\documentclass{article}

\usepackage{PRIMEarxiv}
\usepackage{float} 

\usepackage[utf8]{inputenc} 
\usepackage[T1]{fontenc}    
\usepackage{hyperref}       
\usepackage{url}            
\usepackage{booktabs}       
\usepackage{amsfonts}       
\usepackage{nicefrac}       
\usepackage{microtype}      
\usepackage{lipsum}
\usepackage{graphicx}       
\usepackage{amsmath}       
\usepackage{listings}
\usepackage{xcolor}
\usepackage{fancyhdr}       
\usepackage{graphicx}       
\graphicspath{{media/}}     
\usepackage{caption}
\usepackage{multirow}
\usepackage{soul}

\usepackage[most]{tcolorbox} 
\usepackage{fancyvrb}        
\usepackage{xcolor}          
\usepackage{soul}            

\definecolor{aigold}{RGB}{244,210, 1} 
\definecolor{aigreen}{RGB}{210,244,211} 

\sethlcolor{aigreen}

\definecolor{aired}{RGB}{255,180,181}

\definecolor{aigold}{RGB}{255,180,181}

\definecolor{aiblue}{RGB}{173,216,230} 
 
\definecolor{lightred}{rgb}{1,0.9,0.9} 

\newcommand{\hlred}[1]{{%
  \sethlcolor{lightred}
  \hl{#1}
}}

\newcommand{\hllightgreen}[1]{{%
  \sethlcolor{aigreen}
  \hl{#1}
}}
\newcommand{\hlblue}[1]{{%
  \sethlcolor{aiblue}
  \hl{#1}
}}

\newcounter{textbox}
\renewcommand{\thetextbox}{\arabic{textbox}} 

\newcommand{\textboxcaption}[1]{
  \refstepcounter{textbox}
  \noindent\small\textbf{Text Box \thetextbox:} #1\par\medskip
}

\tcbset{
  custombox/.style={
    width=512.18663pt,
    top=3pt,
    fonttitle=\bfseries\small\sffamily, 
    colback=white,
    colframe=black,
    colbacktitle=black,
    boxrule=0.5pt, 
    colback=white, 
    enhanced,
    rounded corners, 
    center,
    boxed title style={
            sharp corners,
            size=small,
            colframe=blue!75!black, 
        },
    attach boxed title to top left={yshift=-0.1in,xshift=0.15in},
    boxed title style={boxrule=0pt,colframe=white,},
    before=\setlength{\baselineskip}{-2pt}, 
    before upper=\setlength{\baselineskip}{8pt}, 
    after=\setlength{\baselineskip}{8pt}, 
  }
}
\newtcolorbox{LLMbox}[2][]{custombox,title=#2,#1}

\newtcbox{\mybox}[1][green]{on line,
arc=0pt,outer arc=0pt,colback=#1!10!white,colframe=#1!50!black,
boxsep=0pt,left=0pt,right=0pt,top=0pt,bottom=0pt,
boxrule=0pt,bottomrule=0pt,toprule=0pt}

\tcbset{
  customboxmultipage/.style={
    width=474.18663pt,
    top=3pt,
    fonttitle=\bfseries\small\sffamily, 
    colback=white,
    breakable,
    colframe=black,
    colbacktitle=black,
    boxrule=0.5pt, 
    colback=white, 
    rounded corners, 
    center,
    boxed title style={
            sharp corners,
            size=small,
            colframe=blue!75!black, 
        },
    attach boxed title to top left={yshift=-0.1in,xshift=0.15in},
    boxed title style={boxrule=0pt,colframe=white,},
  }
}
\newtcolorbox{LLMboxmultipage}[2][]{customboxmultipage,title=#2,#1}

\lstdefinelanguage{json}{
    basicstyle=\ttfamily\scriptsize, 
    numbers=left,
    numberstyle=\tiny\color{gray}, 
    stepnumber=1,
    numbersep=10pt, 
    showstringspaces=false,
    breaklines=true,
    frame=none,
    backgroundcolor=\color{white},
    literate=
     *{0}{{{\color{blue}0}}}{1}
      {1}{{{\color{blue}1}}}{1}
      {2}{{{\color{blue}2}}}{1}
      {3}{{{\color{blue}3}}}{1}
      {4}{{{\color{blue}4}}}{1}
      {5}{{{\color{blue}5}}}{1}
      {6}{{{\color{blue}6}}}{1}
      {7}{{{\color{blue}7}}}{1}
      {8}{{{\color{blue}8}}}{1}
      {9}{{{\color{blue}9}}}{1}
      {:}{{{\color{red}:}}}{1}
      {,}{{{\color{red},}}}{1}
      {\{}{{{\color{brown}\{}}}{1}
      {\}}{{{\color{brown}\}}}}{1}
      {[}{{{\color{brown}[}}}{1}
      {]}{{{\color{brown}]}}}{1},
}

\lstdefinelanguage{yaml}{
  basicstyle=\ttfamily\footnotesize, 
  morekeywords={slices, sources, model, layer_range, merge_method, base_model, parameters, t, filter, value, dtype}, 
  keywordstyle=\color{blue},
  comment=[l]{\#},
  commentstyle=\color{gray},
  morestring=[b]',
  morestring=[b]",
  stringstyle=\color{orange},
  sensitive=true,
}

\title{Fine-tuning large language models for domain adaptation: Exploration of training strategies, scaling, model merging and synergistic capabilities
\thanks{\textit{\underline{Citation}}: 
\textbf{M.J. Buehler, et al., Fine-tuning large language models for domain adaptation: Exploration of training strategies, scaling, model merging and synergistic capabilities. Pages.... DOI:000000/11111.}} 
}

\author{
  Wei Lu \\
  Civil and Environmental Engineering \\  
  Laboratory for Atomistic and Molecular Mechanics (LAMM) \\
  MIT \\
  Cambridge, MA, USA 
    \And
     Rachel K. Luu \\
  Materials Science and Engineering \\
   Laboratory for Atomistic and Molecular Mechanics (LAMM) \\
  MIT \\
  Cambridge, MA, USA 
   \And
  Markus J. Buehler \\
  Civil and Environmental Engineering and Mechanical Engineering \\  
  Center for Computational Science and Engineering \\
  Schmarzman College of Computing \\  
  Laboratory for Atomistic and Molecular Mechanics (LAMM) \\
  MIT \\
  Cambridge, MA, USA\\ 
  \\
  \texttt{mbuehler@MIT.EDU} 
}

\begin{document}
\maketitle

\begin{abstract}
The advancement of Large Language Models (LLMs) for domain applications in fields such as materials science and engineering depends on the development of fine-tuning strategies that adapt models for specialized, technical capabilities. In this work, we explore the effects of Continued Pretraining (CPT), Supervised Fine-Tuning (SFT), and various preference-based optimization approaches, including Direct Preference Optimization (DPO) and Odds Ratio Preference Optimization (ORPO), on fine-tuned LLM performance. Our analysis shows how these strategies influence model outcomes and reveals that the merging of multiple fine-tuned models can lead to the emergence of capabilities that surpass the individual contributions of the parent models. We find that model merging is not merely a process of aggregation, but a transformative method that can drive substantial advancements in model capabilities characterized by highly nonlinear interactions between model parameters, resulting in new functionalities that neither parent model could achieve alone, leading to improved performance in domain-specific assessments. We study critical factors that influence the success of model merging, such as the diversity between parent models and the fine-tuning techniques employed. The insights underscore the potential of strategic model merging to unlock novel capabilities in LLMs, offering an effective tool for advancing AI systems to meet complex challenges. Experiments with different model architectures are presented, including the Llama 3.1 8B and Mistral 7B family of models, where similar behaviors are observed. Exploring whether the results hold also for much smaller models, we use a tiny LLM with 1.7 billion parameters and show that very small LLMs do not necessarily feature emergent capabilities under model merging, suggesting that model scaling may be a key component. In open-ended yet consistent chat conversations between a human and AI models, our assessment reveals detailed insights into how different model variants perform, and show that the smallest model achieves a high intelligence score across key criteria including reasoning depth, creativity, clarity, and quantitative precision. Other experiments include the development of image generation prompts that seek to reason over disparate biological material design concepts, to create new microstructures, architectural concepts, and urban design based on biological materials-inspired construction principles. We conclude with a series of questions about scaling and emergence that could be addressed in future research.  
\end{abstract}


\section{Introduction}
 
The development of Large Language Models (LLMs)~\cite{Vaswani2017AttentionNeed,AlecRadfordImprovingPre-Training,Xue2021ByT5:Models,Jiang2023Mistral7Bb,Phi-2:Research,dubey2024llama3herdmodels} has enhanced our abilities for natural language processing (NLP) in scientific and engineering applications, due to significant advancements across diverse domains, from general-purpose applications to specialized fields such as materials science and engineering~\cite{Buehler2023MeLMProblemsc, Singhal2023, Qu2024, Yu2024, Hu2023DeepScience, Buehler2024X-LoRA:Design,BuehlerMechGPTModalities, Luu2023BioinspiredLLM:Materials}. These models, including prominent open-source architectures like Llama and Mistral, have demonstrated strong capabilities in understanding and generating human-like text. However, their application in technical fields require fine-tuning strategies that adapt these models to specific domain challenges and technical requirements, which are often poorly understood. In the field of biomateriomics, for instance, researchers aim to develop systematic explorations of knowledge across scales, domains and application areas including biological material design inspiration~\cite{Cranford2012Biomateriomics,Groen2011IntroducingMateriomics,Lee2022AnDiscovery,Arevalo2023LearningMaterials,Smith2023Megalibraries:Discovery,Buehler2023MeLMProblemsc}. These and other challenges can be addressed synergistically using multimodal reasoning engines that have, at its core, capabilities derived from LLMs. One rationale is that LLMs have shown strong capabilities to integrate diverse concepts and provide an integrative modeling strategy for diverse contexts seen in biological materials engineering~\cite{BuehlerGraphReasoning2024,Buehler2023MeLMProblemsc}. 
 
\begin{figure}
    \centering
    \includegraphics[width=1\linewidth]{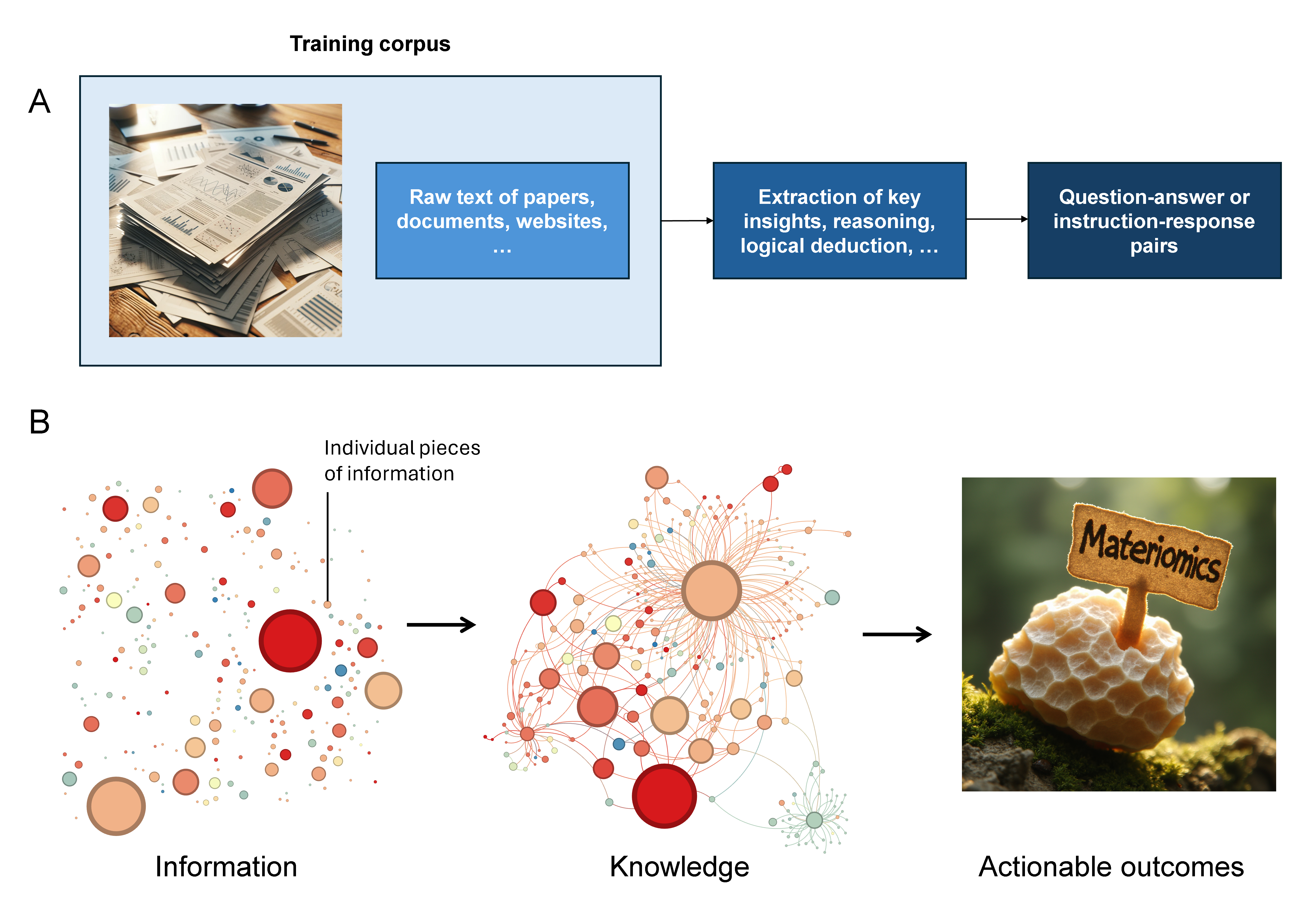}
    \caption{Overview of the approach used in this study, including the scientific training corpus and information processing. Panel A: The training corpus comprises raw text from various sources such as papers, documents, and websites. This text undergoes extraction of key insights, reasoning, and logical deduction, leading to the generation of question-answer or instruction-response pairs. Panel B: Visualization of the transformation from individual pieces of information (here shown as scattered nodes of varying sizes) to a structured network of interconnected insights, illustrating the consolidation of knowledge through the training process. This overall schematic illustrates the goals of this research, to build models for complex problems that integrate distinct features, modalities, and concepts. The image above ``Actionable outcomes'' was generated using \url{lamm-mit/leaf-flux}.}
    \label{fig:Fig_1}
\end{figure}

Fine-tuning LLMs for domain-specific applications involves more than simply retraining on specialized data; it requires the exploration of strategies to endow the model with new knowledge while retaining capabilities learned in earlier training stages, to yield optimal model performance. This is particularly challenging since most of the time it is not feasible to train models from scratch due to cost, or because the original datasets are not available. This is a particular concern in open-source models like Llama or Mistral, where certain details about the training process have been released, but the full datasets during pre-training , fine-tuning, and alignment phases are unknown~\cite{dubey2024llama3herdmodels,Jiang2023Mistral7Bb}. 

An often-effective strategy that has been used in earlier work is low-rank adaptation (LoRA)~\cite{Hu2021LoRA:Models}, where a set of small trainable low-rank tensors is added to linear layers of a larger model to adapt it towards new capabilities~\cite{Hu2021LoRA:Models,Buehler2023MechGPTModalities,Luu2023BioinspiredLLM:Materials}. While this can be an effective method, there are limits to how much a model can be improved and how much new knowledge can be incorporated. 
Other research suggested that continued pre-training (CPT) on domain-specific corpora can help better introduce new knowledge within the target domain~\cite{siriwardhana2024domainadaptationllama370binstructcontinual}, enhancing its relevance and accuracy. However, this typically requires a host of additional strategies to make a model useful for downstream applications, such as instruction following, chat interaction, agentic use, tool calling, and others. Supervised fine-tuning (SFT) is a method used to refine these models by directly teaching them to perform well on specific tasks through curated datasets. However, the potential for further enhancing model performance and unlocking new capabilities through advanced optimization techniques remains a critical area of exploration. In this context, preference-based optimization strategies, such as Direct Preference Optimization (DPO)~\cite{rafailov2024directpreferenceoptimizationlanguage} and Odds Ratio Preference Optimization (ORPO)~\cite{hong2024orpomonolithicpreferenceoptimization}, have emerged as promising approaches. Unlike traditional Reinforcement Learning (RL) methods~\cite{christiano2023deepreinforcementlearninghuman}, which often require explicit reward functions and complex environment models, DPO and ORPO focus on optimizing models based on direct feedback or preferences. These methods offer a flexible and efficient means of refining model behavior, particularly when the goal is to align the model’s outputs with human expectations or domain-specific criteria, such as being able to reason over or logically deduce answers in a particular domain, such as materials science.

Another area of recent interest is the practice of model merging~\cite{garipov2018loss,utans1996weight,goddard2024arceesmergekittoolkitmerging}, where multiple, differently trained models are combined to create a new model with potentially superior capabilities. Earlier experiments have shown that this process is not simply additive; as we will show it leads to highly nonlinear interactions between the parameters of the merged models, resulting in the emergence of new functionalities that neither parent model possessed individually. Such emergent behavior suggests that model merging could be a powerful tool for advancing LLM capabilities, enabling the development of models that are not only more accurate but also more adaptable to complex, real-world challenges.

As this brief review shows, there are a myriad of possible strategies, but relatively little data is available in terms of systematic explorations. LLMs are highly complex models, and training is expensive and time-consuming, and often developers focus on a particular approach that yielded acceptable results. Here we take a different approach and specifically investigate the effects of various fine-tuning and optimization strategies on LLM performance, with a particular focus on a systematic, consistent set of experiments as summarized in Table~\ref{tab:approaches_summary}. Figure~\ref{fig:Fig_1} shows an overview of the training corpus and information processing, covering both the general process of utilizing different types of data (raw, processed/distilled, conversational, etc.) and a visualization of the transformation from individual pieces of information to a structured network of interconnected insights.
 
\begin{table}[h]
\centering
\scriptsize 
\setlength{\tabcolsep}{3pt} 
\begin{tabular}{|p{3.5cm}|p{4cm}|p{4cm}|p{3.5cm}|}
\hline
\textbf{Approach} & \textbf{Definition} & \textbf{Example} & \textbf{Dataset Type} \\ \hline
CPT (Continued Pre-Training) & Continuing the training of a language model on a specific domain or additional data after initial pre-training (no template/instruction format use). & Enhancing model knowledge in specialized fields like materiomics, bioinformatics or broader fields like materials science. & Raw text from papers plus text with reasoning, summarization, distillation \\ \hline
SFT (Supervised Fine-Tuning) & Fine-tuning a pre-trained model using labeled data with supervised learning techniques. & Adapting models to specific tasks such as QA, reasoning, scientific method, or dialogue systems. & Question-answer or instruction-answer pairs, conversations \\ \hline
DPO (Direct Preference Optimization) & Optimizing a model by directly learning from preferences, often involving human feedback or ranking. & Aligning models with human preferences for content generation, physics-awareness, reasoning, or human considerations. & Positive/negative examples \\ \hline
ORPO (Odds Ratio Preference Optimization) & A reference model-free monolithic odds ratio preference optimization algorithm that eliminates the need for a separate preference alignment phase. & Refining models without a baseline reference, using direct odds ratio for preference tasks. & Positive/negative examples \\ \hline
Model Merging & Combining multiple models or checkpoints into a single model, e.g. using techniques like spherical linear interpolation (SLERP). & Creating hybrid models that incorporate strengths from different pre-trained models. & No training data used \\ \hline
\end{tabular}
\caption{Summary of model development approaches explored with datasets used, and a brief description of capabilities.}
\label{tab:approaches_summary}
\end{table}

The plan of the paper is as follows. We first present a brief introduction into key concepts, then present results along with a discussion, followed by conclusions. We present detailed methods and references to codes and model weights, and discuss our dataset creation, development and training mechanisms. 

\section{Results and Discussion}
\label{sec:Results_Discussion}

\begin{figure} 
    \centering
    \includegraphics[width=1\linewidth]{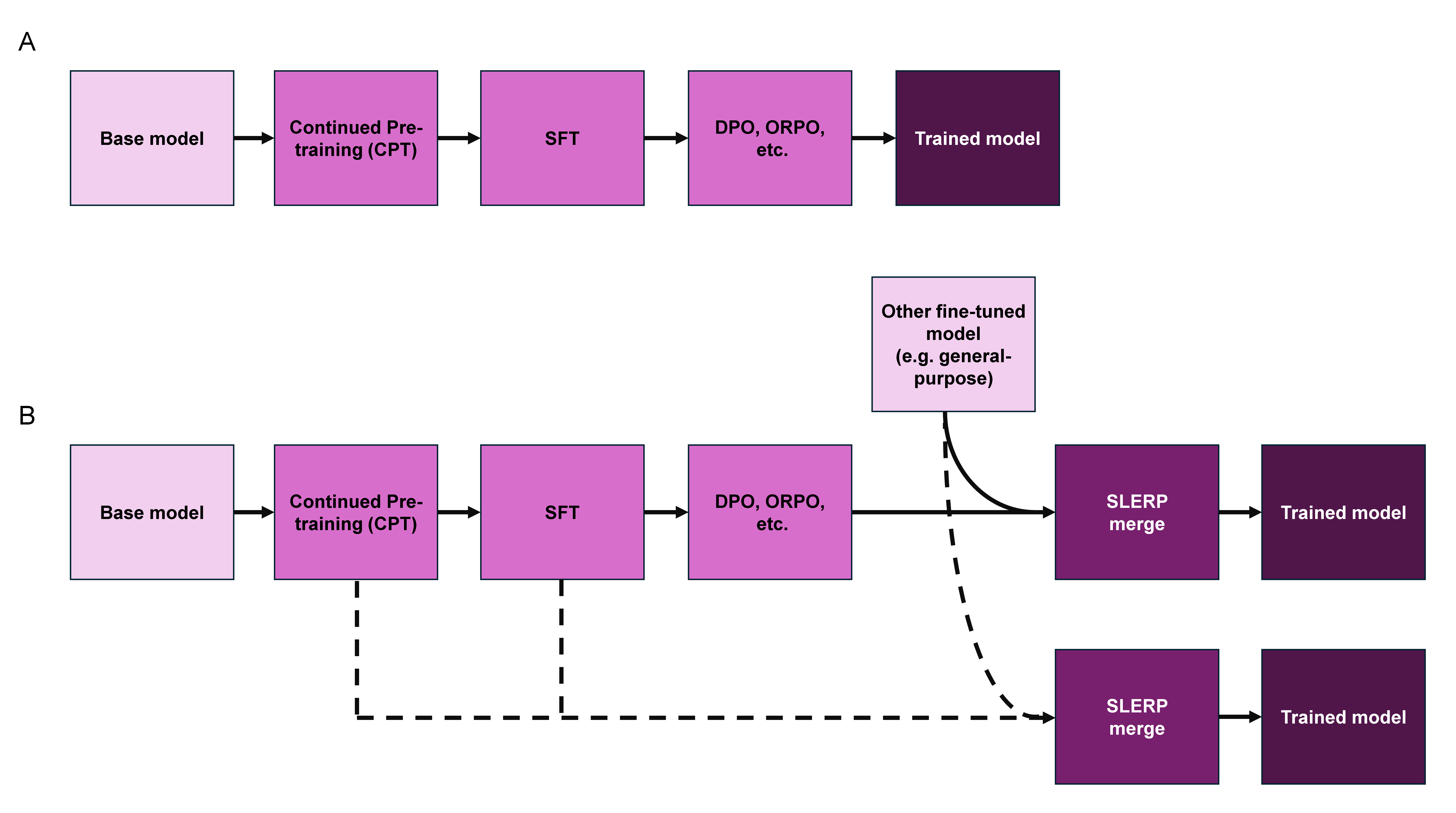}
    \caption{Model training, merging and assessment stages. Panel A: A conventional training pipeline where a base model undergoes Continued Pre-Training (CPT), followed by Supervised Fine-Tuning (SFT), and then optimized using methods like Direct Preference Optimization (DPO) or Odds Ratio Preference Optimization (ORPO) to produce a trained model. Assessment of the model can be performed at each of the stages, such as using the SFT results for benchmarking. Panel B: An alternative pipeline where, after CPT, SFT, and optimization (e.g., DPO, ORPO), the model is further enhanced by merging it with another fine-tuned model (e.g., a general-purpose model). Merging can be done with models extracted from various training stages, such as after CPT, SFT or at the final stage.}
    \label{fig:Fig_1000}
\end{figure}

We follow the process depicted in Figure~\ref{fig:Fig_1000} in developing models and conducting assessments. Figure~\ref{fig:Fig_1000}A shows a conventional linear training pipeline where a base model undergoes Continued Pre-Training (CPT), followed by Supervised Fine-Tuning (SFT), and then optimized using methods like Direct Preference Optimization (DPO) or Odds Ratio Preference Optimization (ORPO) to produce a trained model. Figure~\ref{fig:Fig_1000}B shows an alternative pipeline where, after CPT, SFT, and optimization (e.g., DPO, ORPO), the model is further enhanced by merging it with another fine-tuned model (e.g., a general-purpose model). We note that model merging can be done with models extracted from various training stages, such as after CPT, SFT or at the final stage.

\begin{figure}
    \centering
    \includegraphics[width=.75\linewidth]{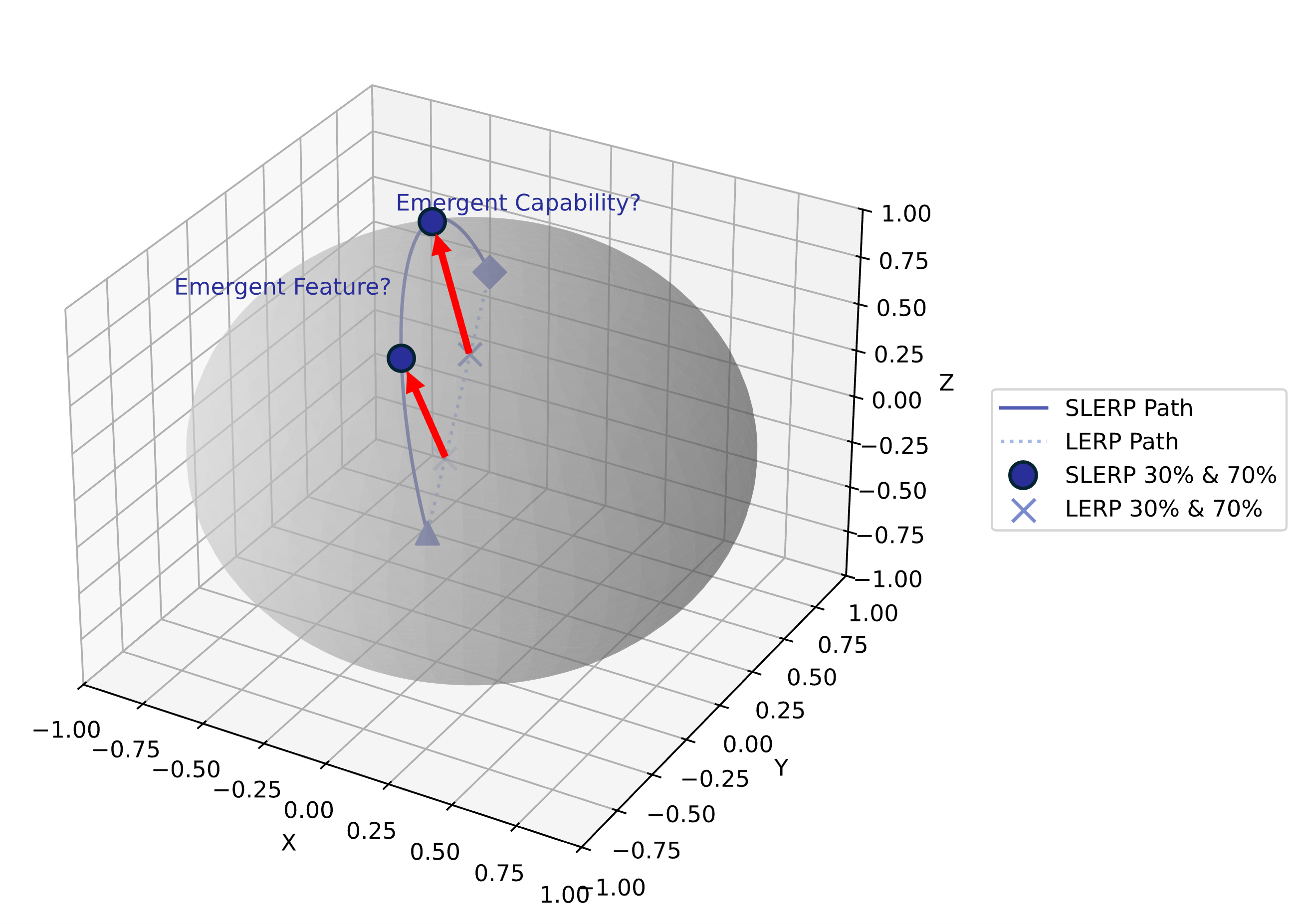}
    \caption{Comparison of SLERP (Spherical Linear Interpolation) and LERP (Linear Interpolation) between two points on a unit sphere, illustrating their application in merging Large Language Model (LLM) parameters. SLERP interpolates between points $\mathbf{p}_1$ and $\mathbf{p}_2$ along a spherical path on the surface of the sphere, calculated as $\text{SLERP}(t)$, where $t$ is the interpolation parameter (equations see main text). In contrast, LERP interpolates linearly between the same two points, following a straight line through the sphere. Intermediate points at 30\% and 70\% along both paths are highlighted, showing the difference in how SLERP and LERP handle interpolation. In the context of LLMs, SLERP is particularly effective for merging model parameters from different pre-trained models, facilitating the emergence of new abilities that neither parent model possessed alone. The smooth, nonlinear path of SLERP helps to preserve the underlying structure of the model parameters, represented by the unit sphere, potentially unlocking novel interactions between features that lead to enhanced performance and the development of emergent capabilities. The sphere in this context represents the inherent structure of the model's parameter space, and by maintaining the geometric relationship between the parameters, SLERP ensures that the interpolation respects this original structure and does not puncture it (as the LERP points would), leading to meaningful and coherent blending of capabilities rather than random, unstructured changes. A key point is that because the merged points are both congruent with the model geometry (that is, they lie on the sphere used here for demonstration) and because they realize new points previously not accessed, emergent features and capabilities could potentially be unlocked.}
    \label{fig:Fig_11}
\end{figure}

For the purpose of the analysis, we go into the details of model merging strategies. In this work we focus on (Spherical Linear Interpolation (SLERP, details see Materials and Methods, Section~\ref{sec:Materials_Methods}), as we found it to be the most effective method. SLERP is a mathematical technique originally introduced in the field of computer graphics for smoothly interpolating between rotations represented by quaternions \cite{shoemake1985animating}. SLERP has found widespread application in various domains that require smooth transitions between orientations or states, including robotics, physics simulations, as well as real-time graphics. For instance, in robotics, SLERP is used for the practical parameterization of rotations, allowing for seamless motion planning and control \cite{grassia1998practical}. In physics simulations and computer graphics, SLERP is crucial for visualizing and animating rotations in a way that preserves the continuity and smoothness of motion \cite{hanson2006visualizing, eberly2001game}. By maintaining the geometric relationships between interpolated states, SLERP ensures that transitions are both smooth and physically meaningful, making it a useful tool in scenarios where precise and continuous interpolation is required. Figure~\ref{fig:Fig_11} shows the basic concepts of SLERP (versus linear interpolation, LERP), visually. A key aspect of this strategy is that the smooth, nonlinear path helps to preserve the underlying structure of the model parameters. The sphere in this context represents the inherent structure of the model's parameter space, and by maintaining the geometric relationship between the parameters, SLERP ensures that the interpolation respects this original structure and does not puncture it (as linear combination of points would), leading to a more meaningful and coherent blending of capabilities rather than random, unstructured changes. Because the merged points are both congruent with the model geometry (that is, they lie on the sphere used here for demonstration) and because they realize new points previously not accessed, emergent features and capabilities could potentially be unlocked.

\begin{figure}
    \centering
    \includegraphics[width=.8\linewidth]{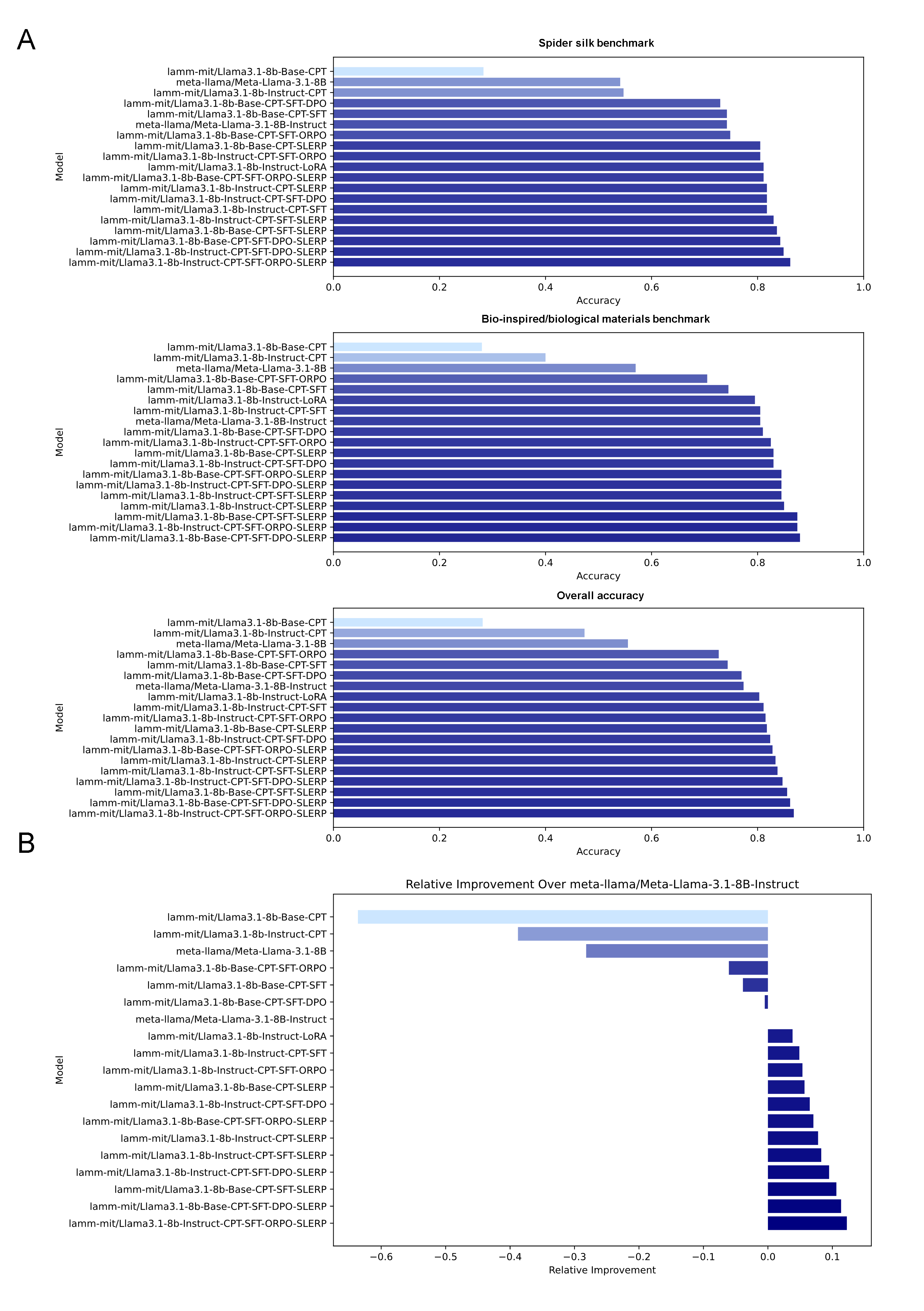}
    \caption{Performance evaluation $P$ of \texttt{Llama-3.1} model variants across benchmarks. Panel A: Accuracy results for various variants on different benchmarks: Spider Silk, Bio-inspired/Biological Materials, and Overall Accuracy. The models were evaluated after undergoing different training and optimization strategies (CPT, SFT, ORPO/DPO, model merging). Panel B: Relative improvement of model variants over the \texttt{meta-llama/Meta-Llama-3.1-8B-Instruct} baseline model. This highlights how each training strategy contributes to the model's performance gains or losses across the various benchmarks, providing insight into the effectiveness of different approaches. It is notable that models that underwent CPT, SFT, and to some extent preference optimization (e.g., DPO, ORPO) show a deterioration in performance, as indicated by negative relative improvement values. However, after applying the SLERP merging technique, these same models exhibit significant performance gains, surpassing the baseline model. This highlights the effectiveness of model merging in combining the strengths of different specialized models, resulting in a robust final model with superior overall performance. Overall, the results show that the models that have undergone SLERP merging (especially those combined with DPO and ORPO strategies) generally show the highest accuracy across benchmarks. Merging in this case is always done with \texttt{meta-llama/Meta-Llama-3.1-8B-Instruct}. All models have been trained with the same datasets in all stages, as shown in Table~\ref{tab:dataset_summary}.}
    \label{fig:Fig_2}
\end{figure}

In the following, we present a series of results from assessment experiments conducted with different model families and training/merging strategies (details on training, models, datasets, and assessment benchmarks, see Materials and Methods). Figure~\ref{fig:Fig_2} depicts a series of performance evaluations of \texttt{Llama-3.1} Model variants across benchmarks. We use two basic models as foundation for our training. First, \texttt{meta-llama/Meta-Llama-3.1-8B}, the base model of the Llama family that has not been fine-tuned and aligned. Second, the \texttt{meta-llama/Meta-Llama-3.1-8B-Instruct} model that has been fine-tuned and aligned to conduct question-answer interactions, along with a host of other capabilities~\cite{dubey2024llama3herdmodels}. Except for the LoRA case~\cite{Luu2023BioinspiredLLM:Materials}, all of our experiments include CPT (see Table~\ref{tab:approaches_summary} for an overview of the training stages and acronyms used) as the first step, with the aim to endow the base model with domain knowledge from our materials science corpus of papers and distilled, extracted and processed data sourced from scientific studies. We then implement a range of variations, such as CPT only, CPT-SFT, CPT-SFT-ORPO and CPT-SFT-DPO. At each stage, we also implement model merging with the with the \texttt{meta-llama/Meta-Llama-3.1-8B-Instruct} model. 
Overall, the results reveal that the models that have undergone SLERP merging (especially those combined with DPO and ORPO strategies) generally show the highest accuracy across benchmarks. The best strategy without model merging is found to be the Instruct-CPT-SFT-DPO strategy.

\begin{figure}
    \centering
    \includegraphics[width=.8\linewidth]{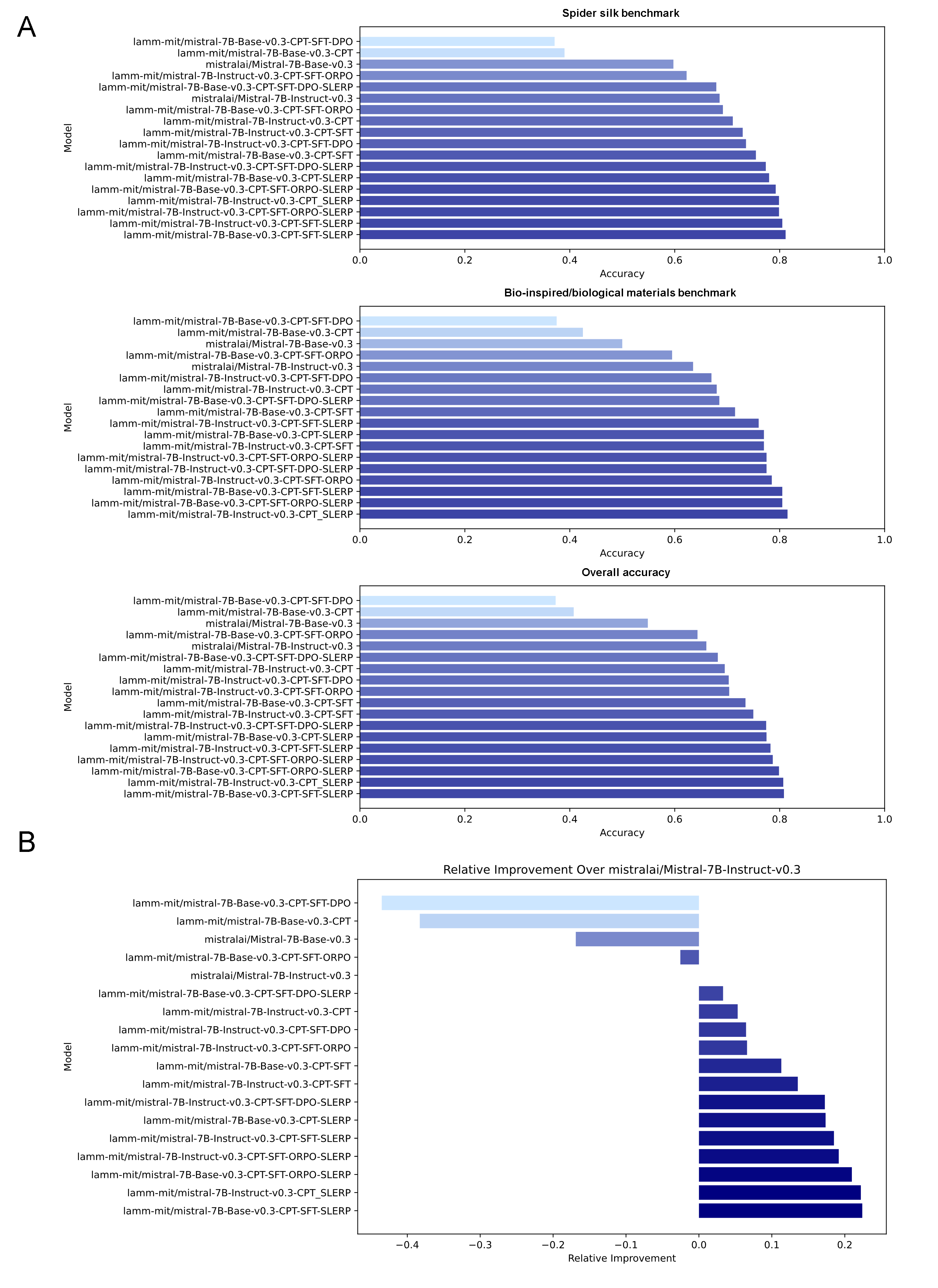}
    \caption{Performance evaluation $P$ of \texttt{Mistral-7B-v0.3} model variants. Panel A: Accuracy results for various Mistral-7B-v0.3 model variants on the Spider Silk, Bio-inspired/Biological Materials, and Overall Accuracy benchmarks. Initial models trained with CPT and SFT show moderate performance. Models further optimized using ORPO or DPO exhibit significant improvements in accuracy across all benchmarks. Model merging results in further significant improvements. The relative improvements are even more pronounced than those seen in the \texttt{Llama-3.1} models (here exceeding 20\% versus around 12\%), indicating the particular effectiveness of these techniques for the Mistral series. Panel B: Relative improvement of model variants over the baseline \texttt{mistralai/Mistral-7B-Instruct-v0.3} model. The Base model subjected to CPT alone initially show a decrease in relative performance. However, after SFT, ORPO and especially after applying SLERP merging, especially with ORPO or DPO optimization, these models demonstrate substantial positive relative improvement, surpassing the baseline by a greater margin than the improvements seen in the \texttt{Llama-3.1} models. This highlights the powerful impact of these combined strategies in enhancing the overall performance of the Mistral models. It is notable that a direct merge of the Base-CPT-SFT model results in significant performance, close to the Instruct-CPT-SFT strategy. Merging is always done with  \texttt{mistralai/Mistral-7B-Instruct-v0.3}. The same training set is used for all experiments, as defined in Table~\ref{tab:dataset_summary}. }
    \label{fig:Fig_3_B}
\end{figure}

We now conduct the same series of experiments using \texttt{Mistral-v0.3} model variants~\cite{Jiang2023Mistral7Bb} across benchmarks. As in the previous set of results, we use the same dataset across all cases, and we present both non-merged cases and merges with the \texttt{mistralai/Mistral-7B-Instruct-v0.3} model.
Figure~\ref{fig:Fig_3_B} depicts an overview the performance evaluations across benchmarks for this case. As before, the results show that these models that have also undergone SLERP merging generally show the highest accuracy across benchmarks. The best strategy without model merging is found to be the Base-CPT-SFT strategy, albeit the performance of the Instruct-CPT-SFT strategy is very similar.

\begin{figure}
    \centering
    \includegraphics[width=.8\linewidth]{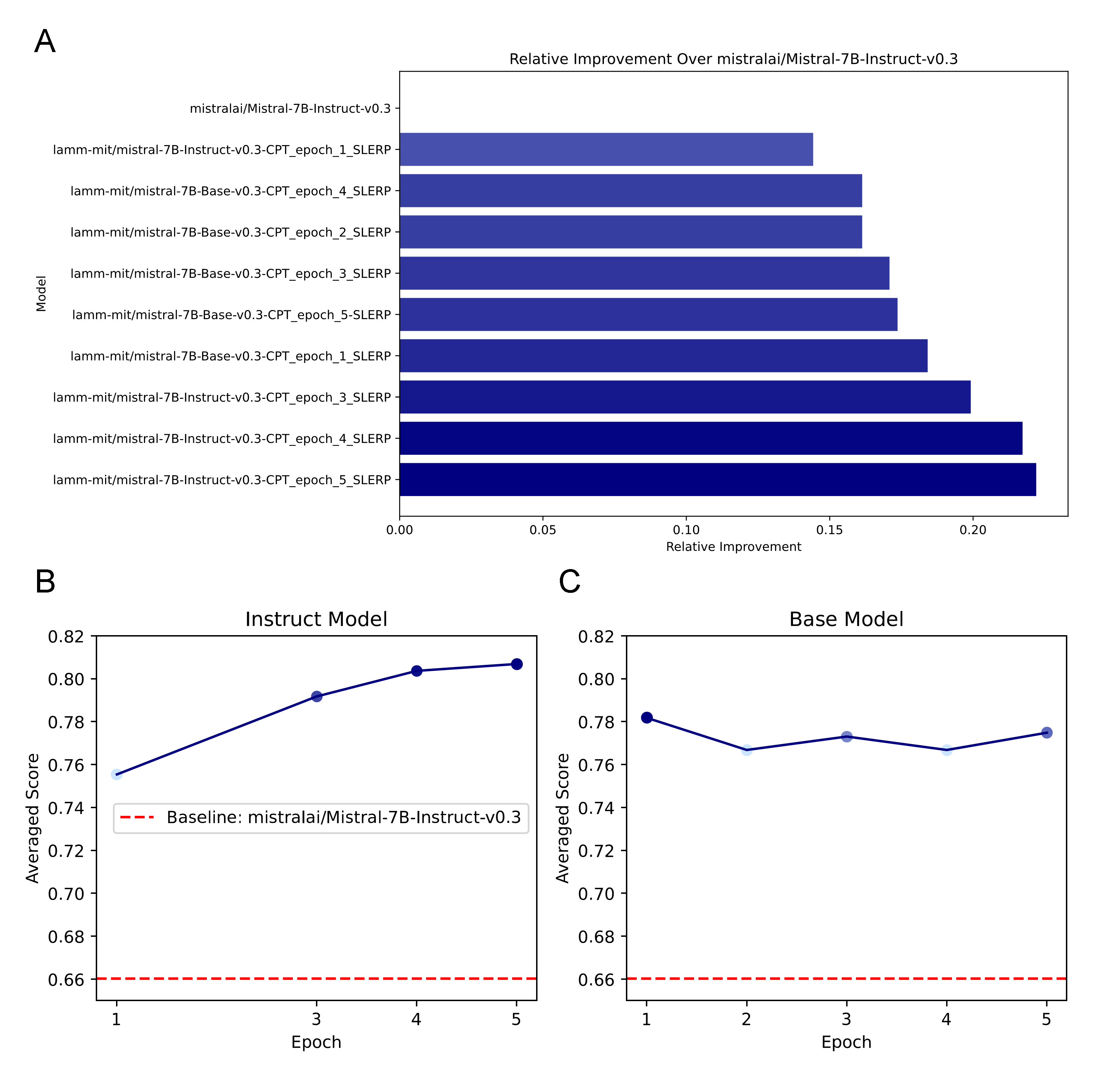}
   \caption{Comparison of averaged scores across different epochs for both the Base and Instruct models fine-tuned with the SLERP method. The baseline score for the \texttt{mistralai/Mistral-7B-Instruct-v0.3} model is indicated by the red dashed line in both subplots. Panel A shows an overview of the results, in similar format as the earlier performance assessments (Figures~\ref{fig:Fig_2} and ~\ref{fig:Fig_3_B}), showing performance across all models and variants of CPT epochs used. Panel B shows the performance of the Instruct model over five training epochs, while panel C subplot illustrates the performance of the Base model over five epochs. }

    \label{fig:Fig_27}
\end{figure}

The CPT stage involves five epochs. To explore the effect of the number of epochs in this phase, we computed performance of the direct CPT-SLERP merges for the Mistral models from different training epochs. It is noted that the original merges assessed (and SFT, DPO/ORPO training stages) in Figure~\ref{fig:Fig_3_B} were conducted based on CPT results from epoch 5.
Figure~\ref{fig:Fig_27} depicts a comparison of averaged scores across different epochs for both the Base and Instruct models, using the SLERP method. Figure~\ref{fig:Fig_27}A shows an overview of the results, in similar format as the earlier performance assessments, depicting performance across all models and variants of CPT epochs used. Figure~\ref{fig:Fig_27}B shows the performance of the Instruct model as a function of the number of CPT training epochs, and Figure~\ref{fig:Fig_27}C illustrates the performance of the Base model. 
We can see that the Instruct model demonstrates a consistent improvement in performance with each epoch, peaking at the best score by epoch 5, indicating that it benefits significantly from continued training. In contrast, the Base model shows a more fluctuating performance, with its highest score at epoch 1, followed by slight declines and only a minor recovery at epoch 5. This suggests that while the Base model starts strong, it does not consistently improve with additional training, potentially indicating a saturation point. Both models, however, consistently outperform the baseline score set by the original \texttt{mistralai/Mistral-7B-Instruct-v0.3} model, underscoring the effectiveness of the SLERP method, and consistent with the earlier results. The more substantial improvement of the Instruct model over the baseline highlights its robustness in instruction-tuned tasks, making it the preferable choice for such applications, particularly when extended fine-tuning is feasible.

\subsection{Detailed analysis of key factors in model merging}

As the results in Figures~\ref{fig:Fig_2} and ~\ref{fig:Fig_3_B} clearly reveal, SLERP appears to significantly improve model performance due to its ability to respect the geometric properties of the parameter space. However, this analysis did not yet reveal whether we have a significant synergistic effect. To examine this, we plot the results differently, comparing the actual measured performance with an expected performance that is computed by simply averaging the scores of the two parent models. To properly define all key variables, the performance of a merged model is defined as \( P_{\text{merged};P_1, P_2} \) (measured per the benchmark), while the expected, averaged score \( E (P_1, P_2) \) is calculated as the linear average of the performances of the two parent models:
\[
E  (P_1, P_2)  = \frac{P_1 + P_2}{2}.
\]

\begin{figure}
    \centering
    \includegraphics[width=1\linewidth]{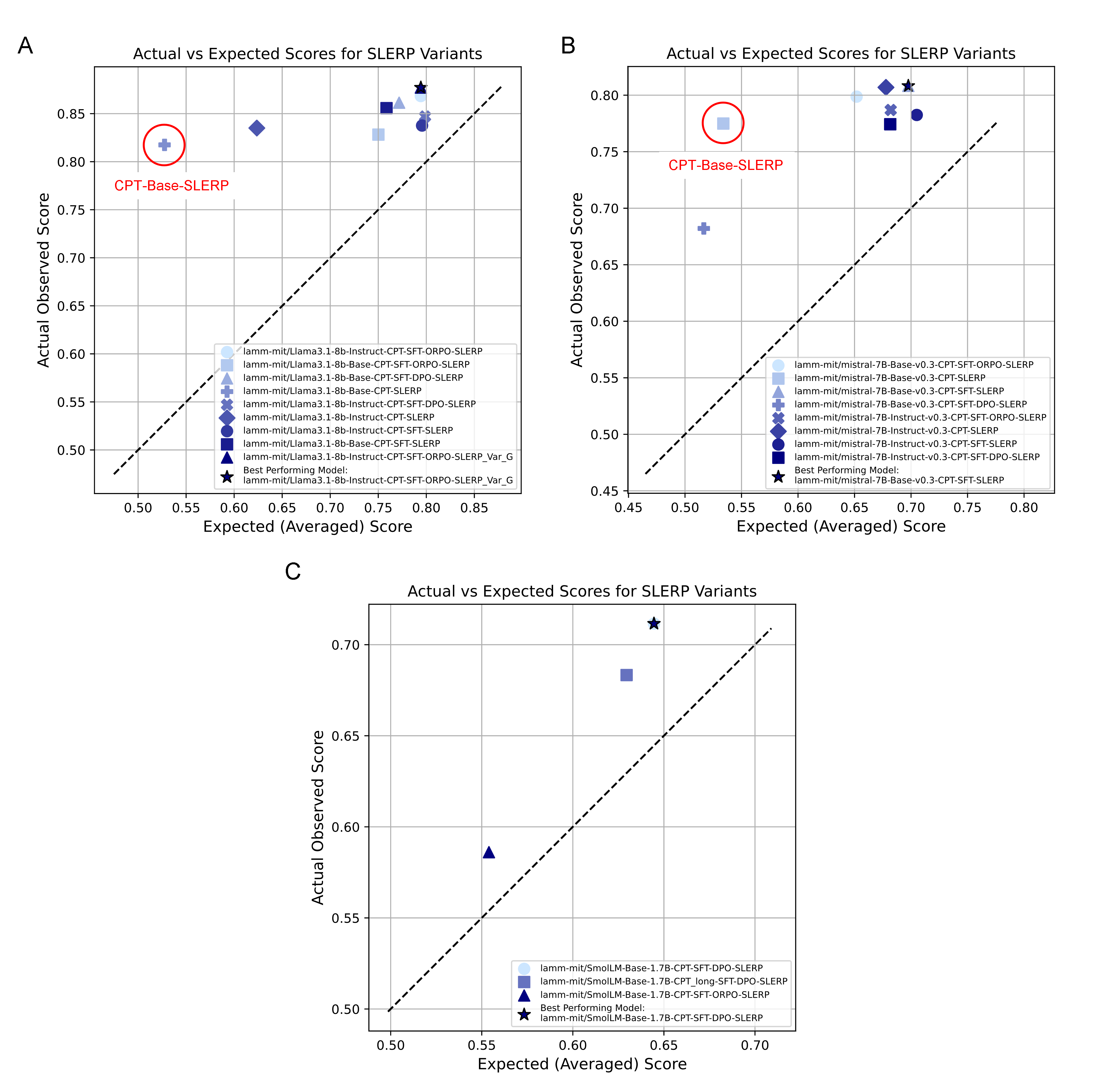}
    \caption{Exploration of performance of SLERP variants for different cases, plotting the actual observed performance \( P_{\text{merged};P_1, P_2} \) over a linear, expected score $E  (P_1, P_2)$ based on a simple average of the score of both parent models (linear combination). The deviation from the diagonal shows clear nonlinear, synergistic effects, where the actual observed model performance is much greater than a simple averaging of the capabilities of the parent models alone. Results are shown for both the \texttt{Llama-3.1} (panel A) and \texttt{Mistral-7B-v0.3} (panel B) model series, respectively. Panel C shows results for the much smaller \texttt{SmolLM} family of models, where the deviation of the observed score from the expected score is not as significant, and even below the best performance of one of the pre-merged models (see, Figure~\ref{fig:Fig_109} for a detailed analysis).  Results for the Llama and Mistral models are similar across all experiments and show a clear nonlinear effect of model merging. As marked with a red circle, \textcolor{red}{$\circ$}, the CPT-Base-SLERP strategy tends to yield some of the highest deviation from the expected score and is, at the same time, a relatively straightforward training strategy. }
    \label{fig:Fig_5}
\end{figure}

Using these metrics, Figure~\ref{fig:Fig_5} shows a detailed exploration of performance of SLERP variants for different cases, plotting the actual observed performance over an estimated, expected score based on a simple average of the score of both parent models (linear combination). 

Notably, the strong deviation from the diagonal reveals  nonlinear, synergistic effects, where the actual observed model performance is much greater than a simple averaging of the capabilities of the parent models alone. Results are shown for both the \texttt{Llama-3.1-8B} and \texttt{Mistral-7B-v0.3} model series, respectively, for a variety of training strategies and datasets used in the process. We find that the results are similar for both models. 
An important distinction that can be seen in the analysis is that for the Llama models, the best performing model (\texttt{lamm-mit/Llama3.1-8b-Instruct-CPT-ORPO-SLERP}) is based off the Llama Instruct model, whereas for the Mistral model (\texttt{lamm-mit/mistral-7B-Base-v0.3-CPT-SFT-SLERP}) it is based off the Mistral Base model.

To better understand the mechanics behind the observed effects, we briefly discuss the mathematical underpinnings of SLERP merging. Unlike linear interpolation, which assumes a flat Euclidean space, SLERP explores a richer parameter space by interpolating along a curved path on a unit sphere (we refer also to Figure~\ref{fig:Fig_11}). This approach allows SLERP to uncover regions in the parameter space that might represent combinations of parameters more effective than those found in either model alone. SLERP further balances the specialized knowledge learned by each model, combining their strengths without simply averaging them. By avoiding high-loss regions that linear interpolation might pass through, SLERP ensures a smoother transition, potentially leading to better generalization in the merged model. The non-linear nature of SLERP's path also considers the complex interactions between parameters, which can reveal beneficial interactions that a simple linear combination would miss. Furthermore, SLERP may act as a form of regularization, preventing overfitting to the idiosyncrasies of a single model's training data, thus enhancing generalization. Finally, SLERP helps mitigate the effects of catastrophic forgetting, preserving knowledge from both models when one has been fine-tuned or trained after the other. These factors combine to make SLERP a powerful tool for model merging, leading to a merged model that often performs better than either of the original models on their own.

Hence, we believe that the observed effectiveness of SLERP in merging models can be attributed to its ability to enhance non-linear interactions between parameters by exploring the spherical geometry of the parameter space. Given two sets of model parameters \( \theta_1 \) and \( \theta_2 \), each parameter can be seen as a vector in a high-dimensional space. The interpolation performed by SLERP respects the curvature of this space, allowing for combinations of parameters that are not simply linear but involve deeper, non-linear synergies (see, Figure~\ref{fig:Fig_11}). Consider the parameters \( \theta_1 \) and \( \theta_2 \) as consisting of individual components \( \theta_{1,i} \) and \( \theta_{2,i} \) in a given layer of the neural network. SLERP combines these parameters as follows:
\[
\theta_{i,\text{merged}} = \|\theta_1\|^{1-t} \|\theta_2\|^t \left( \frac{\sin((1-t) \omega)}{\sin(\omega)} \hat{\theta}_{1,i} + \frac{\sin(t \omega)}{\sin(\omega)} \hat{\theta}_{2,i} \right)
\]
This combination allows for interactions between \( \theta_{1,i} \) and \( \theta_{2,i} \) that are non-linear in nature. For example, if \( \theta_{1,i} \) and \( \theta_{2,i} \) represent weights connected to different features in the network, their spherical combination could activate a new feature \( \phi_i \) that is not present in either model individually:
\[
\phi_i = f\left( \theta_{i,\text{merged}} \cdot x_i \right)
\]
where \( x_i \) is the input feature and \( f(\cdot) \) is the activation function. The non-linear combination of parameters may lead to new behaviors or capabilities, as the interpolated parameters could synergistically enhance or suppress features in ways that the individual models cannot.

SLERP avoids destructive interference by maintaining the angular relationships between the parameter vectors, which can prevent the loss of specialized features learned by either model. The spherical symmetry imposed by SLERP introduces a regularization effect, smoothing the transition between the models and enabling the merged model to generalize better. This process often results in the emergence of new capabilities or improvements in performance that neither of the original models possessed.

The ability of SLERP to uncover these new capabilities can also be understood through the lens of overparameterization and the principles of ensemble methods. Overparameterized neural networks are known to generalize well, even when trained to zero error, due to their increased capacity to capture complex patterns~\cite{belkin2019reconciling}. SLERP leverages this capacity by combining parameters in a non-linear fashion, effectively utilizing the high-dimensional space in which these parameters reside. As a result, the merged model can exhibit emergent properties that are not apparent in either of the original models. SLERP's mechanism resembles ensemble methods, where combining diverse models leads to better generalization~\cite{hansen1990neural}. In this case, the diversity comes from the different training histories and learned features of the two models. The spherical interpolation pathway created by SLERP acts as a continuum of model ensembles, where at each point along the path, the combined parameters may activate new and beneficial feature interactions. SLERP not only preserves the strengths of the individual models but also has the potential to generate entirely new capabilities through its sophisticated interpolation method. This makes it a useful tool for our goal to merge models that complement each other or to create a more versatile and generalizable model from existing pre-trained models.

\begin{figure}
    \centering
    \includegraphics[width=1\linewidth]{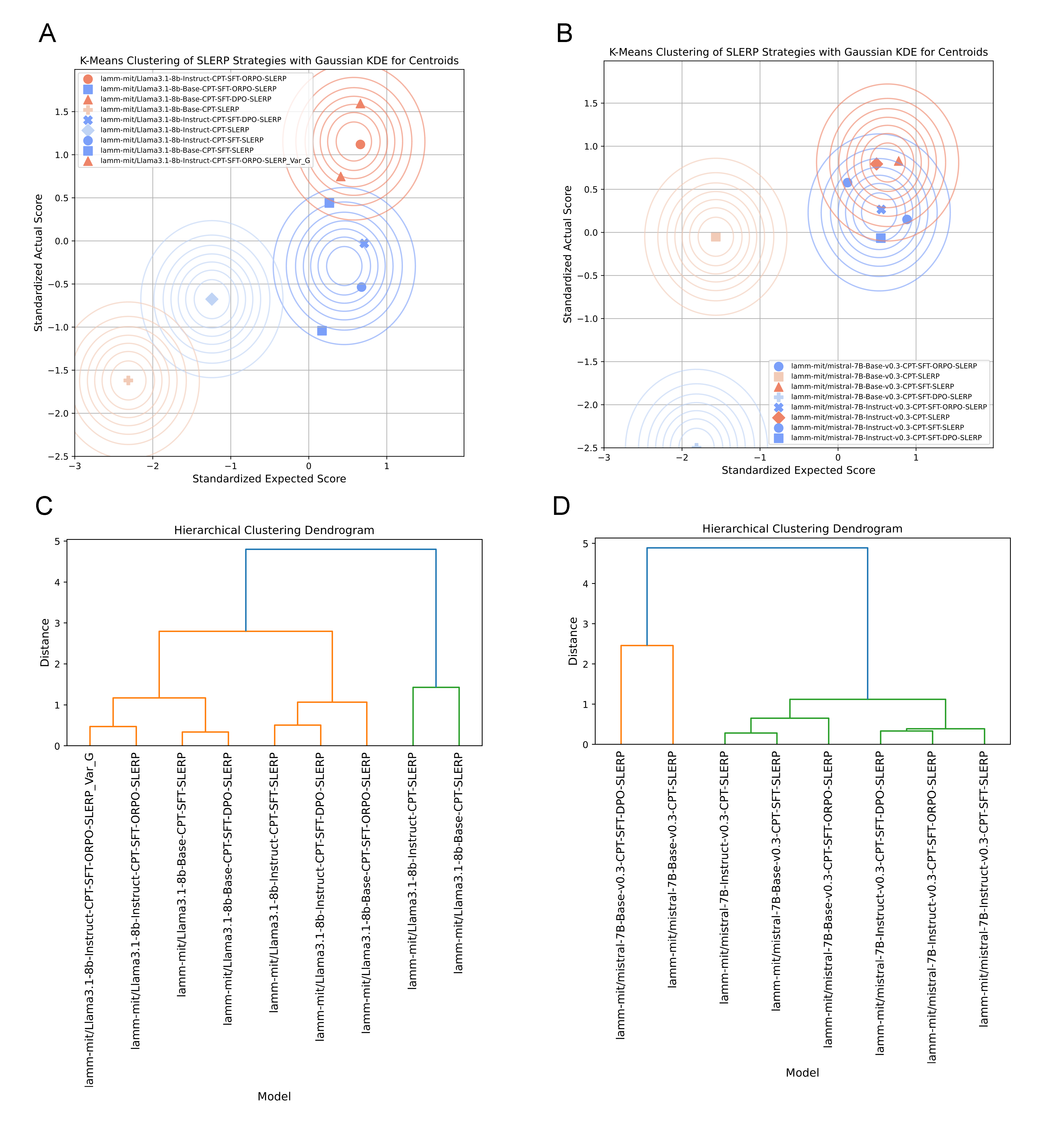}
    \caption{Clustering analysis of SLERP strategies and hierarchical clustering dendrograms for model performance. Panel A: K-Means clustering of \texttt{Llama-3.1-8b} models using standardized expected and actual scores, with Gaussian KDE (Kernel Density Estimation) applied to visualize the centroids. The clustering reveals distinct groupings based on the performance outcomes of various SLERP strategies. Panel B: K-Means clustering of \texttt{Mistral-7B-v0.3} models using the same approach as in panel A. Similar to the Llama models, distinct clusters emerge, highlighting the different performance profiles of the models post-SLERP merging.   Panel A: Hierarchical clustering dendrogram for the \texttt{Llama-3.1-8b} models based on the clustering analysis. Panel B: Hierarchical clustering dendrogram for the \texttt{Mistral-7B-v0.3} models. The dendrograms in panels C and D, respectively, reveal how different models cluster together, indicating that these strategies yield similar performance outcomes. The comparison between the \texttt{Llama-3.1-8b} (panel C) and \texttt{Mistral-7B-v0.3} dendrograms (panel D) shows that while both models respond well to SLERP strategies, the Mistral models exhibit more distinct clustering patterns. 
}

    \label{fig:Fig_6}
\end{figure}

We examine the variations and potential trends in the strategies explored using clustering analysis. Figure~\ref{fig:Fig_6} provides a comprehensive clustering analysis of SLERP strategies applied to both the \texttt{Llama-3.1-8b} and \texttt{Mistral-7B-v0.3} models, and the resulting impact on their performance. We explore the use of two methods. First, K-Means clustering, a partition-based method that groups data into a predefined number of clusters by minimizing the distance between data points and the cluster centroids, providing insight into the natural groupings of models based on their expected and actual performance. Second, we use hierarchical clustering, an agglomerative method that creates a tree-like structure, a dendrogram, to show the nested relationships between models at various levels of similarity, revealing the hierarchical organization and potential sub-groupings within the data.

Figure~\ref{fig:Fig_6}A illustrates the K-Means clustering of the \texttt{Llama-3.1-8b} models using standardized expected and actual scores, with Gaussian KDE (Kernel Density Estimation) applied to visualize the centroids. The analysis reveals distinct groupings that correspond to different SLERP strategies, indicating that specific merging techniques produce closely related performance outcomes.
Figure~\ref{fig:Fig_6}B presents a similar K-Means clustering for the \texttt{Mistral-7B-v0.3} models. Here too, distinct clusters emerge, showing that the SLERP strategies significantly influence the models' performance profiles. Notably, the clustering patterns observed in the Mistral models are more pronounced compared to the Llama models, suggesting that the Mistral architecture might be more sensitive to these optimization and merging strategies.

Across both the Llama and Mistral models, the K-Means analysis clearly delineates two performance-based clusters. Models that incorporate multiple fine-tuning strategies, especially ORPO, consistently form clusters with higher actual scores, outperforming models that rely on simpler strategies. This suggests that the complexity and thoroughness of the fine-tuning process play a crucial role in achieving better model performance, as indicated by the clustering results. Next we explore hierarchical clustering as a way to better break down these distinctions. 

To do this, we use a dendrogram analysis. A dendrogram is a tree-like diagram that displays the arrangement of clusters generated through hierarchical clustering. This visualization helps elucidate the relationships among the models, with closely related models (in terms of performance) clustering together. The dendrogram reveals that models employing similar training strategies are grouped into distinct subclusters, highlighting the effectiveness of these approaches in shaping model performance.
Figure~\ref{fig:Fig_6}C introduces the hierarchical clustering dendrogram for the \texttt{Llama-3.1-8b} models, and Figure~\ref{fig:Fig_6}D for the \texttt{Mistral-7b} models. 
The dendrogram demonstrates how different models cluster together, indicating similar performance outcomes. When comparing the dendrograms of the Llama and Mistral models, it becomes evident that while both models are positively influenced by SLERP strategies, the Mistral models show more defined clustering patterns. This suggests a stronger impact of the various strategies on the Mistral architecture.

\begin{figure}
    \centering
    \includegraphics[width=.75\linewidth]{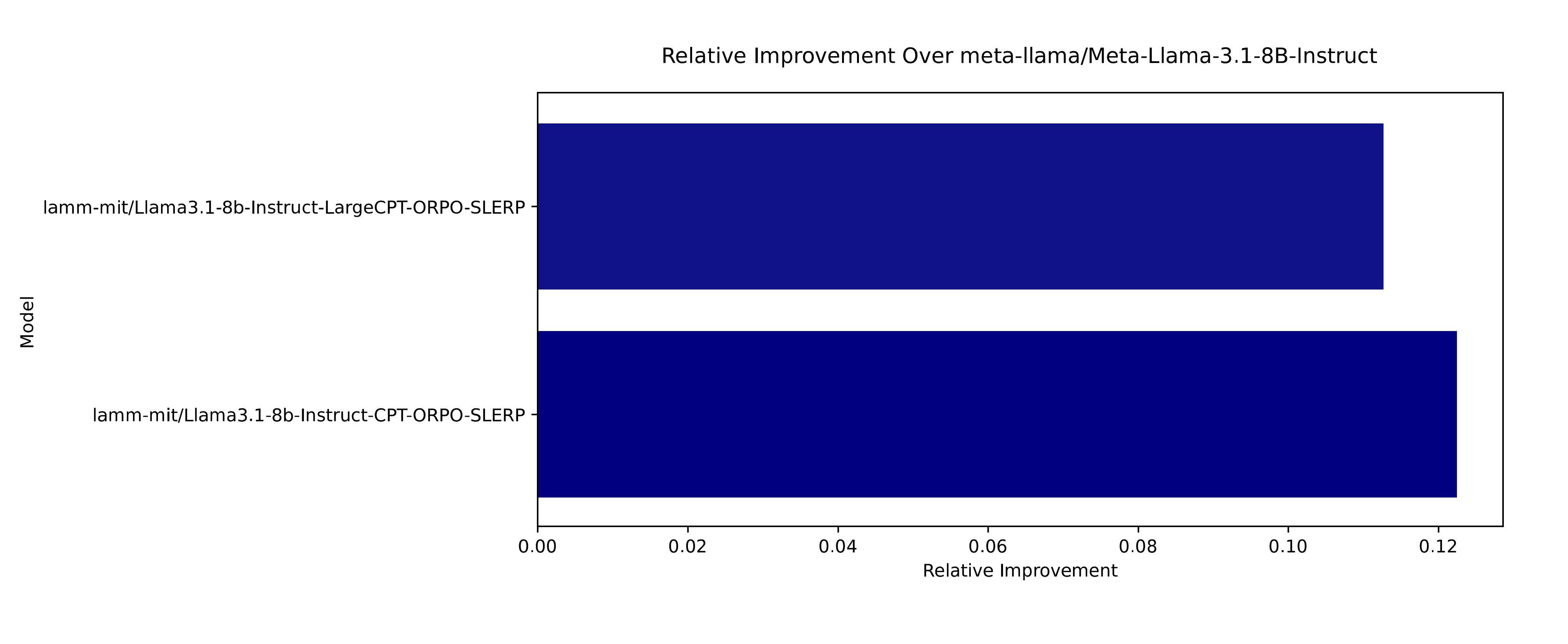}
    \caption{Effect of using a larger CPT dataset using the extended dataset, along with the \texttt{lamm-mit/magpie-ultra-v0.1} dataset, including a more varied quality with a higher content including more defected text. As can be seen performance decreases, underscoring the effect of higher quality, clean data for positive training outcomes. Future experiments could be done to discern these effects more clearly, especially focusing on the effect of various training data compositions. }
    \label{fig:Fig_24}
\end{figure}

Figure~\ref{fig:Fig_24} shows the effect of using a larger CPT dataset using the extended dataset of 8,000 papers, but with more varied format including more defected text, when training the Llama series. As can be seen, performance decreases, underscoring the effect of higher quality, clean data for positive training outcomes. As mentioned earlier, the extended dataset was constructed using a mix of PDF2Text and Nougat OCR~\cite{Blecher2023Nougat:Documents}; we found these methods to yield more variable text quality. While, for instance, Nougat can successfully render equations in Markup format, it also leads to a relatively frequent occurrence of unknown symbols, page breaks, repeated characters, and other defects. These methods did not cause issues when the data was further processed into question-answer pairs or summaries, rational explanations, and so on, but there is an apparent negative effect on CPT as the data is provided in raw format. 

Likewise, a similar test case with the extended dataset was conducted with the Mistral series of models. The variant trained on the original integrated dataset achieved the best overall benchmark of 0.81, whereas the variant trained on the extended dataset achieved 0.80. These results suggest that future experiments could be conducted to assess the effect of this particular dataset variation on that model architecture's performance. We leave this to future work; noting that the overall effects of model merging and the use of Base vs. Instruct models as basis are stable, with differences, however, in which exact strategy yields the best results: For the Llama and Mistral models, it was Instruct-CPT-SFT-ORPO-SLERP and Instruct-CPT-ORPO-SLERP. These observations are further complicated by the effect of prompting, which can skew results one way or the other. An overarching theme, however, is that consistently, SLERP merging yields super performance. For a straightforward and computationally effective way to implement a fine-tuning strategy, the procedure Instruct-CPT-SLERP is probably the best overall choice. While it does not yield the best performance for all scenarios, it generally yields strong performance. The differences show that nuanced benchmarking and prompt engineering can be critical. 
 
\subsection{Mechanistic analysis to elucidate key steps with highest impact on performance}

\begin{figure}
    \centering
    \includegraphics[width=1\linewidth]{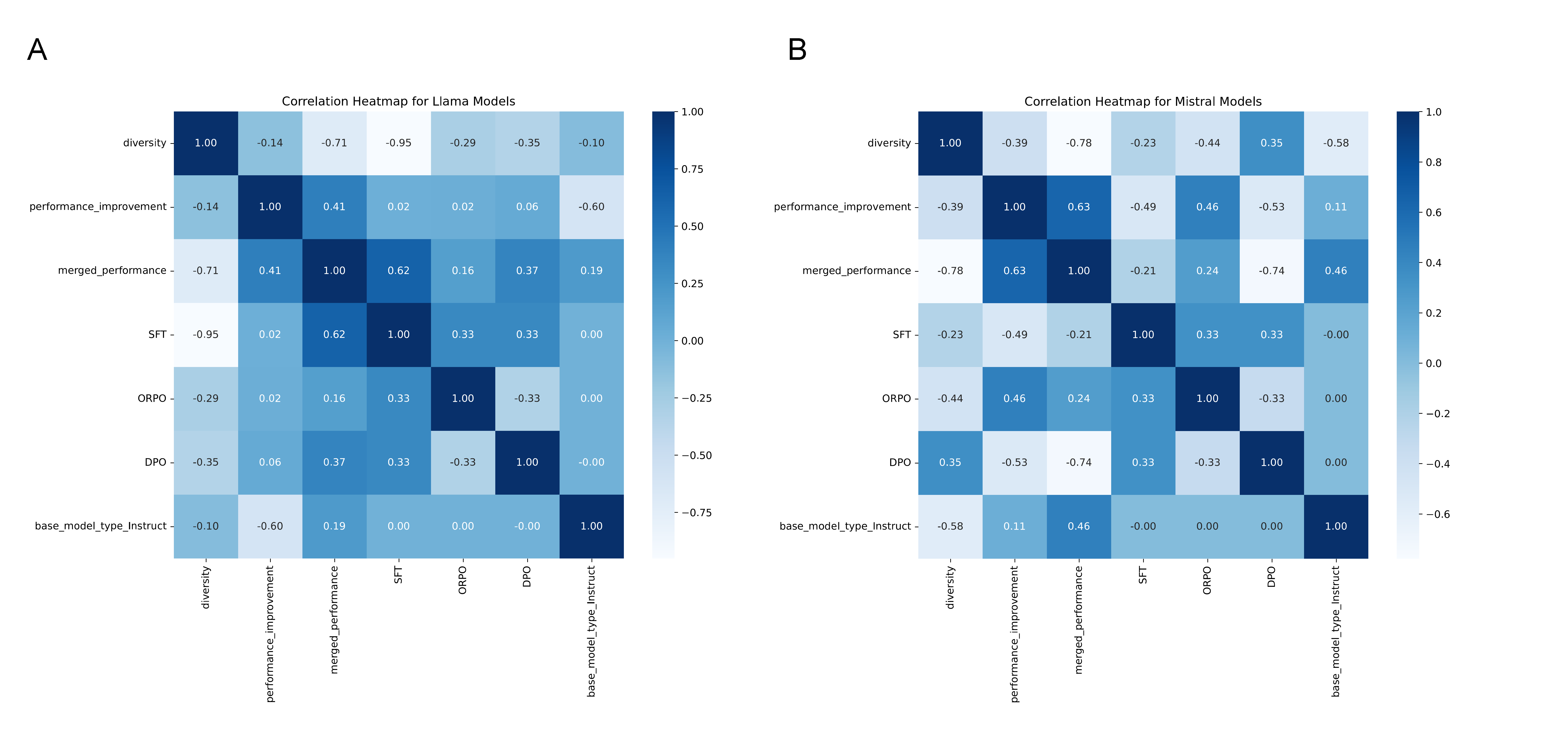}
    \caption{Visualization of correlation heatmaps to assess the relationships between various model attributes and the performance of merged models, where the performance of the merged model is the primary outcome of interest. The attributes considered include the diversity between parent models, the performance improvement relative to the parent models, SFT, DPO/ORPO, and whether the model is based on the Base or Instruct architecture. The correlation coefficients range from -1 to 1, with positive values indicating a direct relationship between the attribute and the merged model performance, and negative values indicating an inverse relationship.
    Panel A, Llama Models: SFT shows the strongest positive correlation with merged performance, suggesting that models incorporating SFT tend to achieve better results after merging. Conversely, diversity between parent models has a strong negative correlation with merged performance, implying that greater differences between parent models are associated with lower merged performance. ORPO and the Instruct architecture exhibit moderate positive correlations with merged performance, indicating that these factors also contribute positively, though less significantly than SFT.
    Panel B, Mistral Models: In the Mistral models, performance improvement shows a robust positive correlation with merged performance, particularly in instruction-tuned models, which also show a strong positive correlation between the base model type and merged performance. Diversity, however, exhibits a negative correlation with merged performance, similar to the Llama models, though the effect is less pronounced. ORPO demonstrates a moderate positive correlation with performance improvement, suggesting that this optimization method contributes to enhanced performance in Mistral models, albeit not as strongly as SFT in Llama models. The findings suggest that instruction-tuned base models and merging strategies play a crucial role in optimizing Mistral model performance, while diversity and SFT influence these outcomes differently in Llama models. }
    \label{fig:Fig_10}
\end{figure}

As a next step in the analysis we focus on correlation heatmaps to illustrate the relationships between various model attributes and the performance of merged models. As shown in Figure~\ref{fig:Fig_10}, the performance of a merged model is denoted as \( P_{\text{merged}} \), while the performance of the two parent models is denoted as \( P_1 \) and \( P_2 \). Performance improvement is defined as the difference between the performance of the merged model and the maximum performance of the two parent models:
\[
\text{Performance Improvement} = P_{\text{merged}} - \max(P_1, P_2)
\]
Diversity between parent models is measured as the absolute difference between their individual performances:

\[
\text{Diversity} = |P_1 - P_2|
\]

To elucidated overall trends that can be gleaned from the results, Figure~\ref{fig:Fig_10} depicts correlation heatmaps for the fine-tuned Llama and Mistral models. The data reveals distinct relationships between various metrics. In Llama models, a strong negative correlation between diversity and SFT suggests that higher diversity reduces reliance on supervised fine-tuning, whereas performance improvement shows moderate positive correlations with both merged performance and SFT, indicating that these factors contribute to improved outcomes. In contrast, Mistral models exhibit a more robust positive correlation between performance improvement and merged performance, especially in instruction-tuned models, where the Base model type significantly enhances merged performance. ORPO, while contributing to performance improvements in both models, has a more pronounced impact in Mistral models. Overall, the findings suggest that diversity tends to reduce SFT dependency, particularly in Llama models, while instruction-tuned Base models in Mistral benefit more from merging strategies, emphasizing the importance of model selection and optimization methods.

\begin{figure} 
    \centering
    \includegraphics[width=.75\linewidth]{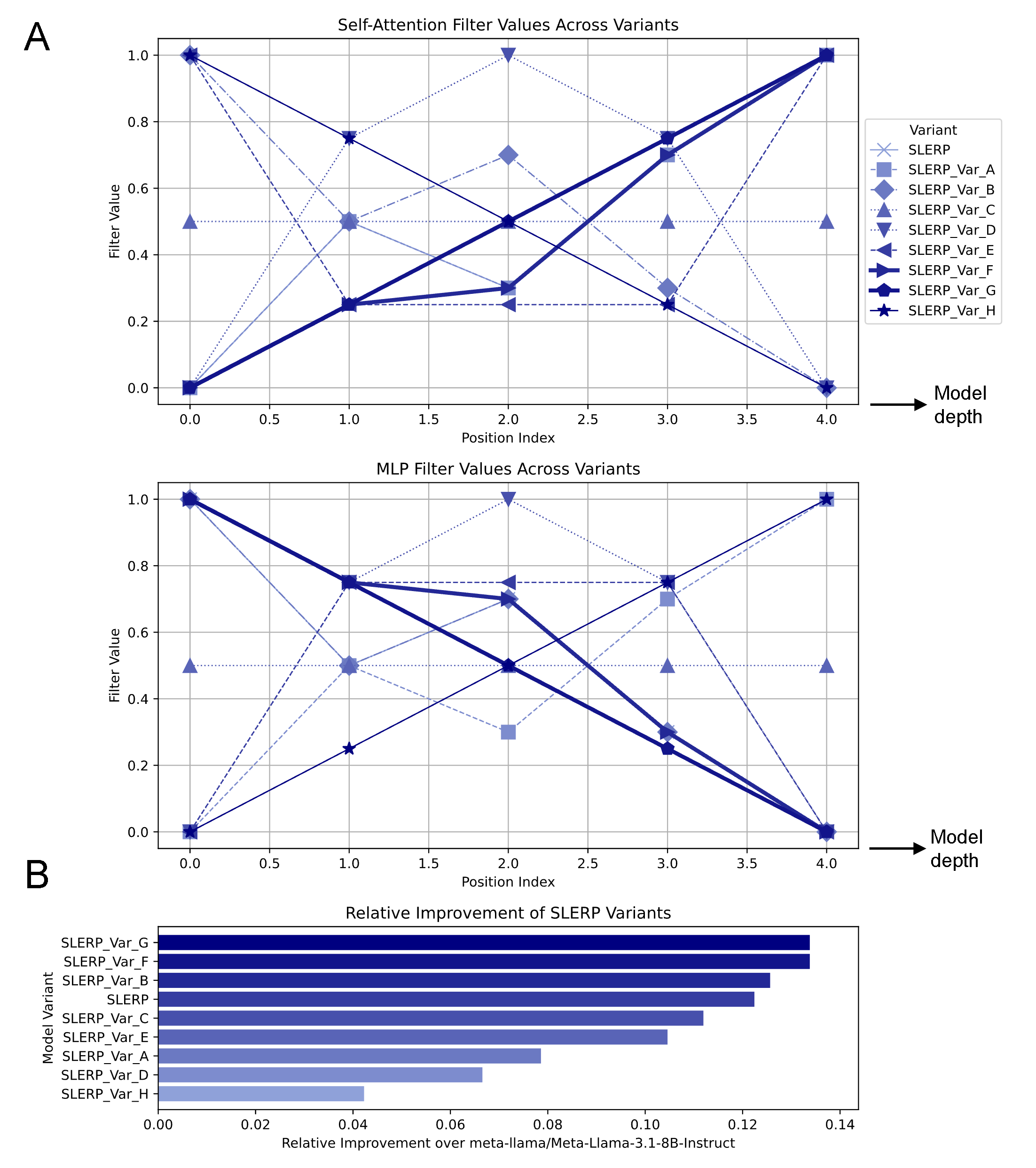}
    \caption{Model merging incorporates varying the relative density of how parameters are combined across the layers of the LLMs involved (see, in Figure~\ref{fig:Fig_11} for the two points merged at 30\% vs. 70\%). The analysis shown here focuses on variants of the original SLERP merge used in the earlier examples, for the best performing strategy in the case of the \texttt{Llama-3.1} Model variants(CPT-SFT-ORPO, then merge). As depicted visually, we vary the self-attention filter values distinctly from the multilayer perceptron (MLP) values (panel A). Different weighting schemes are employed, starting from the reference case that was chosen based on earlier work~\cite{goddard2024arceesmergekittoolkitmerging}. Resulting performance measures are summarized in panel B, showing that Variants G and F show the best performance (where G is a simple linear progression).}
    \label{fig:Fig_4}
\end{figure}

In model merging, there are several parameter choices, including the relative density of the parameters that are preserved across the layers of the LLMs being merged. This is exemplified in Figure~\ref{fig:Fig_11} for the two points merged at 30\% vs. 70\% along their SLERP paths. In Figure~\ref{fig:Fig_4} we conduct a systematic analysis on variants of the original SLERP merge used in the earlier examples, with a range of alternative options, for the best performing strategy in the case of the \texttt{Llama-3.1} Model variants (CPT-SFT-ORPO-SLERP). As depicted visually, we vary the self-attention filter values distinctly from the multilayer perceptron (MLP) values (Figure~\ref{fig:Fig_4}A). Different weighting schemes are employed, starting from the reference case that was chosen based on earlier work~\cite{goddard2024arceesmergekittoolkitmerging}. Resulting performance measures are summarized in Figure~\ref{fig:Fig_4}B, showing that Variants G and F show the best performance (where G is a simple linear progression across the depth of the LLM) (detailed performance assessments for other variants are shown in Figure~\ref{fig:Fig_4}).

\subsection{Contrasting assessments with very small LLMs}
While the models studied earlier were modest in size, around 7-8 billion parameters, recent research has resulted in even smaller, yet useful, models that can be particularly useful for edge computing applications, or deployment on devices such as mobile phones or robotic systems. We now examine whether such models also show the marked effects observed earlier due to model merging. We conduct this analysis using the \texttt{SmolLM} model series, specifically the 1.7 billion parameter model. This choice is partially motivated by the complete open access of the model, training strategy, and training data. As in the earlier analyses, we start with the base model and successively apply CPT, SFT and DPO (we found that for this small model, DPO worked better than ORPO). Though never reaching the level of absolute performance of finetuned 7B or 8B models, in almost all fine-tuning cases with SmolLM, we find the most significant performance increases relative to its original model with the CPT-SFT-DPO version of SmolLM being the top performing variant.

\begin{figure} 
    \centering
    \includegraphics[width=.8\linewidth]{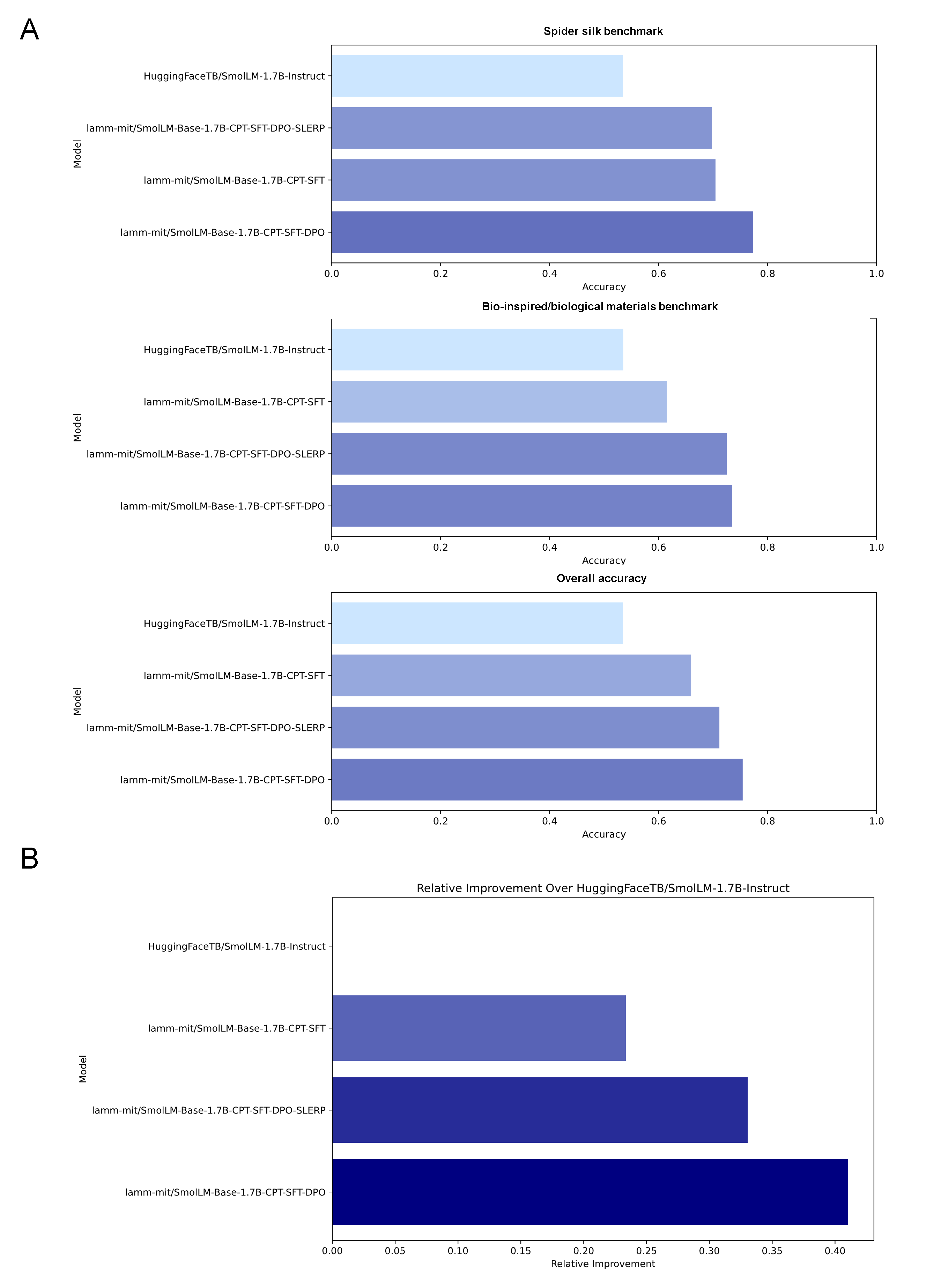}
    \caption{Comparative performance analysis of SmolLM-1.7B models across benchmarks. Panel A: Accuracy results for various \textbf{SmolLM-1.7B} model variants on the Spider Silk, Bio-inspired/Biological Materials, and Overall Accuracy benchmarks. The \texttt{HuggingFaceTB/SmolLM-1.7B-Instruct} model serves as the baseline, albeit training is done solely on the Base model (\texttt{HuggingFaceTB/SmolLM-1.7B}). While applying CPT and SFT strategies improves performance, the addition of DPO yields further accuracy gains. However, unlike in the much larger Llama or Mistral models, here, SLERP merging does not yield the best performance overall. This is more clearly shown in panel B, where we plot the relative improvement over the baseline model (\texttt{HuggingFaceTB/SmolLM-1.7B-Instruct}) across the benchmarks. Notably SLERP combined with DPO yields a slight reduction in performance over the CPT-SFT-DPO case, in stark contrast to the earlier results for Llama and Mistral. The emergent behaviors triggered by SLERP in larger models are not observed here, indicating a potential threshold effect. This suggests that SLERP's ability to unlock novel capabilities may depend on the scale of the model, with smaller models like the 1.7B parameter SmolLM failing to exhibit these emergent properties. These findings underscore the critical role of model scale in realizing advanced capabilities, contributing to the broader understanding of scaling laws in neural networks. Note, in B, the top bar is zero since this model is used as the reference to compute model improvement. It is kept in the visualization for consistency with the earlier plots (where some variants yielded performance degradation, that is, negative values).
}
    \label{fig:Fig_20}
\end{figure}

As depicted in Figure~\ref{fig:Fig_20} while we observe a significant emergence of new capabilities when applying SLERP to large-scale language models in the 7B and 8B parameter ranges, these emergent behaviors were absent in smaller models, such as those with 1.7B parameters. This may suggest a threshold effect where SLERP's potential to unlock novel abilities is contingent on model size. Smaller models might lack the same level of complexity as larger 7B to 8B models that have notably richer high-dimensional parameter spaces and capabilities especially for reasoning and knowledge recall. These findings underscore the importance of model scale in the manifestation of emergent properties and provide critical insights into the interplay between interpolation techniques and model complexity. Our results contribute to the broader understanding of scaling laws in neural networks, highlighting the conditions under which advanced capabilities may be realized.  A summary of the observed performance over the expected, averaged performance of the base model is shown in Figure~\ref{fig:Fig_5}C.

\subsection{Further quantification of the effects of model merging across all model architectures}
\begin{figure}
    \centering
    \includegraphics[width=1\linewidth]{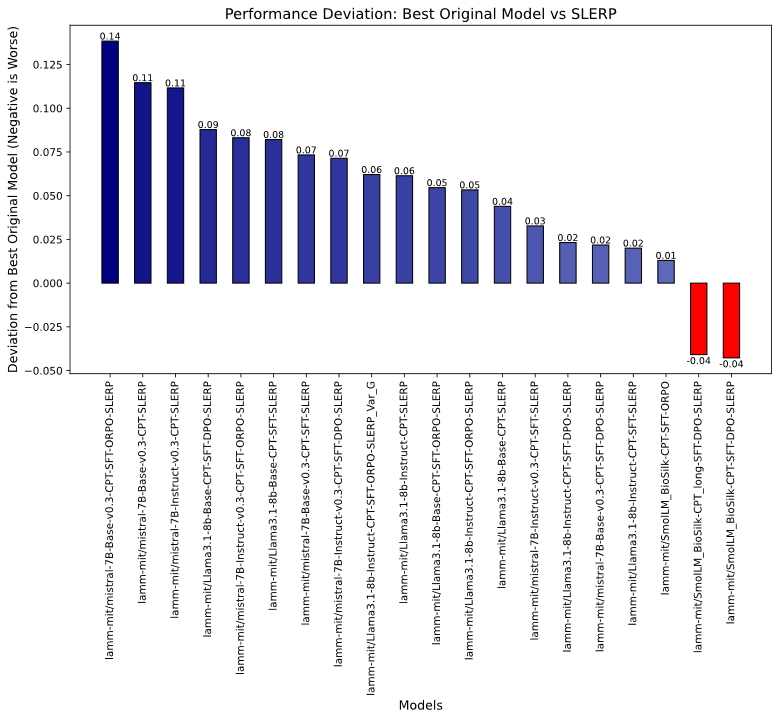}
    \caption{Performance deviation of SLERP merged models compared to the best original model. The plot illustrates the deviation in performance between models merged using SLERP and their best-performing original counterparts (either without SLERP or with instruction tuning). The deviation is calculated as the difference between the best original model's performance and the SLERP model's performance. Negative deviations, where SLERP underperforms relative to the best original model, are highlighted in red. Positive deviations, indicating better performance of the SLERP model, are shown in shades of blue, with darker blue representing greater improvements. Models are ordered from most significant improvement to most significant underperformance.}

    \label{fig:Fig_109}
\end{figure}

To better understand whether or not and to what degree model merging improves performance over either one of the two models used for merging, we present the analysis shown in Figure~\ref{fig:Fig_109}. The plot shows performance deviation of SLERP merged models compared to the best original model used as source. 

The results reveal that the deviation in performance between models merged using SLERP and their best-performing original counterparts, whereby the deviation is calculated as the difference between the best original model's performance and the SLERP model's performance. Hence, negative deviations, where SLERP underperforms relative to the best original model, are marked in red. Positive deviations, indicating better performance of the SLERP model, are shown in shades of blue, with darker blue representing greater improvements.  

\subsection{Interactive examples for open-ended cross-material reasoning and material design tasks}

In our next experiment, we conduct interactive conversations with a set of the models, using consistent system prompts and identical user input. We aim to test multi-turn capabilities of the models, assess responsiveness to system prompts and instructions, and capability to produce structured output (JSON). We will further assess the quality of synthesis of each model, along a set of criteria that include depth of reasoning, creativity, clarity and whether or not quantitative predictions are featured. 
Each of the conversations unfolds as follows:
\begin{LLMbox}{Conversation with: ...}
\begin{lstlisting}
(*@\bf\hlblue{System:}@*) Your are a helpful and friendly assistant. You are an expert in materials science with special knowledge in biological materials. You provide concise but accurate responses. You are an expert in combining different concepts, often not obviously related, and provide accurate accounts of correct relationships or analogies. 

(*@\bf\hlred{User:}@*) Collagen and leaves, discuss their relationship.

(*@\bf\hllightgreen{Assistant:}@*) ...

(*@\bf\hlred{User:}@*) Thank you! Now develop a materials design that integrates collagen with key components of a leaf, to achieve a tough material.

(*@\bf\hllightgreen{Assistant:}@*) ...

(*@\bf\hlred{User:}@*) Summarize these design features in JSON format.

(*@\bf\hllightgreen{Assistant:}@*) ...

\end{lstlisting}
\end{LLMbox}

Text Boxes~\ref{TexBox:lamm-mit/Llama3.1-8b-Instruct-CPT-SFT-DPO}, \ref{TexBox:lamm-mit/Llama3.1-8b-Instruct-CPT-SFT-ORPO-SLERP-Var_G}, \ref{TexBox:lamm-mit/mistral-7B-v0.3-Base-CPT-SFT-DPO}, \ref{TexBox:lamm-mit/mistral-7B-v0.3-Base-CPT-SFT-SLERP}, and \ref{TexBox:lamm-mit/SmolLM-Base-1.7B-CPT-SFT-DPO} present the results of conversations of a human user with a selection of five models (best performing models, and DPO trained models).

All five models demonstrate a strong ability to connect seemingly unrelated concepts, such as collagen and leaves, and to propose innovative materials designs. The output features rich Markup formatting (note, the raw source is shown here). The models show a consistent understanding of the biological and materials science concepts involved, suggesting that their fine-tuning on domain-specific content has been effective. The differences in the responses mainly pertain to the depth of explanation, the creativity of the proposed designs, and the clarity of the output in both natural language and structured formats like JSON.

As shown in Text Box~\ref{TexBox:lamm-mit/Llama3.1-8b-Instruct-CPT-SFT-DPO}, the model lamm-mit/Llama3.1-8b-Instruct-CPT-SFT-DPO provides a well-organized and detailed discussion, drawing clear parallels between the structure and function of collagen and leaves. The proposed material design is robust, incorporating key components like a collagen-based matrix, vascular-like channels, and mesophyll-like cells. The response is notable for its comprehensive breakdown of each component's role, leading to a thorough and scientifically grounded design. The JSON summary is precise, reflecting the structure of the proposed design effectively.

\begin{table}[h]
\scriptsize
\centering
\begin{tabular}{|p{2.5cm}|p{2.5cm}|p{5cm}|p{5cm}|}
\hline
\textbf{Model} & \textbf{Key Strengths} & \textbf{Weaknesses} & \textbf{Observations} \\ \hline
\texttt{lamm-mit/Llama3.1-8b-Instruct-CPT-SFT-DPO} & Detailed and comprehensive design; strong scientific grounding & May be overly complex for some applications; focuses heavily on structural components, potentially missing opportunities for simpler solutions & The model excels in providing a thorough and detailed analysis. The robust material design effectively integrates various components, showcasing strong scientific reasoning. However, the complexity of the design may limit its practicality in certain applications. The JSON summary is precise, but the depth of detail could be overwhelming in less technical contexts.  \\ \hline
\texttt{lamm-mit/Llama3.1-8b-Instruct-CPT-SFT-ORPO-SLERP-Var\_G} & Concise and inventive; broader application vision & Lacks depth in certain areas; the brevity may sacrifice some detail in the explanation of the material properties & This model offers a concise yet insightful analysis. Its focus on innovation is evident in the integration of chloroplast-inspired nanoparticles for energy harvesting, which is unique among the models. However, the lack of depth in the explanation of mechanical properties could be a drawback in more technical discussions. The JSON representation is clear and well-structured, reflecting the model’s focus on innovative applications.  \\ \hline
\texttt{lamm-mit/mistral-7B-v0.3-Base-CPT-SFT-DPO} & Comprehensive integration of components; strong mechanical focus & Potentially too focused on mechanical properties, which might limit creativity in design & The model provides a detailed and comprehensive approach, particularly in its focus on the mechanical properties of the proposed material. The integration of multiple materials, such as cellulose nanofibers and silk fibroin, is handled well, though the emphasis on mechanical properties may limit the exploration of other innovative aspects. The JSON summary effectively captures the design’s complexity, though it may be seen as overly technical for broader audiences.   \\ \hline
\texttt{lamm-mit/mistral-7B-v0.3-Base-CPT-SFT-SLERP} & Simple and straightforward; clear and accessible design & Lacks depth and innovation; may be seen as too simplistic compared to other models & This model provides a straightforward and accessible design, focusing on essential components like collagen fibers and cellulose nanofibers. While the simplicity makes it accessible, it also limits the depth and creativity of the design. The JSON summary is basic but clear, making it suitable for less technical audiences but potentially insufficient for more advanced applications.   \\ \hline
\texttt{lamm-mit/SmolLM-Base-1.7B-CPT-SFT-DPO} & Highly creative and detailed; innovative features like self-healing & Complexity might make it challenging to implement; could be seen as too speculative for practical use & The model stands out for its creativity and innovation, particularly in its inclusion of self-healing and shape-memory features. These advanced characteristics make it suitable for high-end applications, though the complexity of the design may pose challenges in practical implementation. The JSON summary is comprehensive, effectively conveying the unique aspects of the design, but the speculative nature of some features may limit its immediate applicability.  \\ \hline
\end{tabular}
\caption{Summary of model performance, strengths, weaknesses, and design strategies based on conversations related to collagen and leaves. Evaluation conducted using \texttt{GPT-4o}.}
\label{tab:ft_strategies_collagen_leaves}
\end{table}

Text Box~\ref{TexBox:lamm-mit/Llama3.1-8b-Instruct-CPT-SFT-ORPO-SLERP-Var_G} shows results for the \texttt{lamm-mit/Llama3.1-8b-Instruct-CPT-SFT-ORPO-SLERP-Var\_G} model. This model delivers a more concise but also insightful analysis of collagen and leaves. The material design focuses on integrating collagen fibrils with cellulose microfibrils and chloroplast-inspired nanoparticles. This resembles an inventive approach to enhancing the material's properties. This model excels in identifying the potential applications of the designed material, showcasing a broader vision for its use. The JSON representation is clear and well-structured, effectively summarizing the design features.

Text Box~\ref{TexBox:lamm-mit/mistral-7B-v0.3-Base-CPT-SFT-DPO} captures results of the \texttt{lamm-mit/mistral-7B-v0.3-Base-CPT-SFT-DPO} model. The responses are found to be comprehensive, with a strong focus on the mechanical properties of collagen and leaves. The proposed material design is detailed, incorporating collagen fibrils, cellulose nanofibers, silk fibroin, and a nanocellulose-based matrix. This model particularly stands out for its emphasis on the integration of these components to enhance the material's toughness and durability. The JSON summary is thorough, capturing the complexity of the design and its potential applications.

Next, Text Box~\ref{TexBox:lamm-mit/mistral-7B-v0.3-Base-CPT-SFT-SLERP} shows results for the \texttt{lamm-mit/mistral-7B-v0.3-Base-CPT-SFT-SLERP} model. This result illustrates a more straightforward and less detailed analysis compared to the others. While the connection between collagen and leaves is adequately explained, the material design is simpler, focusing on collagen fibers, cellulose nanofibers, chlorophyll, and pectin. This response is notable for its clarity and simplicity, making it accessible but perhaps lacking the depth seen in other models. The JSON summary is basic but effective in conveying the key elements of the design.

Finally, Text Box~\ref{TexBox:lamm-mit/SmolLM-Base-1.7B-CPT-SFT-DPO} showcases the results of the smallest model in this study, \texttt{lamm-mit/SmolLM-Base-1.7B-CPT-SFT-DPO}. The model offers an inventive, creative, and highly detailed response, integrating a broad range of components such as collagen fibrils, nanocrystalline cellulose, and an alginate adhesive, resulting in a material creatively referred to as ``Leafy-Coraline (LC) Composite''. This model excels in proposing a novel composite material with self-healing and shape-memory properties, reflecting a relatively high level of creativity and technical understanding. The JSON summary is comprehensive, capturing the innovative aspects of the design and its potential applications effectively. In spite of its size, this model provides excellent responses. 

A summary of key observations is shown in Table~\ref{tab:ft_strategies_collagen_leaves}. We can see that each model demonstrates strengths in different areas, from detailed explanations and innovative designs to clear and concise JSON summaries. The variations in depth, creativity, and technical detail among the models highlight the diversity of approaches and the potential for each to be suited to different types of tasks or applications. Overall, these models provide a strong foundation for further exploration and development in the field of bio-inspired materials design. 
A detailed breakdown of the analysis, generated using \texttt{GPT-4o} by analyzing the raw text of the conversations, is shown in Table~\ref{tab:intelligence_scores}, showing the results of an analysis conducted by sharing the raw text with  \texttt{GPT-4o} and prompting the model to identify criteria and assess the different versions. 

\begin{table}[h]
\centering
\scriptsize
\begin{tabular}{|p{3cm}|p{1.5cm}|p{1.5cm}|p{1.5cm}|p{1.5cm}|p{1.5cm}|p{1.5cm}|p{1.5cm}|}
\hline
\textbf{Model} & \textbf{Depth of Reasoning} & \textbf{Creativity} & \textbf{Clarity} & \textbf{Quantitative Predictions} & \textbf{Total Score (Max 40)} & \textbf{Average Score (Max 10)} & \textbf{Normalized Intelligence Score} \\ \hline
\texttt{lamm-mit/SmolLM-Base-1.7B-CPT-SFT-DPO} & 10 & 10 & 9 & 9 & 38 & 9.5 & 10.0  \\ \hline
\texttt{lamm-mit/Llama3.1-8b-Instruct-CPT-SFT-DPO} & 9 & 8 & 9 & 7 & 33 & 8.25 & 8.7  \\ \hline
\texttt{lamm-mit/Llama3.1-8b-Instruct-CPT-SFT-ORPO-SLERP-Var\_G} & 8 & 9 & 8 & 6 & 31  & 7.75 & 8.2  \\ \hline
\texttt{lamm-mit/mistral-7B-v0.3-Base-CPT-SFT-DPO} & 8 & 7 & 8 & 7 & 30 & 7.5 & 7.9  \\ \hline
\texttt{lamm-mit/mistral-7B-v0.3-Base-CPT-SFT-SLERP} & 6 & 6 & 7 & 5 & 24 & 6.0 & 6.3  \\ \hline
\end{tabular}
\caption{Summary of model performance with individual criterion scores, total intelligence score, average score, and normalized intelligence score. We assign scores from 1 to 10 for each criterion, including (1) Depth of Reasoning: How well the model explains concepts and connects ideas, (2) Creativity: The uniqueness and innovation in the material design. (3) Clarity: How clearly the model communicates its ideas. (4) Quantitative Predictions: Whether the model includes numerical or quantitative aspects in its response. Evaluation conducted using \texttt{GPT-4o}.}
\label{tab:intelligence_scores}
\end{table}



\subsection{Agentic use in image generation: Applications in cross-domain knowledge integration for materials and urban design}

We show several examples to highlight the capabilities and potential of fine-tuned LLMs, showcasing a particular application in materials design. The overall goal is to explore how the LLMs developed here can be used to reason over complex materials principles and use the insights developed through multi-step prompting to create a prompt for image generation. We note that our fine-tuned SmolLM based model performs extremely well for this task and yields quite creative prompts that integrate various ideas and concepts (underscoring its potential as a creative agent). 

While slight variations of prompting are used to yield different examples (each result presented here includes a detailed presentation of all features), the general goal is to think about design principles that we can extract from combining different biological materials. For instance, we ask the model to think about ways to combine design elements from spider silk and collagen to make a strong, lightweight but tough material, and to also incorporate design cues from leaf microstructures. The approach can be used to focus directly on material microstructures but can also be used to yield cross-domain results, such as architectural ideas or city design. 

In our first example we prompt the \texttt{lamm-mit/SmolLM-Base-1.7B-CPT-SFT-DPO} model as follows (Text Box~\ref{TexBox:lamm-mit/SmolLM-Base-1.7B-CPT-SFT-DPO_ImageGen_1} shows the entire conversation):

\begin{LLMbox}{Conversation template used with: \texttt{lamm-mit/SmolLM-Base-1.7B-CPT-SFT-DPO} }
\begin{lstlisting}
(*@\bf\hlblue{System:}@*) Your are a helpful and friendly assistant. You excel in combining disparate concepts and develop well thought-out designs of organic, bio-inspired material microstructures. 
          
(*@\bf\hlred{User:}@*) First, think about specific design principles you can extract from combining dragline silk and collagen fibers to make a strong, lightweight but tough material. Incorporate design cues from leaf microstructure.

(*@\bf\hllightgreen{Assistant:}@*) ...

(*@\bf\hlred{User:}@*) Use these thoughts to develop an image generation prompt for an AI system like Stable Diffusion. Instead of using technical terms, you carefully describe the image using geometric features such as lines, shapes, circles, holes, porous features, patterns, and so on. 

(*@\bf\hllightgreen{Assistant:}@*) ...

(*@\bf\hlred{User:}@*)  Give me one short prompt that I can use in my AI generator, but be specific about the features to be incorporated and how.

(*@\bf\hllightgreen{Assistant:}@*) Here's a concise prompt to get you started:

"Generate a futuristic, eco-friendly architectural concept utilizing a biomimetic composite material that integrates the structural efficiency of spider silk with the adaptive porosity of plant tissues. Utilize the following key features:

* Fibrous architecture inspired by spider silk, represented by sinuous lines and curved forms.
* Interconnected, spherical nodes reminiscent of plant cell walls, emphasizing growth and adaptation.
* Open cellular structures echoing the permeable nature of plant leaves, suggesting dynamic exchanges and self-regulation capabilities.
* Gradations of opacity and transparency inspired by the varying densities found in plant tissues, highlighting functional differentiation and multi-functionality."

This prompt encourages the AI generator to incorporate specific visual elements drawn from nature, ensuring a cohesive and engaging final output.

\end{lstlisting}
\end{LLMbox}

\begin{figure}
    \centering
    \includegraphics[width=1\linewidth]{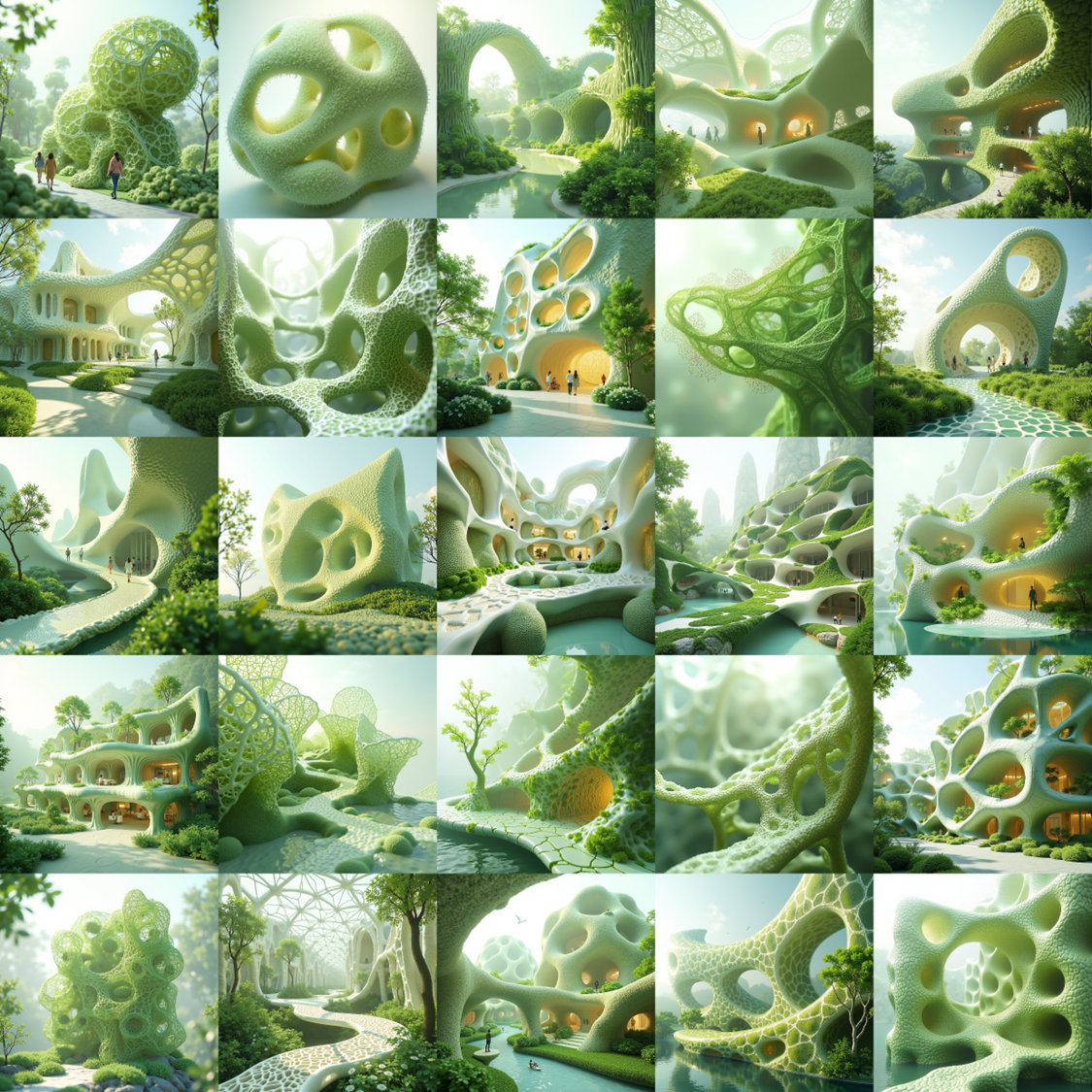}
    \caption{Image generation results developed by \texttt{lamm-mit/SmolLM-Base-1.7B-CPT-SFT-DPO}, leading to this prompt used for image generation: \texttt{Generate a futuristic, eco-friendly architectural concept utilizing a biomimetic composite material that integrates the structural efficiency of spider silk with the adaptive porosity of plant tissues. Utilize the following key features: * Fibrous architecture inspired by spider silk, represented by sinuous lines and curved forms. * Interconnected, spherical nodes reminiscent of plant cell walls, emphasizing growth and adaptation. * Open cellular structures echoing the permeable nature of plant leaves, suggesting dynamic exchanges and self-regulation capabilities. * Gradations of opacity and transparency inspired by the varying densities found in plant tissues, highlighting functional differentiation and multi-functionality.} The images illustrate a architectural vision that draws inspiration from biomorphic forms including leaf microstructures, integrating sustainable design principles with natural elements, creating open, flowing spaces that emphasize harmony between advanced architectural techniques and the organic world.
}

    \label{fig:Fig_2000}
\end{figure}

A collection of images produced in this way is shown in Figure~\ref{fig:Fig_2000}. The resulting images resemble a visionary architectural concept where biomorphic structures seamlessly blend with nature, creating a futuristic and sustainable environment. The designs are inspired by natural forms such as honeycombs, coral, and cellular structures, characterized by fluid, curving lines and intricate lattice-like frameworks that evoke the organic world. These structures are integrated with greenery, emphasizing harmony with the environment and suggesting the use of innovative, eco-friendly materials. The open, flowing spaces with large archways and natural light emphasize a connection with the outdoors, creating a sense of tranquility and well-being. As can be seen in some of the images, human figures within these spaces highlight the livability and community-centric design, suggesting a vision where technology and nature coexist harmoniously. The images present a forward-looking approach to architecture, where sustainability, aesthetics, and advanced techniques converge to create a new paradigm for living and public spaces. A close inspection of the resulting designs further suggest a clear emergence of the leaf microstructure patterns, a result of the prompt and the fine-tuned ability of the generative model to incorporate this particular design idea. 

These design ideas have the potential to be implemented, with specific functionality emphasized. Real-life examples of bioinspired architectural designs include “The Hive”, a building on the NUT campus designed by Heatherwick Studio. Its honeycomb structure mirrors the cellular organization of hives, providing a modified modular hexagonal form that improves structural efficiency, particularly by optimizing airflow and ventilation systems. Another compelling example is “Little Island” in New York, which is designed to resemble floating leaves and create a dynamic urban landscape. The design shows resilience to climate change, offering flexibility in adapting to changes in water levels.

As illustrated in these examples, bioinspired designs in architecture go beyond mere aesthetics by incorporating natural elements into urban facilities, often enhancing functionality. Specifically, as shown in Figure~\ref{fig:Fig_2000}, these designs have the potential to improve sustainability and energy efficiency by optimizing material usage and structural topology compared to conventional designs. For instance, the designs in row three reflect design simplicity and potential material saving incorporating non-uniform cellular structures tailored for different space usages. The structural design also satisfies the load path analysis, with larger column sizes at lower floors compared to the top ones, which also shows the load-bearing capacity enhancement. Moreover, energy efficiency can be improved by optimizing heating and cooling systems, as seen in designs that mimic leaf veins, such as the first design in row one. The second design in row five demonstrates a bio-inspired roof design, with beam thickness inspired by leaf veins to maximize lighting and energy efficiency. Additionally, these designs enhance the user experience by integrating natural elements into urban facilities, such as bridges and pathways within the designs. However, to actualize these designs in real-life applications, further research is needed, which includes studying wind effects and façade analysis considering the integration of extended plants and ensuring the loading and structural integrity, as well as selecting materials that are compatible with greenery.

Furthermore, bioinspired structures featuring cellular patterns have the potential to be further enhanced with topology optimization for various purposes, such as optimizing material usage while maintaining structural integrity and architectural features. We note that these bioinspired designs reveal innovative design approaches for environmental integration and sustainability improvements, offering a promising start for exploring the vast design space, although more research is needed to fully validate the application in the real world.

In our second example we prompt the \texttt{lamm-mit/mistral-7B-v0.3-Base-CPT-SFT-DPO} model as follows (Text Box~\ref{TexBox:lamm-mit/mistral-7B-v0.3-Base-CPT_SFT_DPO_ImageGen_2} shows the entire conversation):

\begin{LLMbox}{Conversation template used with: \texttt{lamm-mit/mistral-7B-v0.3-Base-CPT-SFT-DPO}}
\begin{lstlisting}

(*@\bf\hlblue{System:}@*) Your are a helpful and friendly assistant. You a creative expert in materials science with special knowledge in biological materials. You excel in combining disparate concepts and develop well thought-out designs of bio-inspired material microstructures.

(*@\bf\hlred{User:}@*) First, think about design principles you can extract from combining spider silk and collagen to make a strong, lightweight but tough material. Incorporate design cues from leaf microstructure. 

(*@\bf\hllightgreen{Assistant:}@*)  ...

(*@\bf\hlred{User:}@*) Use these thoughts to develop an image generation prompt, incorporating the keyword 'leaf microstructure' when appropriate. Instead of using technical terms, you carefully describe the image using generic features such as lines, shapes, circles, holes, patterns, and so on. 

(*@\bf\hllightgreen{Assistant:}@*) ...

(*@\bf\hlred{User:}@*)  Give me one short prompt that I can use in my AI generator.

(*@\bf\hllightgreen{Assistant:}@*) Specifically for a direct and concise image generation prompt related to the bio-inspired material microstructure, here's a one-liner:

"Generate an image of a golden spider web network intertwined with collagen veins, forming a dynamic, leaf-inspired microstructure amidst a lush green background."
\end{lstlisting}
\end{LLMbox}

\begin{figure}
    \centering
    \includegraphics[width=1\linewidth]{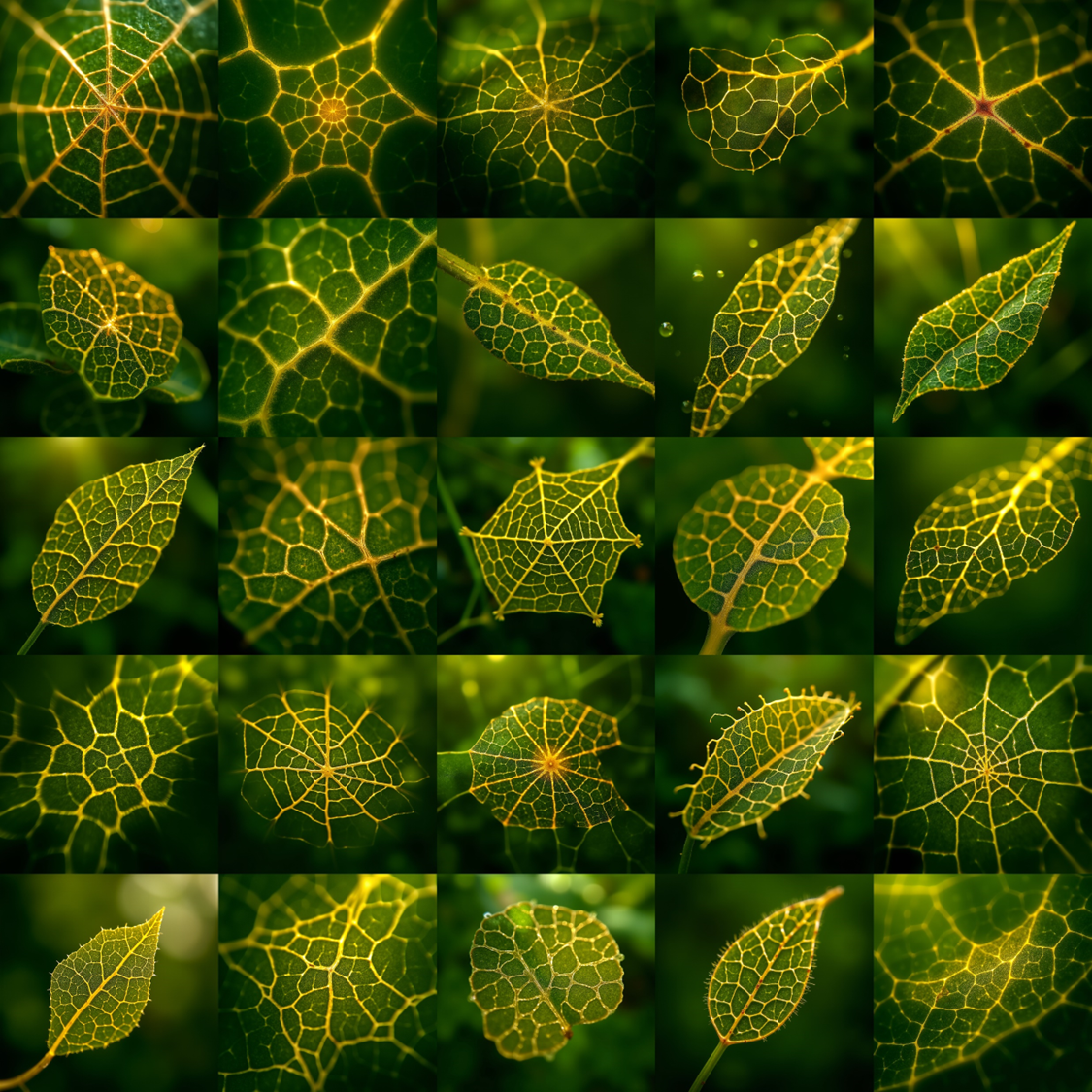}
    \caption{Image generation results developed by \texttt{lamm-mit/mistral-7B-v0.3-Base-CPT-SFT-DPO}, leading to this prompt used for generation: \texttt{Generate an image of a golden spider web network intertwined with collagen veins, forming a dynamic, leaf-inspired microstructure amidst a lush green background.} The images show a novel design of leaf-like structures with prominent vein structures, some of which exhibit intricate spider web-like patterns, all set on a green background. }

    \label{fig:Fig_2001}
\end{figure}

A collection of images produced in this way is shown in Figure~\ref{fig:Fig_2001}.  The images showcase close-up views of a novel form of a biological material, highlighting their intricate vein patterns and microstructures. The leaves exhibit a variety of geometric vein arrangements, ranging from polygonal to radial patterns, all sharply contrasted against the green leaf surfaces by bright yellow to gold veins. The diversity in leaf shapes and vein structures offers a variety of structural options, including a prominent featuring of spider-web like motifs. Several of the leaf patterns in the image resemble spider webs or spider web-like structures. The veins in these particular leaves form radial patterns that converge toward a central point, much like the structure of a 2D orb spider web. This similarity is especially pronounced in some of the leaves where the vein network is more symmetrical and evenly spaced, creating a web-like appearance. These spider web-like patterns, some of which resemble projections of various 3D webs such as cobwebs and sheet webs, add an interesting visual for studies related to natural design and biomimicry. A close inspection of the resulting images shows that textures and depth are captured in fine detail. We find that the soft yet focused lighting accentuates these patterns.

\begin{figure}
    \centering
    \includegraphics[width=1\linewidth]{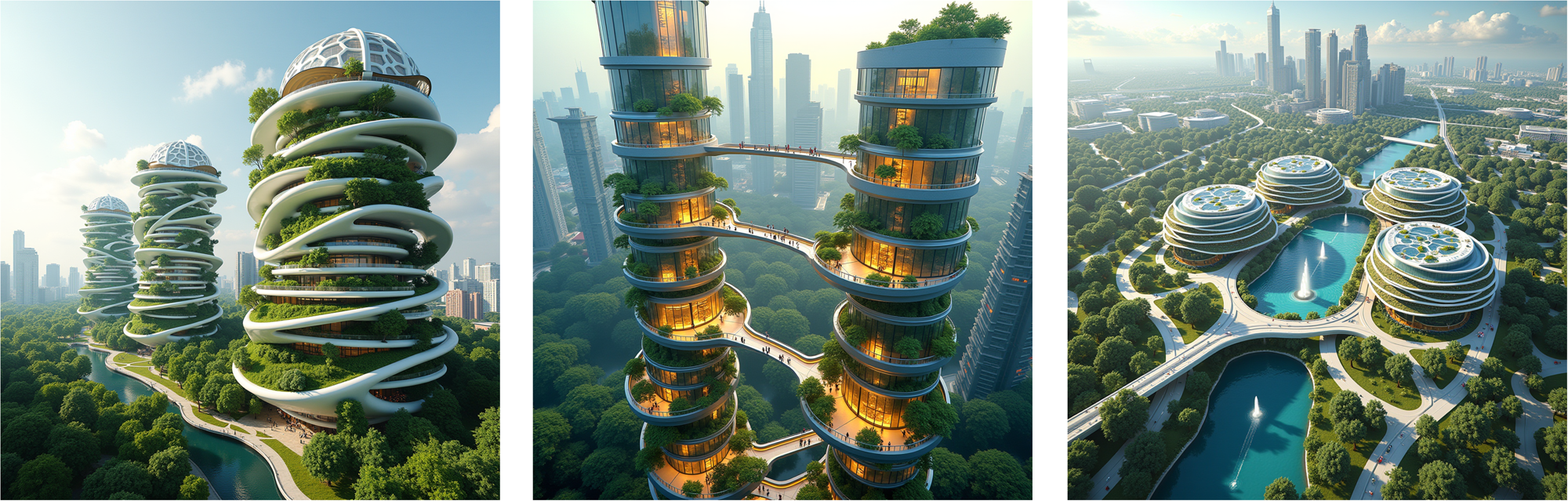}
    \caption{Sample images specifically prompting the model to develop urban design ideas based on a set of biological materials, specifically spider silk, collagen and leaves, developed using  \texttt{lamm-mit/SmolLM-Base-1.7B-CPT-SFT-DPO}. The images illustrate conceptual urban designs integrating biomimetic architecture and ecological sustainability. The left image depicts spiraling vertical towers with embedded greenery, evoking natural patterns (generation prompt: \texttt{Utilize the spiral geometry of a nautilus shell to construct a series of interconnected, curved towers that form the city's skyline. The towers' spiraling walls house lush greenery, generating a perpetual canopy of foliage that filters sunlight and provides shade. Atop each tower stands a sleek, aerodynamic dome housing a state-of-the-art research facility, promoting collaboration among scientists and innovators across disciplines.}). %
    The center image showcases interconnected cylindrical structures through elevated walkways, enhancing both social and ecological connectivity (generation prompt: \texttt{Imagine a cityscape where towering skyscrapers twist and curve along the dual axis grid, their rooftops adorned with lush greenery and shimmering solar panels. At the heart of each tower lies a vibrant, neon-lit hub - a Circular Economy Center - housing cutting-edge research and innovation in sustainable technologies. Connected by elevated walkways and hoverbikes, this city fosters a thriving ecosystem of collaboration and progress.}). 
    The right image presents a broader urban layout, where domed buildings and landscaped water features are seamlessly integrated, reflecting a holistic approach to urban planning that prioritizes regeneration and biodiversity. The prompt used for this case is much longer and included as Text Box~\ref{TexBox:city_prompt}}.

    \label{fig:Fig_2005}
\end{figure}

Figure~\ref{fig:Fig_2005} shows a few additional sample images specifically prompting the model to develop urban design ideas based on a set of biological materials, including spider silk, collagen and leaves, developed by  \texttt{lamm-mit/SmolLM-Base-1.7B-CPT-SFT-DPO}. The images presented illustrate a conceptual approach to urban design that synthesizes advanced architectural techniques with principles of ecological integration and sustainability. The structures exhibit biomimetic design, characterized by their spiraling, organic forms that mimic natural patterns, possibly influenced by leaf microstructures. This design approach aligns with the principles of biophilic architecture~\cite{kellert2008biophilic}, which aims to reconnect urban environments with the natural world by incorporating natural elements into the built environment.

Several similar architectural designs have been realized, validating the potential for these generated concepts. Examples include Azabudai Hills, a district in Tokyo featuring curving planted rooftops; CapitaSpring, a skyscraper in Singapore with orthogonal strips of plants embedded in its façade; and the Vertical Forest, designed by Stefano Boeri Architetti, which integrates residential buildings with diverse greenery. However, there remains significant potential to explore and expand these ideas within extended urban systems, for instance, how to interconnect individual structures, ensuring seamless integration with other facilities and alignment with an overall urban planning strategy.

The integration of nature into architecture has roots in movements such as organic architecture, as advocated by Frank Lloyd Wright~\cite{wright1954natural}, where the harmony between human habitation and the natural world is paramount. The designs proposed here, however, push this concept further by embedding extensive greenery directly into the architectural framework, creating vertical gardens and green terraces that are integral to the building’s structure rather than ancillary elements.
Moreover, the buildings are interconnected through elevated walkways, which not only facilitate human movement but also promote ecological connectivity, potentially serving as corridors for urban wildlife and contributing to biodiversity. This interconnectedness suggests a systems-thinking approach to urban design, where the built environment is considered part of a larger ecological network rather than an isolated entity.

The design concepts represented in these images could perhaps be seen as a potential paradigm shift in urban planning, where we move beyond sustainability to focus on regenerative design. This approach aims to create urban environments that not only minimize ecological impact but actively restore and enhance the natural environment. Such a model could potentially represent a significant advancement in urban ecology, proposing a future where cities operate as living systems, integrated with and supportive of their surrounding ecosystems. More work would be necessary to explore this, but this example illustrates a use case where the methods developed here can guide creative research and technology developments. 
 
\begin{figure}[H]
\begin{LLMbox}{Image generation prompt predicted by  \texttt{lamm-mit/SmolLM-Base-1.7B-CPT-SFT-DPO}}
\begin{lstlisting}
Design a futuristic metropolis that embodies the harmonious fusion of advanced technology, eco-friendliness, and human innovation. Utilize the following key geometric features to craft a visually striking and conceptually rich cityscape:

1. **Circular Architecture**: Integrate circular structures throughout the city, emphasizing the use of curvilinear forms reminiscent of ancient Greek temples or modern-day wind turbine farms. These circular buildings should be designed to maximize energy absorption through optimized aerodynamics and minimize visual clutter.

2. **Hierarchical Grid System**: Implement a sophisticated grid layout featuring nested hexagonal and pentagonal patterns, allowing for seamless connections between adjacent districts and optimizing traffic flow. Each node should contain a unique blend of residential, commercial, and recreational areas, fostering a sense of community and accessibility.

3. **Eco-Friendly Infrastructure**: Leverage cutting-edge green technologies by incorporating vertical gardens, rooftop solar panels, and advanced water management systems throughout the city. These elements will not only reduce carbon footprints but also contribute to a vibrant, verdant urban landscape.

4. **Human-Centric Design**: Prioritize pedestrian-friendly streetscapes, ensuring ample space for people-movers, parks, and public art installations. The city's central square should feature a majestic, undulating fountain that serves as a focal point for social gatherings and cultural events.

5. **Dynamic Lighting**: Employ adaptive lighting solutions that respond to time-of-day changes, weather patterns, and special occasions. This will result in a mesmerizing display of color shifts, shadows, and illumination effects that evoke a sense of wonder and enchantment.

In your generated image, please note the emphasis on capturing the essence of this futuristic city, blending technological advancements with timeless aesthetic values.
\end{lstlisting}
\end{LLMbox}
\textboxcaption{Image generation prompt based on a task given to \texttt{lamm-mit/SmolLM-Base-1.7B-CPT-SFT-DPO}, to think about specific design principles you can extract from combining dragline silk and collagen fibers to make a resilient, organic and living material and to incorporate design cues from leaf microstructures, to ultimately yield an image generation prompt for a futuristic city design.}
\label{TexBox:city_prompt}
\end{figure}

\section{Conclusions}
This study addressed fundamental questions in the fine-tuning of large language models (LLMs) for domain-specific knowledge, exploring how different optimization strategies and datasets influence model performance, and assessed effects of model size and capabilities. Our investigation focused on a host of techniques applied consistently across models/architectures and parameter numbers. These included  Continued Pre-Training (CPT), Supervised Fine-Tuning (SFT), Direct Preference Optimization (DPO), and Odds Ratio Preference Optimization (ORPO). The goal was to determine how these methods impact the specialization of LLMs, particularly in the context of engineering or science domains. 

A key finding of our research is that model scale plays a crucial role in the efficacy of fine-tuning strategies. Larger models, such as those with 7B and 8B parameters, not only exhibited substantial improvements in domain-specific tasks but also showed the emergence of novel capabilities—an outcome not observed in smaller models like the 1.7B parameter SmolLM model. This observation, as shown in a comparative plot (Figure~\ref{fig:Fig_109}) suggests a threshold effect, where the benefits of advanced optimization techniques, including model merging through SLERP (Spherical Linear Interpolation), become significantly more pronounced as model size increases. There are notable differences between the Llama and Mistral family of models that deserve further investigation. However, such investigations are hampered by the lack of detailed insights into the datasets used for training (during pre-training and fine-tuning) as well as a lack of details on specific training approaches. A summary of the key insights including the most effective approach is shown in Table~\ref{tab:ft_strategies}.

\begin{table}[h!]
\centering
\scriptsize
\begin{tabular}{|p{3cm}|p{3cm}|p{6cm}|}
\hline
\textbf{Model family} & \textbf{Base Model} & \textbf{FT Strategy} \\ \hline
Llama (8B)  & Instruct & CPT-SFT-ORPO-SLERP \\ \hline
Mistral (7B) & Base     & CPT-SFT-SLERP \\ \hline
SmolLM  (1.7B)& Base     & CPT-SFT-DPO \\ \hline
\end{tabular}
\caption{Summary of fine-tuning strategies and best base models used. For all cases, CPT-SFT is a critical step, but some model architectures show great improvements with additional DPO or ORPO steps. Model merging is effective primarily in larger models, whereas it has a detrimental effect in the smallest model considered (Figure~\ref{fig:Fig_109}).}
\label{tab:ft_strategies}
\end{table}

\begin{figure}
    \centering
    \includegraphics[width=.75\linewidth]{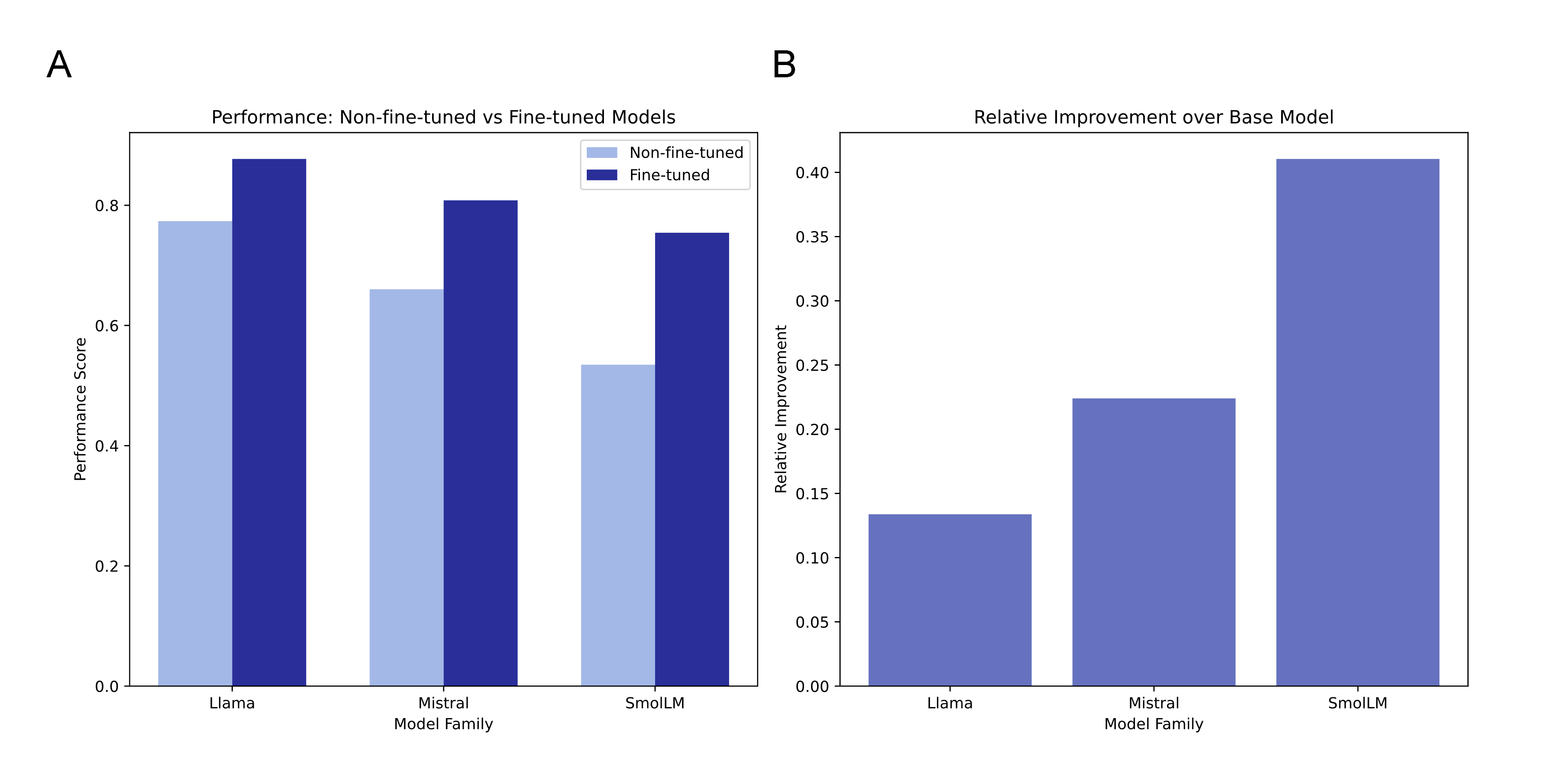}
    \caption{Summary of performance and performance gains. Panel A: Comparison of performance scores between non-fine-tuned and fine-tuned models across three model families: Llama, Mistral, and SmolLM. Fine-tuned models consistently outperform their non-fine-tuned counterparts, with the Llama model showing the highest performance overall. Panel B: Relative improvement in performance over the base models for each model family. SmolLM exhibits the greatest relative improvement, followed by Mistral and Llama, indicating that the fine-tuning process significantly enhances performance, particularly for the SmolLM model. The color palette transitions from light blue (non-fine-tuned) to dark blue (fine-tuned), visually emphasizing the performance gains achieved through fine-tuning.}
    \label{fig:Fig_25}
\end{figure}

The comparison between the \texttt{Llama-3.1-8b} and \texttt{Mistral-7B-v0.3} models highlighted how SLERP, when combined with SFT, DPO and ORPO, can effectively unlock synergistic properties, leading to refined performance outcomes that cluster distinctly in our analysis. However, these improvements were not mirrored in the smaller \texttt{SmolLM-1.7B} models, which showed a deterioration of performance under model merging, underscoring the importance of scale in realizing the full potential of fine-tuning techniques. Figure~\ref{fig:Fig_25} shows an overview of the significant performance gains achieved through fine-tuning across different model families, emphasizing how larger models like Llama and Mistral benefit more substantially from these advanced techniques in terms of the overall scale of performance reached. 

As shown in Figure~\ref{fig:Fig_25}A, fine-tuning consistently enhances model performance, with Llama leading in absolute performance scores. However, Figure~\ref{fig:Fig_25}B reveals an interesting trend: Despite SmolLM's lower baseline, its relative improvement from fine-tuning is the most pronounced, illustrating how smaller models, while not reaching the peak performance of their larger counterparts, can still gain substantially from targeted optimization strategies. This differential improvement pattern underscores the critical role of model size in both the efficacy and scope of fine-tuning processes. Consequently, while the larger Llama and Mistral models are better equipped to leverage the full spectrum of fine-tuning techniques like DPO and ORPO combined with SLERP, SmolLM demonstrates that even smaller-scale models can achieve meaningful performance boosts, albeit with limitations imposed by their smaller capacity.

\begin{figure}
    \centering
    \includegraphics[width=.65\linewidth]{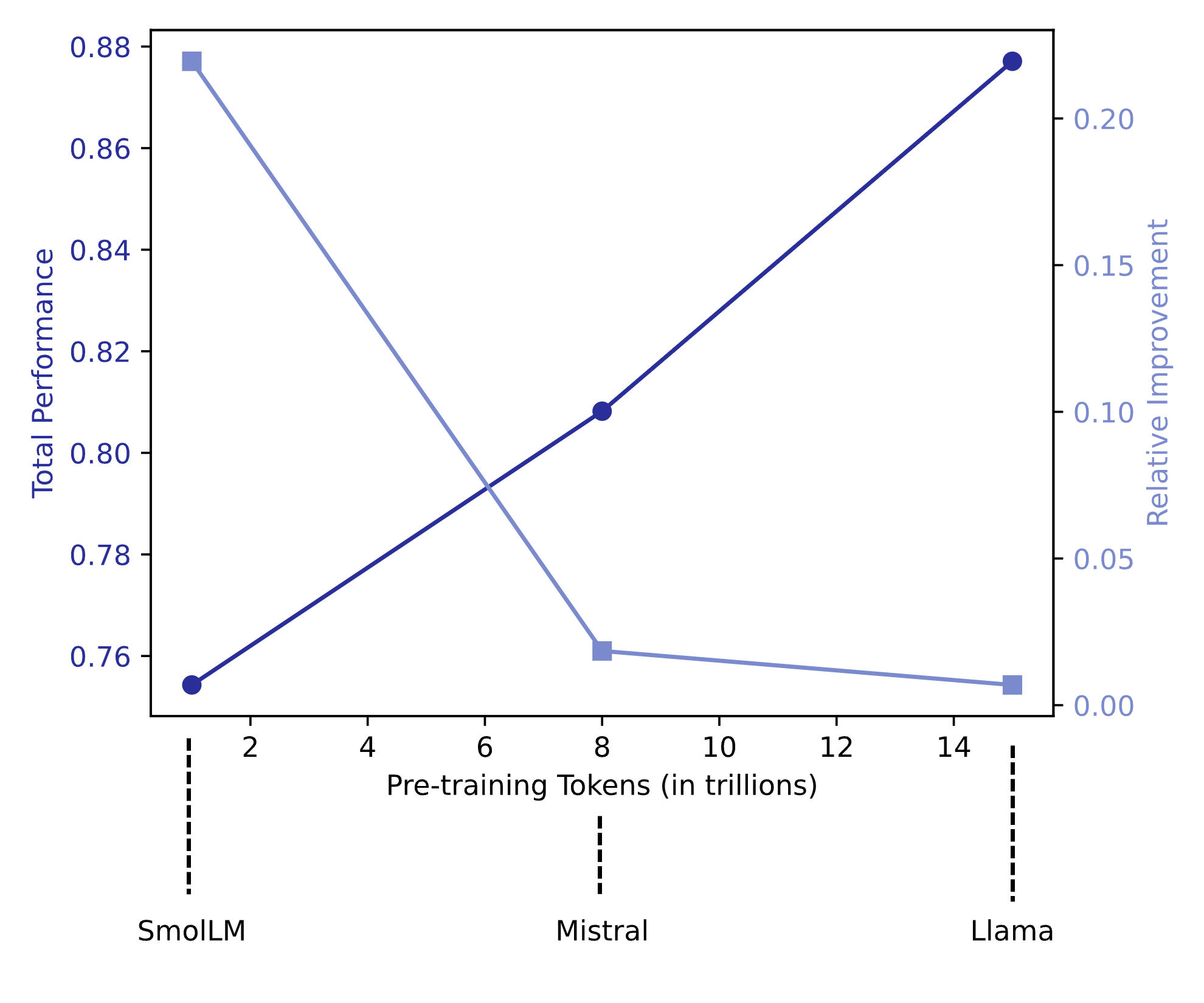}
    \caption{Performance over the number of pretraining tokens (1 trillion for SmolLM, 8 trillion for Mistral, and 15 trillion for Llama-3.1). This is an approximate analysis as each model/architecture has unique features to it, but an overall trend can be observed that more pre-training yields better performance, but also diminishes relative improvements that can be achieved.}
    \label{fig:Fig_26}
\end{figure}

Figure~\ref{fig:Fig_26} shows performance over the number of pretraining tokens (note, 1 trillion for SmolLM, 8 trillion for Mistral, and 15 trillion were used for Llama-3.1). While this is an approximate analysis as each model/architecture has unique features, training strategies, and datasets associated with it, but an overall trend can be observed that more pre-training tends to yield better performance, but also diminishes relative improvements that can be achieved. This agrees with the general consensus that for transformer-based models, there seem to be diminishing returns as training data and model size is increased, implying that as performance gets better, exponentially larger and better datasets and larger models must be used.

The results reported here contribute to the ongoing discourse on the optimal strategies for fine-tuning LLMs, particularly in specialized domains where precise and sophisticated understanding is required. The results suggest that while CPT and SFT are foundational, the application of advanced optimization methods like DPO and ORPO, especially when coupled with model merging techniques like SLERP, are critical for achieving high-performance outcomes in larger models. The data shows that there is room to optimize performance for each model, and some basic understanding needs to be developed how different base models perform and how they can best be improved. Overall, the use of CPT, SFT, and ORPO/DPO followed by model merging can be seen as a viable approach for strong domain performance. The use of CPT and the other steps tends to yield much better results than simpler approaches such as LoRA, albeit, at a much higher computational cost. For smaller models, model merging may not be advisable, albeit we would recommend experimentation for model architectures to explore this particular feature and also identify the threshold of model size in which emergent capabilities prominently appear.

There are numerous other parameters and angles to be explore. First, the effect of prompting. We kept the prompting consistent across all experiments. However, each model or model family will likely respond differently to prompts, and it is possible that additional performance gains can be achieved via prompt engineering. We deliberately did not address this issue here as we wanted to focus on an overall, consistent comparison. In some preliminary assessments we did notice that for the Mistral models, the CPT-SFT-DPO series resulted in rather low benchmark performance; where a closer inspection suggested that this may be due to limited ability to follow directions, and lengthy answers. Our benchmarks are consistent in that they test not only domain knowledge but also how closely models follow directives to answer in a certain way (single word, as done done here). Further experiments with the Mistral CPT-SFT-DPO series could entail using a more diverse dataset during DPO, including additional instruction-following foci. These and other variations are left to future work. 

We presented various materials-specific applications of the models that go beyond question-answer benchmarking, focusing on general reasoning capabilities to integrate complex, disparate biological materials concepts, step-by-step analysis, structured outputs, and others. Results of consistently prompted conversations with various models were shown in Text Boxes~\ref{TexBox:lamm-mit/Llama3.1-8b-Instruct-CPT-SFT-DPO}, \ref{TexBox:lamm-mit/Llama3.1-8b-Instruct-CPT-SFT-ORPO-SLERP-Var_G}, \ref{TexBox:lamm-mit/mistral-7B-v0.3-Base-CPT-SFT-DPO}, \ref{TexBox:lamm-mit/mistral-7B-v0.3-Base-CPT-SFT-SLERP}, and \ref{TexBox:lamm-mit/SmolLM-Base-1.7B-CPT-SFT-DPO}. These conversations of a human user with models provided important insights into the different strengths and weaknesses of the models in realistic use-cases. 

Building on this, focusing on another real-world use case of our models, we prompted the LLMs to develop image generation prompts, showing a powerful use case of the LLMs in conjunction with a pipeline of models interacting via agentic modalities (see, e.g. Figures~\ref{fig:Fig_2000}, \ref{fig:Fig_2001} and \ref{fig:Fig_2005} ranging from futuristic architecture, bio-inspired material microstructures, to urban/cityscape design). These studies could be further expanded by incorporating a feedback loop, where in a multi-agent framework, images generated can be analyzed by vision-LLMs (e.g. Cephalo~\cite{buehler2024cephalomultimodalvisionlanguagemodels}) and the insights fed back into the LLMs for improved reasoning and adjustments. 

Future research could further go deeper into the mechanisms behind the observed emergent behaviors, exploring how they can be harnessed across different architectures and domains, perhaps using tools from statistical mechanics or thermodynamics that may offer a fundamental perspective of how multi-particle systems behave and how these behaviors can be modeled. Other work to focus on interpretable insights could help also, following recent work~\cite{olah2024scaling}. Additionally, understanding the limitations and potential of smaller models remains an important area of inquiry, particularly as we seek to fine-tune LLMs for specific tasks without the extensive computational resources required for larger models. Other avenues include scale-up of the experiments to 70B or 405B models in the Llama series. Given the lessons learned from the present paper, a good strategy could be to do CPT on the Instruct model and then merge with the original Instruct model using SLERP. On the other side, more research could be done with the small LLMs. Our result based on SmolLM are interesting as they provide a model with reasonable performance on a tiny scale; this model could, for instance, become an effective tool when combined with in-context learning such as retrieval-augmented generation (RAG)~\cite{Run-llama/llama_index:Applications}. 

This study sheds light on the nuanced role of model scale and fine-tuning strategies in the development of domain-specific LLMs. By advancing our understanding of these dynamics, we move closer to unlocking the full capabilities of AI in specialized fields. The high degree of complexity across parameters and variables leave this to be a challenging field of study that offers many opportunities for future research. More work could also be done in improving the datasets. We find that using a larger dataset is not necessarily beneficial for downstream performance. We also found similar results for the Mistral model, where we did not see a significant decrease but an almost identical performance. Combined with other recent studies, this indicates that data quality is a major issue that can be addressed by further distillation, processing and perhaps filtering data components for relevance. The use of DPO or ORPO is particularly intriguing as it offers avenues for improving scientific accuracy and aligning the model with particular styles of responses (e.g. reasoning in a systematic way, step-by-step). Dataset curation could target several of these aspects and more experiments could shed light on the effects of these on performance. In the same vein, the incorporation of visual cues (e.g. figures, plots, microstructures, etc.) as done in recent work~\cite{buehler2024cephalomultimodalvisionlanguagemodels} can be another source for data.  

\section{Materials and methods}
\label{sec:Materials_Methods}
 
We provide details about the materials and methods used to conduct this study. 

\subsection{Dataset curation and processing}

The dataset used for training includes scientific papers from broad domains of biological materials, mechanics/mechanical properties, and spider silk. Earlier work focused on around 1,000 papers in training; here, the dataset consists of the original training set and an additional set of \textasciitilde4,300 papers in the realm of spider silk. All of the training was done with this integrated dataset unless mentioned otherwise. A few experiments were done where training also included an extended dataset of \textasciitilde3,800 more papers on biological materials, see details in Section~\ref{Integrated_data_definition}.

\subsubsection{Spider silk materials dataset}

\begin{figure}
    \centering
    \includegraphics[width=1\linewidth]{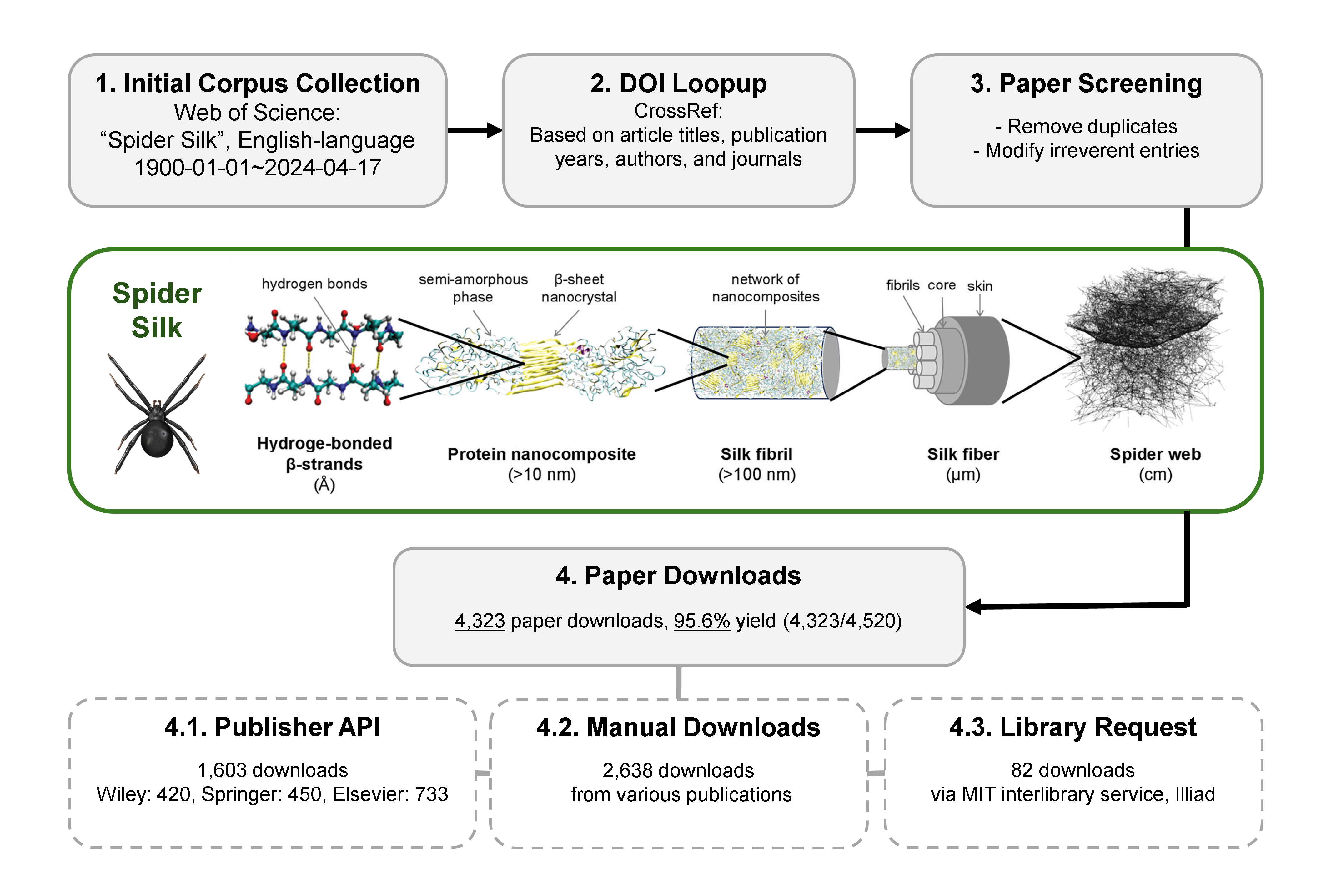}
    \caption{Paper collection process, here exemplified for the spider silk dataset. A total of 4,323 spider silk-related papers were collected and downloaded in PDF format. The process involved four key steps: collecting 4,520 papers from Web of Science, performing DOI lookups for missing entries using the \texttt{habanero.Crossref} Python client, screening for duplicates and irrelevant entries, and finally, downloading the papers through publisher APIs, manual downloads, and library requests. The final yield was 95.6\%. Image adapted from~\cite{Lu2023SpiderSilk} with permission.
}
    \label{fig:Fig_30}
\end{figure}

We introduce a new dataset, tailored to studies on spider silk. As the first step, we curated a set of PDF files using the method outlined in Figure~\ref{fig:Fig_30}. The process involved four key steps: 
\begin{enumerate}
    \item Collecting 4,520 papers from Web of Science,
    \item Performing digital object identifier (DOI) lookups for missing entries using the \texttt{habanero.Crossref} Python client,
    \item Screening for duplicates and irrelevant entries, and
    \item Downloading 4,323 papers through publisher APIs, manual downloads, and library requests.
\end{enumerate}
A total of 4,323 spider silk-related papers were downloaded in PDF format and used for training. 4,520 papers were initially collected from the Web of Science search engine on April 17, 2024, using the keywords ``spider silk''. The search was limited to English-language articles published between 1900-01-01 and 2024-04-17. For the collected papers that were missing DOIs, we used the \texttt{habanero.Crossref} Python client to interact with the CrossRef API, conducting DOI lookups based on the article titles, publication years, authors, and journals, where available. After initial collection, the paper entries were screened before downloading. Ultimately, 4,323 papers were successfully downloaded in PDF format, resulting in a 95.6\% yield. The remaining papers were not downloaded due to duplication, irrelevant content, or unavailability.

Among the 4,323 papers, 1,603 were downloaded using the APIs of three publishers (420 from Wiley, 450 from Springer Nature, and 733 from Elsevier), 2,638 were manually downloaded, and 82 out of 98 interlibrary requests were provided by MIT Illiad. All collected paper details are summarized in the corresponding supplementary information (\textquoteleft SI\_source\_articles\_1.csv\textquoteright) for easier identification.

\subsubsection{Biological materials dataset}

The original dataset developed from 1,034 biological materials papers was described in earlier work~\cite{Luu2023BioinspiredLLM:Materials}. The detailed information on original biological materials dataset is summarized in the Supplementary Information \textquoteleft SI\_source\_articles\_2.csv\textquoteright. A secondary much larger dataset was also created, which we refer to as the `extended dataset'. This extended dataset consists of 3,826 biological materials-related papers, captured from a larger scope of search terms including ``biological materials mechanical structure'' – a broader scope than the previous search terms ``biological materials mechanical hierarchical structure''. From the results, only a portion of the entries were retrieved from API-supported publishers including 2,159 articles from Elsevier, 749 articles from Wiley, and 998 articles from Springer Nature, rendering 3,826 articles. These articles were also retrieved in plain text format through the publisher API or by processing PDFs using python package PDF2TEXT, leading to generally more unpredictable and varied text and formats. The remaining articles can be retrieved through following the previously established procedure, ideally all in PDF format, but for the scope of this study, the larger dataset provides us a brief look into the effects of training with a larger and more varied dataset. The detailed information on the extended dataset of biological materials is available in \textquoteleft SI\_source\_articles\_3.csv\textquoteright 

\subsubsection{Integrated dataset}
\label {Integrated_data_definition}
For all training unless mentioned otherwise, we used a combination of the original 1,034 biological materials papers and the 4,323 spider silk papers. For select  cases in the `extended dataset', the combination of the 1,034 biological materials papers, 4,323 spider silk papers, and the 3,826 extended biological materials papers, was used for training to explore the effect of larger data in varied formats. This dataset included more papers, but with a variegated format and great number of errors due to the use of less effective PDF-to-text translation methods. 

\subsection{Dataset processing and preparation}

For best-quality raw dataset used in CPT we use Marker \url{https://github.com/VikParuchuri/marker}, a tool to convert PDFs into markup format. We found this work work well overall with consistent quality. 

\subsection{Dataset distillation and preparation for supervised learning}

The raw text from the scientific papers is processed using general-purpose LLMs (mostly \texttt{GPT-4o} or \texttt{GPT-4o-mini}) in different ways to yield higher-quality data, including data suitable for supervised learning consisting of question-answer pairs. We used the following strategies and objectives:

\begin{enumerate}
    \item Question-answer pairs: Based on chunks of text, we developed question-answer pairs 
    \item Quantitative material property extraction: We extract quantitative material properties from papers, yielding a list of facts and properties of materials  
    \item Extraction of summaries: Based on chunks of text, we developed summaries of content for a well-reasoned clean description of content
    \item Organizing the research content in papers into structured JSON format, with the following key components:

    \begin{itemize}
        \item \textbf{Title:} The \texttt{title} key contains the title of the article, summarizing the study's primary focus.
        
        \item \textbf{Insights:} The \texttt{insights} key includes an array of key findings or interpretations derived from the research. Each entry in this array provides a brief summary of significant insights.
    
        \item \textbf{Facts:} The \texttt{facts} key consists of an array of factual statements based on observations or results obtained during the study. These entries present concise, verifiable information.
    
        \item \textbf{Details:} The \texttt{details} key offers a more detailed exploration of specific aspects of the research. This array includes information on methods, structural characteristics, and other technical elements relevant to the study.
    
        \item \textbf{Comparisons:} The \texttt{comparisons} key comprises an array of comparative analyses made within the research. These comparisons may involve structural similarities with other proteins or evolutionary analogies.
    
        \item \textbf{Questions:} The \texttt{questions} key lists the primary research questions that the study aimed to address. Each entry corresponds to a significant inquiry within the research framework.
    
        \item \textbf{Answers:} The \texttt{answers} key provides corresponding answers to the research questions listed under the \texttt{questions} key. These entries summarize the study's conclusions or findings related to each question.
    \end{itemize}
\end{enumerate}

This approach involves extracting research content from scientific papers, allowing for clear and concise data extraction from the entire context of the scientific paper. The process begins with the development of question-answer pairs to address specific research inquiries. Additionally, quantitative material properties are extracted to compile verifiable facts and key findings. Summaries of the content are generated to provide well-reasoned, clean descriptions of the research. 

The JSON format includes key components such as the title of the paper, insights, factual statements, detailed methods, comparisons, research questions, and corresponding answers, creating a comprehensive and organized representation of the study.

As an example:
\begin{tcolorbox}[colframe=gray!75!black, colback=gray!10, title=JSON Structured Data, boxrule=0.5mm, top=2mm, bottom=2mm, left=2mm, right=2mm]
\begin{lstlisting}[language=json]
{
  "title": "Recombinant Spider Silk Nanomembranes: Properties, Formation, and Applications",
  "insights": [
    "The self-assembly of recombinant spider silk proteins can create functional nanomembranes with desirable mechanical properties.",
    "These nanomembranes can support cell growth and tissue engineering applications due to their biocompatibility and protein permeability.",
    "The unique properties of the membranes, such as elasticity and toughness, make them suitable for medical applications."
  ],
  "facts": [
    "The nanomembranes have a thickness of 250 nm to 440 nm.",
    "Ultimate engineering strain of over 220% and toughness of 5.2 MPa were recorded.",
    "The membranes support the growth of human keratinocytes, which form a confluent layer within three days.",
    "They are permeable to human blood plasma proteins and small molecules like Rhodamine B and dextrans of various sizes."
  ],
  "details": [
    "Nanomembranes formed from the self-assembly of FN-4RepCT spider silk protein.",
    "Membranes are cm-sized, free-standing, and have a thickness variability of +/-110 nm.",
    "Mechanical tests show a bursting pressure of 86 Pa and maximum vertical displacement of 4.3 mm.",
    "Effective mass diffusivity for small molecules through membranes was found to be significantly less than in an obstacle-free medium.",
    "Biocompatibility was demonstrated by minimal immune response and degradation within 2-4 weeks."
  ],
  "comparisons": [
    "The mechanical properties of the spider silk membranes are similar to thicker silk membranes despite being nanometer thick.",
    "Protein permeability of the developed nanomembranes exceeds that of previously produced synthetic or bio-derived membranes.",
    "The FN-4RepCT silk protein has enhanced stability and bioactivity compared to membranes reinforced with inorganic fillers or subjected to film compression."
  ],
  "questions": [
    "What are the main properties of the spider silk nanomembranes?",
    "How do the mechanical properties of these membranes compare to traditional silk membranes?",
    "What types of proteins can permeate through the spider silk nanomembranes?",
    "How quickly can human keratinocytes grow on the nanomembranes?",
    "What methods were used to characterize the mechanical and permeability properties of the membranes?",
    "What applications might benefit from the properties of these nanomembranes?",
    "How does temperature and concentration affect the thickness of the membranes during formation?",
    "Why is the lower permeability to 100 nm gold particles significant?",
    "What role do cell-binding motifs from fibronectin play in cell attachment and growth?",
    "What challenges are faced in developing biodegradable membranes for tissue engineering?"
  ],
  "answers": [
    "The main properties include ultrathin thickness (250-440 nm), high elasticity, significant toughness (5.2 MPa), and protein permeability.",
    "Despite being nanometer thick, the spider silk membranes exhibit comparable mechanical properties to thicker silk membranes, including high toughness and strain.",
    "The membranes are permeable to human blood plasma proteins as well as small molecules like Rhodamine B (0.5 kDa) and dextrans of sizes 3 and 10 kDa.",
    "Human keratinocytes can form a confluent layer within three days of seeding on either side of the membrane.",
    "Mechanical and permeability properties were characterized using bulge inflation tests, diffusion coefficient evaluations, and fluorescence assays.",
    "Applications include tissue engineering, organ-on-a-chip systems, and potential use in surgical transplants due to rapid keratinocyte growth.",
    "Thickness is controlled by adjusting the initial protein concentration and formation time, showing a linear relationship.",
    "The lower permeability to 100 nm gold particles indicates selective permeability which is crucial for ensuring membranes only allow certain biomolecules to pass.",
    "Cell-binding motifs facilitate integrin-mediated binding, enhancing the attachment and proliferation of cultured cells on the membrane surface.",
    "Challenges include ensuring both biocompatibility and protein permeability while maintaining mechanical strength in biodegradable membranes."
  ]
}
\end{lstlisting}
\end{tcolorbox}

For use in DPO/ORPO, we additionally construct rejected answer by instructing the LLM to develop a scientifically incorrect, wrong answer. Here is an example:

\begin{tcolorbox}[colframe=blue!75!black, colback=blue!10, title=Chosen Answer]
\textbf{User:} What molecular mechanisms underlie the hydration-induced cross-linking of aciniform spider silk, and how do these mechanisms impact its mechanical properties?

\textbf{Assistant:} The hydration-induced cross-linking of aciniform spider silk is primarily driven by the formation of $\beta$-sheet structures from the silk protein's existing $\alpha$-helical and random coil regions upon exposure to moisture. When the silk fibers are hydrated, the water molecules interact with the protein chains, leading to a reorganization of the protein structure. This process results in a partial unfolding of the $\alpha$-helical regions and encourages the aggregation of previously disordered or loosely structured regions into $\beta$-sheet nanostructures.

Utilizing solid-state nuclear magnetic resonance (SSNMR) methods, it has been shown that water exposure increases the $\beta$-sheet content of the silk, which rises from approximately 15\% in the native state to as high as 53\% after hydration and subsequent mechanical shear treatment. This structural transition is crucial because $\beta$-sheet structures are associated with increased stiffness and rigidity compared to the flexible and extensible $\alpha$-helical domains.

As a result of the increased $\beta$-sheet content and the subsequent inter-fiber cross-linking, the mechanical properties of hydrated aciniform silk fibers are significantly enhanced. The fibers exhibit higher stiffness, transitioning from a flexible state to a more rigid mat-like structure. For instance, after drying, the material stiffness can reach values around 4.76 GPa, indicating a notable increase in stiffness compared to the native state. This transformation allows the silk to maintain structural integrity while being subjected to mechanical stress, which is particularly beneficial during the spider's feeding process.

Overall, the hydration-induced molecular changes confer a remarkable adaptability to the aciniform silk, enabling it to switch from a flexible, extensible fiber to a strong, rigid material, enhancing its functionality in prey-wrapping and potentially opening avenues for synthetic applications in biomimetic materials.
\end{tcolorbox}

\vspace{0.5cm}

\begin{tcolorbox}[colframe=red!75!black, colback=red!10, title=Rejected Answer]
\textbf{User:} What molecular mechanisms underlie the hydration-induced cross-linking of aciniform spider silk, and how do these mechanisms impact its mechanical properties?

\textbf{Assistant:} The hydration-induced cross-linking of aciniform spider silk occurs because the silk proteins absorb water, which causes the fibers to become thicker and stiffer. This process essentially makes the silk stronger without any specific molecular changes. As a result, the silk fibers can stretch less and are more prone to breaking under tension. Overall, exposure to water simply makes the fibers less flexible.
\end{tcolorbox}

The entire scientific paper is used to construct this structured data. Five distinct sets are prepared for each paper, resulting in around 21,000 for the spider silk dataset and 5,000 for the bio-inspired materials dataset. 

This structured approach ensures that the dataset encapsulates both high-level insights and detailed factual information, facilitating comprehensive analysis and effective training. During CPT, this high-quality data is used together with the raw text (see Table~\ref{tab:dataset_summary} for a detailed breakdown; where unstructured text means that we simply concatenate question-answers, summaries, and other content). For SFT, we only use question-answer pairs.  For DPO/ORPO, triples of prompt, response, rejected response, are used. Details on the specific training sets used for each model and training stage are indicated for each model as shown in Table~\ref{tab:dataset_summary}.

\subsection{Benchmark development and prompting}
In our benchmarking strategy, we evaluated the model's ability to select the correct answer by employing a structured prompting approach. The primary prompt function instructed the model as follows: 

\begin{quote}
    \texttt{You respond with one word or letter. Select the correct answer to this question: \{q\}}
\end{quote}
followed by: 
\begin{quote}
    \texttt{The correct answer is:}
\end{quote}

This constrained the model's response to one word or letter, ideal for multiple-choice or binary (True/False) questions. The benchmarking was conducted with a deterministic setup without sampling (that is, $T=0$) to ensure consistency in results. The answers were summarized using a custom answer check that converted the model's response to a single letter corresponding to the correct choice (e.g., A, B, C, as wellas T, or F).

The benchmarks contain a diverse set of multiple-choice (MC) and true/false (TF) questions focused on biological material properties, applications, and production, categorized under topics like biology, materials science, gene studies, and methodologies. The questions assess understanding of both factual and conceptual information, with some involving numerical calculations or experimental techniques. Scenario-based questions explore, for instance, how spiders adapt their web designs to environmental factors, while logic-related and paraphrasing tasks evaluate reasoning skills. The content emphasizes biological material potential in advanced applications, such as biotechnology and materials engineering. 

\subsection{Benchmark development for model assessment}

Two benchmark datasets were developed, one focused on silk materials (``Spider silk benchmark'', available at \url{lamm-mit/spider-silk-benchmark}) and the other one general biological and bio-inspired materials (``Bio-inspired/biological materials benchmark'', available at \url{lamm-mit/bioinspired-benchmark}).

The general benchmark development process follows earlier work~\cite{Luu2023BioinspiredLLM:Materials} and included developing multiple-choice questions. The questions were developed by domain experts, based on knowledge extracted from relevant articles in the field. The benchmarks developed in this study each contain a question and answer, question types, knowledge categories, assessment areas, and referenced papers, if any, used for question development. New benchmarks were developed for spider silk to assess the new training corpus, and the original benchmark for bio-inspired materials was extended to include double the number of questions.

\begin{table}[h]
\centering
\scriptsize
\begin{tabular}{|p{2cm}|p{11cm}|p{2cm}|}
\hline
\textbf{Stage} & \textbf{Dataset} & \textbf{Fraction Used} \\ \hline

\multicolumn{3}{|c|}{\textbf{Llama} 8B and \textbf{Mistral} 7B (Base/Instruct model based) }\\ \hline
\multirow{4}{*}{CPT} & lamm-mit/bio-silk-mech-mix-80K & 2.0 \\ \cline{2-3}
 & Raw text biological materials papers (lamm-mit/raw-text-bio-marker$^\#$) & 1.0 \\ \cline{2-3}
 & Raw text spider silk materials papers (lamm-mit/raw-text-silk-marker$^\#$) & 1.0 \\ \cline{2-3}
 & lamm-mit/bio-materials-text-60K & 1.0 \\ \hline
\multirow{2}{*}{SFT} & lamm-mit/HuggingFaceH4-ultrachat\_200k & 1.0 \\ \cline{2-3}
 & lamm-mit/bio-silk-mech-mix-q-a-35K & 1.0 \\ \hline
\multirow{3}{*}{DPO/ORPO} & lamm-mit/orpo-dpo-mix-40k & 1.0 \\ \cline{2-3}
 & lamm-mit/bio-inspired-DPO & 1.0 \\ \cline{2-3}
 & lamm-mit/spider-silk-DPO & 1.0 \\ \hline

\multicolumn{3}{|c|}{\textbf{SmolLM} 1.7B (only Base-model based)} \\ \hline
\multirow{5}{*}{CPT} & lamm-mit/bio-silk-mech-mix-80K & 1.0 \\ \cline{2-3}
 & Raw text biological materials papers (lamm-mit/raw-text-bio-marker$^\#$) & 1.0 \\ \cline{2-3}
 & Raw text spider silk materials papers (lamm-mit/raw-text-silk-marker$^\#$) & 1.0 \\ \cline{2-3}
 & lamm-mit/bio-materials-text-60K & 2.0 \\ \hline
\multirow{6}{*}{SFT} & HuggingFaceTB/Magpie-Pro-300K-Filtered-H4 & 1.0 \\ \cline{2-3}
 & HuggingFaceTB/self-oss-instruct-sc2-H4 & 1.0 \\ \cline{2-3}
 & HuggingFaceTB/OpenHermes-2.5-H4 & 0.001 \\ \cline{2-3}
 & HuggingFaceTB/everyday-conversations-llama3.1-2k & 1.0 \\ \cline{2-3}
 & HuggingFaceTB/instruct-data-basics-smollm-H4 & 1.0 \\ \cline{2-3}
 & lamm-mit/bio-silk-mech-mix-q-a-35K & 1.0 \\ \hline
\multirow{3}{*}{DPO/ORPO} & lamm-mit/orpo-dpo-mix-40k & 1.0 \\ \cline{2-3}
 & lamm-mit/bio-inspired-DPO & 1.0 \\ \cline{2-3}
 & lamm-mit/spider-silk-DPO & 1.0 \\ \hline

\end{tabular}
\caption{Datasets used for training different stages of Llama (8 billion parameters), Mistral (7 billion parameters) and SmolLM (1.7 billion parameters) models. The Mistral-based training strategy uses the same dataset, with the CPT phase and domain-specific SFT and ORPO/DPO data added, as in the development of the Zephyr model~\cite{tunstall2023zephyrdirectdistillationlm}. The SmolLM mix of data is different as we follow the distribution and use of data as in the development of the SmolLM-Instruct model (\url{HuggingFaceTB/SmolLM-1.7B-Instruct}); however, with our domain specific data added during the SFT phase. The DPO phase had not been used in the original SmolLM-Instruct model. Note: $^\#$For the list of paper references in this dataset, see Supplementary Information.  }
\label{tab:dataset_summary}
\end{table}

\subsubsection{Spider silk benchmark}
We established a benchmark specifically for spider silk, comprising 159 question-answer pairs designed to evaluate model performance. The question types include both multiple choice and true or false, and can be divided into two sets, each designed for different purposes: (1) 105 basic questions primarily assess knowledge and understanding of spider silk, while (2) the remaining 54 advanced questions explore various scenarios involving logic, reasoning, and more. 

The details are as follows:
\begin{enumerate}
    \item Basic Query Set: We first prepared a query set focuses on the knowledge understanding of spider silk, consisting of 105 question-answer pairs, which includes 50 multiple choice questions and 55 true or false questions. These questions were further classified into descriptive, conceptual, analytical, numerical, comparative, and experimental categories, covering knowledge in various research areas such as materials, biology, application, gene, production, and methodology.
    \item Advanced Query Set: To further access different scenarios involving logic, reasoning, and creativity, 54 additional questions were designed. These were categorized by question types into 31 multiple choice and 23 true or false questions. Additionally, this advanced query set was organized by task types, including (1) scenario-based questions (e.g., experiment designs under specified scenarios, spider behavior or the mechanical performance of webs and silks under specified conditions), (2) advanced reasoning questions (e.g., reading comprehension, summarization/simplification, paraphrasing, numerical calculation), and (3) logic-related questions (e.g., fallacy identification, negation, semantic parsing, contradiction).
\end{enumerate}

This spider silk-specific benchmark is designed to evaluate not only knowledge recall but also the model’s capabilities in logic, reasoning, and creativity, ensuring a comprehensive assessment of the LLMs' cognitive and reasoning abilities. The benchmark query set is documented as supplementary information, with each entry containing question and answer, question type, knowledge category, assessment area, and referenced paper, if any, used for question development. Text Box~\ref{box:sample_questions_silk} shows a few sample questions.

\begin{figure}[h!]
    \centering
    \begin{LLMbox}[title=Spider silk benchmark]
        \\
        \textbf{Question 1:} \\
        \textit{"Spider silk is a remarkable natural fiber known for its exceptional strength and elasticity. It is produced by spiders using specialized silk glands and can be used in various applications, including medical sutures and biodegradable fishing nets."} \\
        \textbf{What are two key properties of spider silk mentioned in the passage?} \\
        A) Strength and elasticity \\
        B) Biodegradability and colorfulness \\
        C) Weight and flexibility \\
        D) Waterproof and adhesive properties \\
        \\
        \textbf{Answer:} A
        \vspace{0.3cm} \\
        \textbf{Question 2:} \\
        "Due to its impressive mechanical properties, spider silk is being considered for use in protective gear such as bulletproof vests. Its ability to absorb high amounts of energy before breaking makes it an ideal material for such applications." \\
        \textbf{Why is spider silk considered for use in protective gear?} \\
        A) It is very lightweight. \\
        B) It can absorb high amounts of energy before breaking. \\
        C) It is easy to color. \\
        D) It is water-soluble. \\
        \\
        \textbf{Answer:} B
        \vspace{0.3cm} \\
        \textbf{Question 3:} \\
        \textbf{Which protein domains in spider dragline silk are characterized by $\beta$-sheet conformation?} \\
        A) Poly(Gly) \\
        B) Poly(Ala) \\
        C) Poly(Gly-Ala) \\
        D) Poly(Ser) \\
        \\
        \textbf{Answer:} B
        \vspace{0.3cm} \\
        \textbf{Question 4:} \\
        Which environmental condition does not significantly affect the mechanical properties of spider dragline silk? \\
        A) Ambient humidity \\
        B) UV radiation \\
        C) Temperature up to 150°C \\
        D) Low temperatures (-60°C) \\
        \\
        \textbf{Answer:} C
    \end{LLMbox}
    \textboxcaption{Sample questions from the spider silk benchmark.}
    \label{box:sample_questions_silk}

\end{figure}

\subsubsection{Bio-inspired/biological materials benchmark}
In earlier work, a 100 multiple choice question benchmark was developed with challenging questions focusing on biological material mechanics. Since biological materials captures a large span of species and materials, four categories were developed for this domain including:
\begin{enumerate}
    \item General Questions (80): Cover broad topics and general trends
    \item Specific Questions (80): Focus on unique biological material mechanics or article-specific phenomena
    \item Numerical Questions (20): Ask for specific quantitative values documented in the literature
    \item Non-Biological Questions (20): Probe the model to distinguish between synthetic and biological materials, a common weakness noticed in foundational LLMs
\end{enumerate}

In this work, we expanded the benchmark and doubled the previous number of questions, following the same categories as outlined above. Text Box~\ref{box:sample_questions_second_dataset} shows a few sample questions. 

\begin{figure}[h!]
    \centering
    \begin{LLMbox}[title=Sample Questions: Bio-inspired/biological materials benchmark]
        \\
        \textbf{Question 1:} \\
        When describing gradient structures, they can be gradients in? \\
        A) dimension \\
        B) composition \\
        C) both dimension and composition \\
        \textbf{Answer:} C
        \vspace{0.3cm} \\
        \textbf{Question 2:} \\
        What is scanning electron microscopy used for? \\
        A) Imaging biological materials \\
        B) Testing biological materials for mechanical strength \\
        C) Measuring strain rate sensitivity of biological materials \\
        \textbf{Answer:} A
        \vspace{0.3cm} \\
        \textbf{Question 3:} \\
        What do Ashby plots for biological materials typically display? \\
        A) Strength as a function of density \\
        B) Strain rate to deformation and failure mode \\
        C) Microstructure to macrostructure \\
        \textbf{Answer:} A
        \vspace{0.3cm} \\
        \textbf{Question 4:} \\
        Which of the following biological materials is most notable for impact resistance? \\
        A) Stomatopod dactyl club \\
        B) Plant stems \\
        C) Bird feathers \\
        \textbf{Answer:} A
    \end{LLMbox}
    \textboxcaption{Sample questions from the bio-inspired/biological materials benchmark.}
    \label{box:sample_questions_second_dataset}
\end{figure}

\subsection{Training approach}

Table~\ref{tab:dataset_summary} summarizes the datasets used for each of the stages of training. While we use the same data to train the Llama and Mistral models, we alter the mix of SFT and DPO/ORPO training data to match the original SmolLM-Instruct model development, but with the domain-specific data added into it.

Training of the Llama and Mistral models was conducted on one 8xH100 node with 8 GPUs, whereas the SmolLM model was trained on a single  GPU. All training was conducted using the Hugging Face alignment handbook  (\url{https://github.com/huggingface/alignment-handbook}), using TRL trainers (for SFT, DPO, and ORPO). Additional details pertaining to each of the training stages is provided below.

\subsubsection{Continued pre-training (CPT)}
During CPT, we provide raw text to the model. A start token is added at the beginning of each chunk. We enhance training efficiency by using sample packing, where multiple short examples are packed into the same input sequence to match the total sequence length used. Training scripts that specify all parameters used are available at \url{https://github.com/lamm-mit/LLM-finetuning}.

\subsubsection{Supervised fine-tuning}
We use the relevant chat template associated with each model to provide samples in question-answer format, where the user role is used for the question and the assistant role for the answer. As in CPT, we use sample packing, whereby multiple short examples are packed together to match the total sequence length used.  Training scripts that specify all parameters used are available at \url{https://github.com/lamm-mit/LLM-finetuning}.

\subsubsection{DPO training}
We employed Direct Preference Optimization (DPO) as a fine-tuning strategy to enhance our language model's ability to generate preferred responses based on human feedback. As introduced in~\cite{rafailov2024directpreferenceoptimizationlanguage} DPO fine-tuning uses a dataset with chosen and rejected samples.  
In each prompt, one response was labeled as ``chosen'' (the preferred option) and the other as ``rejected''' (the less preferred option). This preference data trains the model to distinguish and prioritize responses that align with desired outputs.

The Hugging Face DPO trainer of the Transformer Reinforcement Learning library was utilized to fine-tune the model by directly maximizing the log-likelihood of the DPO loss. This optimization process is specifically designed to enhance the model's ability to replicate the preferences indicated by the data.

The training was facilitated through the specific model chat template, which enabled interaction-based fine-tuning by incorporating direct feedback into the model’s learning process. This method ensured that the model was trained not only on generating correct responses but also on aligning with human and scientific preferences in a conversational context. 

Training scripts that specify all parameters used are available at \url{https://github.com/lamm-mit/LLM-finetuning}.

\subsubsection{Odds Ratio Preference Optimization (ORPO) training}  
We employed Odds Ratio Preference Optimization (ORPO)~\cite{hong2024orpomonolithicpreferenceoptimization} as a preference alignment strategy to fine-tune our model. ORPO streamlines the process of preference-aligned supervised fine-tuning (SFT) by directly optimizing the model using a log odds ratio term appended to the negative log-likelihood (NLL) loss, thus eliminating the need for an additional reference model or preference alignment phase (realizing a form of preference-aligned SFT). This approach reduces computational and memory overhead while maintaining strong preference adaptation. 
This simplifies the process of aligning language models with human preferences, eliminating the need for complex reinforcement learning frameworks like RLHF (Reinforcement Learning from Human Feedback). Instead of fitting a reward model and then optimizing it, DPO directly adjusts the model's parameters by maximizing the log-likelihood of the preferred response relative to the dispreferred one. This approach results in a more stable, efficient, and computationally lightweight method for aligning language models with human preferences, often outperforming traditional RL-based methods as shown in~\cite{hong2024orpomonolithicpreferenceoptimization}.

In our ORPO training, we used datasets formatted identical to that required for DPO training, containing three key components: prompt, chosen, and rejected. For each prompt, a pair of responses was provided, with one labeled as ``chosen"" (the preferred response) and the other as ``rejected'' (the less preferred response). This dataset structure allowed the ORPOTrainer to optimize the model by minimizing the NLL loss, with an additional penalty applied to the rejected responses and a strong adaptation signal applied to the chosen responses.

The Hugging Face ORPO trainer of the Transformer Reinforcement Learning library was utilized  leveraging the ORPO algorithm to fine-tune the model without the need for a reference model. This setup enabled efficient preference optimization by integrating the log odds ratio directly into the training process, allowing the model to learn from human preferences in a streamlined manner. The use of a chat template facilitated the interaction-based fine-tuning, ensuring that the model was optimized for generating responses that align with the desired outputs in a conversational and scientifically accurate context.

Training scripts that specify all parameters used are available at \url{https://github.com/lamm-mit/LLM-finetuning}.

\subsubsection{End and padding token}
During SFT, ORPO and DPO phases, we set the pad token to be distinct from the EOS token to ensure that end tokens are not masked out during training. 

\subsection{Model merging}
We use MergeKit~\cite{goddard2024arceesmergekittoolkitmerging} to merge models. The primary method for model merging is Spherical Linear Interpolation (SLERP), which we found to perform best overall. 
Given two sets of model parameters \( \theta_1 \) and \( \theta_2 \), SLERP interpolates between them as follows.
Both sets of parameters are first normalized to lie on the unit hypersphere:

\[
\hat{\theta}_1 = \frac{\theta_1}{\|\theta_1\|}, \quad \hat{\theta}_2 = \frac{\theta_2}{\|\theta_2\|}
\]

The cosine of the angle \( \omega \) between the two parameter vectors is computed using the dot product:

\[
\cos(\omega) = \hat{\theta}_1 \cdot \hat{\theta}_2
\]

The interpolation between the two vectors on the unit hypersphere is then given by:

\[
\text{SLERP}(t) = \frac{\sin((1-t) \omega)}{\sin(\omega)} \hat{\theta}_1 + \frac{\sin(t \omega)}{\sin(\omega)} \hat{\theta}_2
\]

where \( t \in [0, 1] \) is the interpolation parameter. When \( t = 0 \), \( \text{SLERP}(t) \) returns \( \hat{\theta}_1 \), and when \( t = 1 \), it returns \( \hat{\theta}_2 \).

The final interpolated parameter vector is then re-scaled to the original magnitude:

\[
\theta_{\text{interpolated}} = \|\theta_1\|^{1-t} \|\theta_2\|^t \times \text{SLERP}(t)
\]

SLERP allows us to merge neural network weights in a way that respects the underlying geometry of the weight space. By interpolating along a spherical path, SLERP avoids the pitfalls of linear interpolation, such as passing through regions of high loss in the parameter space, leading to a smoother and more effective merge \cite{shoemake1985animating}. This method is beneficial in scenarios like transfer learning, ensemble methods, or creating hybrid models that combine the strengths of different pre-trained models.

The effectiveness of SLERP can be linked to principles observed in overparameterization and ensemble methods. Overparameterized neural networks tend to generalize well due to their high capacity, even when trained to near-zero error~\cite{belkin2019reconciling}. SLERP leverages this capacity by non-linearly combining parameters, enabling the emergence of new capabilities through complex interactions that linear interpolation might miss. 

Additionally, SLERP shares similarities with ensemble methods, which benefit from the diversity among models~\cite{hansen1990neural}. The interpolation process in SLERP acts as a continuous ensemble, where the non-linear combination of model parameters across the spherical path can activate new features that neither model exhibited individually. This process is particularly effective when the models being merged have learned complementary features, allowing SLERP to synergistically enhance their strengths and produce novel functionalities.

The overall effectiveness of SLERP in discovering new capabilities and improving performance can be speculated to scale with the diversity between the models being merged and their overall parameter count~\cite{kaplan2020scaling}.

A sample merge script for SLERP merges, for MergeKit, is: 
\begin{tcolorbox}[colback=gray!5!white,colframe=gray!75!black,title=MergeKit YAML (SLERP)]
\begin{lstlisting}[language=yaml,basicstyle=\ttfamily]
slices:
  - sources:
      - model: lamm-mit/Llama3.1-8b-Instruct-CPT-SFT-DPO
        layer_range: [0, 32]
      - model: meta-llama/Meta-Llama-3.1-8B-Instruct
        layer_range: [0, 32]
merge_method: slerp
base_model: meta-llama/Meta-Llama-3.1-8B-Instruct
parameters:
  t:
    - filter: self_attn
      value: [0, 0.5, 0.3, 0.7, 1]
    - filter: mlp
      value: [1, 0.5, 0.7, 0.3, 0]
    - value: 0.5
dtype: bfloat16
\end{lstlisting}
\end{tcolorbox}

The chat template used for the Llama models is 
\begin{tcolorbox}[colback=gray!5!white,colframe=gray!75!black,title=Chat template used for Llama derived models]
\begin{lstlisting}[basicstyle=\ttfamily]
<|begin_of_text|><|start_header_id|>system<|end_header_id|>

system message<|eot_id|>
<|start_header_id|>user<|end_header_id|>

user query 1<|eot_id|>
<|start_header_id|>assistant<|end_header_id|>

response 1<|eot_id|>

<|start_header_id|>user<|end_header_id|>

user query 2<|eot_id|>
<|start_header_id|>assistant<|end_header_id|>

...
\end{lstlisting}
\end{tcolorbox}
The vocabulary size of the Llama family models is 128,000.

The chat template used for the Mistral models is 
\begin{tcolorbox}[colback=gray!5!white,colframe=gray!75!black,title=Chat template used for Mistral derived models]
\begin{lstlisting}[basicstyle=\ttfamily]
<s>[INST] system message

user query 1[/INST] response 1</s>[INST] user query 2[/INST] ...
\end{lstlisting}
\end{tcolorbox}
The vocabulary size of the Mistral family models is 32,768.

The chat template used for the SmoLM models is 
\begin{tcolorbox}[colback=gray!5!white,colframe=gray!75!black,title=Chat template used for SmoLM derived models]
\begin{lstlisting}[basicstyle=\ttfamily]
<|im_start|>system
system message<|im_end|>
<|im_start|>user
user query 1<|im_end|>
<|im_start|>assistant
response 1<|im_end|>
<|im_start|>user
user query 2<|im_end|>
<|im_start|>assistant
response 2<|im_end|>
<|im_start|>assistant 
...
\end{lstlisting}
\end{tcolorbox}
The vocabulary size of the SmoLM family models is 49,152.

\subsection{Clustering analysis and dendrograms}

In this analysis, we standardized the performance scores of the merged models to facilitate meaningful comparisons across different models. The performance of a merged model is denoted as \( P_{\text{merged}} \), while the performance of the two parent models is denoted as \( P_1 \) and \( P_2 \). The standardized scores were computed as follows. 

The expected score \( E_i \) for the \( i \)-th merged model was calculated as the average of the performances of the two parent models:
\[
E_i = \frac{P_1 + P_2}{2}
\]
Let \( \mu_E \) and \( \sigma_E \) represent the mean and standard deviation of the expected scores across all merged models, respectively. The standardized expected score for the \( i \)-th merged model, denoted as \( Z_{E_i} \), was then calculated using the formula:
\[
Z_{E_i} = \frac{E_i - \mu_E}{\sigma_E} = \frac{\left(\frac{P_1 + P_2}{2}\right) - \mu_E}{\sigma_E}
\]

The actual score \( A_i \) for the \( i \)-th merged model is defined as its observed performance:
\[
A_i = P_{\text{merged}}
\]
Let \( \mu_A \) and \( \sigma_A \) represent the mean and standard deviation of the actual scores across all merged models, respectively. The standardized actual score for the \( i \)-th merged model, denoted as \( Z_{A_i} \), was calculated using the formula:
\[
Z_{A_i} = \frac{A_i - \mu_A}{\sigma_A} = \frac{P_{\text{merged}} - \mu_A}{\sigma_A}
\]
See Table~\ref{tab:parameter_definitions} for a summary of key parameters and definitions. 

\begin{table}[h!]
\centering
\scriptsize
\begin{tabular}{|c|l|}
\hline
\textbf{Parameter} & \textbf{Definition} \\ \hline
\( P_{\text{merged}} \) & Performance of the merged model. \\ \hline
\( P_1 \) & Performance of the first parent model. \\ \hline
\( P_2 \) & Performance of the second parent model. \\ \hline
\( E_i \) & Expected score for the \( i \)-th merged model, defined as the average of \( P_1 \) and \( P_2 \). \\ \hline
\( A_i \) & Actual score for the \( i \)-th merged model, equal to \( P_{\text{merged}} \). \\ \hline
\( \mu_E \) & Mean of the expected scores across all merged models. \\ \hline
\( \sigma_E \) & Standard deviation of the expected scores across all merged models. \\ \hline
\( \mu_A \) & Mean of the actual scores across all merged models. \\ \hline
\( \sigma_A \) & Standard deviation of the actual scores across all merged models. \\ \hline
\( Z_{E_i} \) & Standardized expected score for the \( i \)-th merged model. \\ \hline
\( Z_{A_i} \) & Standardized actual score for the \( i \)-th merged model. \\ \hline
\end{tabular}
\caption{Definitions of parameters used in the clustering and dendrogram analysis.}
\label{tab:parameter_definitions}
\end{table}

\subsection{Summary of best-performing model releases on Hugging Face}
The models with the best performance, along with their corresponding Hugging Face hub IDs, are summarized in Table~\ref{tab:ft_strategies_best_model}.
\begin{table}[h!]
\scriptsize
\centering
\begin{tabular}{|p{7cm}|p{7cm}|p{1cm}|}
\hline
\textbf{Model} & \textbf{Hugging Face Hub ID} & \textbf{Notes} \\ \hline
lamm-mit/Llama3.1-8b-Instruct-CPT-ORPO-SLERP\_Var\_G & lamm-mit/Llama3.1-8b-Instruct-CPT-SFT-ORPO-SLERP\_Var\_G-09022024 &  \\ \hline
lamm-mit/Llama3.1-8b-Instruct-CPT-ORPO-SLERP &  lamm-mit/Llama3.1-8b-Instruct-CPT-SFT-ORPO-SLERP-09022024&  \\ \hline

lamm-mit/mistral-7B-v0.3-Base-CPT-SFT-SLERP & lamm-mit/mistral-7B-v0.3-Base-CPT-SFT-SLERP-09022024 &  \\ \hline
lamm-mit/mistral-7B-v0.3-Base-CPT-SFT-DPO & lamm-mit/mistral-7B-v0.3-Base-CPT-SFT-DPO-09022024 &  \\ \hline

lamm-mit/SmolLM-Base-1.7B-CPT-SFT-DPO & lamm-mit/SmolLM-Base-1.7B-CPT-SFT-DPO-09022024 &  \\ \hline
\end{tabular}
\caption{Overview of best-performing models and associated hub ID on Hugging Face.}
\label{tab:ft_strategies_best_model}
\end{table}

\subsection{Multi-turn human-AI conversations}
We define human input and system prompts consistently for all experiments. We use $\mathrm{top}_k$=512, $\mathrm{top}_p=0.9$, and $\mathrm{repetition\_penalty}=1.1$. We set the maximum number of generated tokens to 1024 to allow for long, detailed outputs during multi-turn conversations.  
 
\subsection{Image synthesis}

Image synthesis is conducted using \url{lamm-mit/leaf-FLUX.1-dev}. This model is a fine-tuned version of the \url{black-forest-labs/FLUX.1-dev} base model using leaf images with a keyword `<leaf microstructure>', following the concept proposed in~\cite{ruiz2023dreamboothfinetuningtexttoimage}. The training set can be found at \url{lamm-mit/leaf-flux-images-and-captions}. The model was trained for $N=2,000$ steps using \texttt{adamw8bit}, at a learning rate of $0.0001$, and LoRA adapters applied to all linear layers of rank $r=16$, with $\alpha=16$. The purpose of demonstrating the use of an image generation model utilizing biologically inspired cues is to steer solutions towards organic, abstract, biological forms. Image generation is typically conducted with 25 denoising steps and guidance scale=3.5. An alternative model is available at \url{lamm-mit/leaf-L-FLUX.1-dev}, trained for $N=4,000$ steps using \texttt{adamw8bit}, at a learning rate of $0.0001$, and LoRA adapters applied to all linear layers of rank $r=64$, with $\alpha=64$. The larger FLUX model provides a stronger effect of inducing leaf microstructure patterns. 

For conversations targeted for image synthesis, we use $\mathrm{top}_k$=512, $\mathrm{top}_p=0.9$, and $\mathrm{repetition\_penalty}=1.1$. We set the maximum number of generated tokens to 1024 to allow for long, detailed outputs of the models. 

\section*{Author contributions}
MJB designed the overall research, developed algorithms, codes and conducted the training, assessments and analysis. MJB developed and executed the distillation strategies to generate structured and adversarial data. WL and RL curated the scientific papers for the corpus of raw training data, downloaded the papers, and designed the benchmark questions. MJB wrote the initial paper draft with input from WL and RL, and all authors edited and finalized the manuscript. 

\section*{Conflicts of interest}
The authors declare no conflicts of interest.

\section*{Code, data and model weight availability}

Training scripts are available at \url{https://github.com/lamm-mit/LLM-finetuning}. Model weights can be found at \url{https://huggingface.co/lamm-mit}. Training data is available as listed in Table~\ref{tab:dataset_summary}. References and DOI numbers of the downloaded papers used to construct the dataset are attached as Supplementary Information ( \textquoteleft SI\_source\_articles\_1.csv\textquoteright,  \textquoteleft SI\_source\_articles\_2.csv\textquoteright and  \textquoteleft SI\_source\_articles\_3.csv\textquoteright). 

\section*{Acknowledgments}
This work was supported in part by Google, the MIT Generative AI Initiative, USDA (grant number 2021-69012-35978), with additional support from NIH. This material is partially based upon work supported by the National Science Foundation Graduate Research Fellowship under Grant number 2141064. 

\bibliographystyle{naturemag}
\bibliography{references,references-Mendeley}  

\clearpage

\textboxcaption{Conversation generated by the model: \texttt{lamm-mit/Llama3.1-8b-Instruct-CPT-SFT-DPO}}
\label{TexBox:lamm-mit/Llama3.1-8b-Instruct-CPT-SFT-DPO}

\begin{LLMboxmultipage}{Conversation with: \texttt{lamm-mit/Llama3.1-8b-Instruct-CPT-SFT-DPO}}
\begin{lstlisting}

(*@\bf\hlblue{System:}@*) Your are a helpful and friendly assistant. You are an expert in materials science with special knowledge in biological materials. You provide concise but accurate responses. You are an expert in combining different concepts, often not obviously related, and provide accurate accounts of correct relationships or analogies. 

(*@\bf\hlred{User:}@*) Collagen and leaves, discuss their relationship.

(*@\bf\hllightgreen{Assistant:}@*) At first glance, collagen and leaves may seem like unrelated materials. However, they share some interesting similarities in terms of their structure and function.

Collagen is a protein that forms the structural framework of many tissues in our body, including skin, tendons, ligaments, cartilage, and bone. It has a unique triple-helix structure, known as a tropocollagen helix, which gives it exceptional mechanical properties such as tensile strength and elasticity.

Similarly, leaves have a hierarchical structure that allows them to perform essential functions for plant growth and survival. The main components of a leaf include the epidermis (the outermost layer), mesophyll cells (where photosynthesis occurs), and vascular tissue (which transports water and nutrients). Like collagen, these structures must be strong enough to withstand environmental stresses while also being flexible enough to adapt to changing conditions.

One key similarity between collagen and leaves is their ability to self-assemble into complex networks. In collagen, this process involves the formation of fibrils, which then aggregate into larger fibers and bundles. Similarly, the various cell types and tissues within a leaf work together to form a functional unit that can respond to external stimuli.

Another connection between collagen and leaves lies in their response to stress. Both materials exhibit nonlinear elastic behavior, meaning that their stiffness changes in response to applied force. This property allows them to absorb energy and recover from deformation, which is crucial for maintaining their integrity under varying loads.

In summary, although collagen and leaves come from vastly different sources, they share important structural and functional similarities that allow them to perform similar roles in their respective systems.

(*@\bf\hlred{User:}@*) Thank you! Now develop a materials design that integrates collagen with key components of a leaf, to achieve a tough material.

(*@\bf\hllightgreen{Assistant:}@*) To create a tough material that combines the strengths of collagen and key components of a leaf, we could consider integrating several features:

1. Collagen-based matrix: We start by creating a collagen-based matrix, which provides a robust foundation for the composite material. This matrix would mimic the structure of collagen fibrils found in natural collagen, allowing for high tensile strength and toughness.

2. Vascular-like channels: Next, we incorporate vascular-like channels into the collagen matrix. These channels would mimic the transport system found in leaves, allowing for efficient distribution of nutrients and waste products throughout the material. The channels could be designed using a combination of synthetic polymers and collagen, ensuring both flexibility and durability.

3. Epidermal-like layers: To further enhance the material's performance, we add epidermal-like layers on either side of the collagen matrix. These layers would provide additional protection against environmental stressors and help regulate moisture content, much like the epidermis in leaves. They could be composed of a biocompatible polymer, such as poly(lactic acid) (PLA), which offers good mechanical properties and biodegradability.

4. Mesophyll-like cells: Finally, we introduce mesophyll-like cells into the material. These cells would mimic the photosynthetic capabilities of leaves, providing a source of energy for the material. They could be created using a combination of living cells and synthetic polymers, ensuring both functionality and stability.

By integrating these four components - collagen matrix, vascular-like channels, epidermal-like layers, and mesophyll-like cells - we can create a tough, multifunctional material that combines the best attributes of collagen and key components of a leaf. This material could find applications in biomedical engineering, where its ability to distribute nutrients and support cellular activity would be particularly valuable.

(*@\bf\hlred{User:}@*) Summarize these design features in JSON format.

(*@\bf\hllightgreen{Assistant:}@*) {
    "name": "Collagen-Leaf Composite Material",
    "description": "Tough, multifunctional material combining collagen and key components of a leaf.",
    "components": [
        {
            "type": "matrix",
            "material": "collagen"
        },
        {
            "type": "channels",
            "material": "synthetic polymer + collagen"
        },
        {
            "type": "layers",
            "material": "poly(lactic acid)"
        },
        {
            "type": "cells",
            "material": "living cells + synthetic polymer"
        }
    ],
    "functions": [
        "High tensile strength and toughness",
        "Efficient nutrient and waste product distribution",
        "Environmental stress resistance",
        "Moisture regulation",
        "Photosynthetic capability"
    ]
}

\end{lstlisting}
\end{LLMboxmultipage}

\clearpage
 
\textboxcaption{Conversation generated by the model: \texttt{lamm-mit/Llama3.1-8b-Instruct-CPT-SFT-ORPO-SLERP-Var\_G}}
\label{TexBox:lamm-mit/Llama3.1-8b-Instruct-CPT-SFT-ORPO-SLERP-Var_G}

\begin{LLMboxmultipage}{Conversation with: \texttt{lamm-mit/Llama3.1-8b-Instruct-CPT-SFT-ORPO-SLERP-Var\_G}}
\begin{lstlisting}
(*@\bf\hlblue{System:}@*) Your are a helpful and friendly assistant. You are an expert in materials science with special knowledge in biological materials. You provide concise but accurate responses. You are an expert in combining different concepts, often not obviously related, and provide accurate accounts of correct relationships or analogies. 

(*@\bf\hlred{User:}@*) Collagen and leaves, discuss their relationship.

(*@\bf\hllightgreen{Assistant:}@*) While collagen and leaves may seem unrelated at first glance, they share some interesting connections.

1. **Structural Similarities**: Both collagen and leaf structures exhibit hierarchical organization. Collagen is a protein that forms the primary structural component of connective tissues like skin, tendons, and bones. It has a helical structure that self-assembles into fibrils, which then form fibers and eventually tissues. Similarly, leaves have a hierarchical structure, starting from individual cells to complex vascular networks.

2. **Mechanical Properties**: Collagen provides strength and elasticity to tissues due to its unique triple-helix structure. Leaves also need to withstand mechanical stresses such as wind and rain. The arrangement of cellulose microfibrils within plant cell walls gives them tensile strength and flexibility, similar to how collagen's structure contributes to tissue resilience.

3. **Biological Functionality**: In living organisms, collagen plays a crucial role in maintaining tissue integrity and facilitating movement (e.g., in tendons). Leaves, on the other hand, perform photosynthesis, absorbing sunlight to produce energy for plants through chlorophyll. However, both components contribute to the overall health and function of the organism - collagen supports bodily functions, while leaves support life by producing oxygen and organic compounds.

4. **Materials Science Applications**: Understanding the properties of collagen can inform biomaterials design, particularly in medical applications where tissue repair or replacement is needed. Similarly, studying the mechanics of leaves can inspire the development of lightweight yet strong materials for engineering applications.

In summary, while collagen and leaves serve distinct biological purposes, they share commonalities in terms of structural organization, mechanical properties, and potential applications in materials science.

(*@\bf\hlred{User:}@*) Thank you! Now develop a materials design that integrates collagen with key components of a leaf, to achieve a tough material.

(*@\bf\hllightgreen{Assistant:}@*) Designing a material that combines the beneficial properties of collagen with those of leaves requires understanding the unique characteristics of each component. Here's a hypothetical approach:

**Material Design: "Leaf-Collagen Hybrid"**

**Key Components:**

1. **Collagen Fibrils**: These would be the primary structural elements, providing strength and elasticity.
2. **Cellulose Microfibrils**: Inspired by the leaf's cell wall structure, these would add rigidity and resistance to deformation.
3. **Chloroplast-Inspired Nanoparticles**: Incorporating nanoparticles that mimic the light-absorbing capabilities of chloroplasts could enhance the material's ability to harness energy from external sources.

**Manufacturing Process:**

1. **Self-Assembly**: Collagen fibrils would be allowed to self-assemble into a network, mimicking natural tissue formation.
2. **Nanoparticle Integration**: Chloroplast-inspired nanoparticles would be dispersed throughout the collagen matrix, potentially using techniques like electrostatic deposition or chemical bonding.
3. **Cellulose Microfibril Reinforcement**: Cellulose microfibrils would be embedded within the collagen matrix, either through mechanical mixing or chemical cross-linking.

**Properties:**

1. **Toughness**: The combination of collagen's elasticity and cellulose's rigidity should result in a material that can absorb significant amounts of energy without breaking.
2. **Lightweight**: The use of cellulose microfibrils and nanoparticles could help reduce the material's weight while maintaining strength.
3. **Energy Harvesting**: The chloroplast-inspired nanoparticles might enable the material to convert environmental energy (e.g., light) into usable power, making it suitable for applications like wearable technology or biomedical devices.

**Potential Applications:**

1. **Biomedical Implants**: The Leaf-Collagen Hybrid could be used in implants that require both strength and biocompatibility, such as artificial joints or tissue scaffolds.
2. **Wearable Technology**: Its energy harvesting capabilities make it an attractive candidate for smart clothing or accessories that can power small electronics.
3. **Environmental Monitoring**: The material's ability to respond to environmental stimuli could be leveraged for monitoring air quality or detecting changes in temperature.

This hypothetical design illustrates how integrating key components of collagen and leaves could lead to the creation of a tough, lightweight, and energy-harvesting material with diverse applications.

(*@\bf\hlred{User:}@*) Summarize these design features in JSON format.

(*@\bf\hllightgreen{Assistant:}@*) ```json
{
    "name": "Leaf-Collagen Hybrid",
    "components": [
        {
            "type": "collagen fibrils",
            "function": "primary structural element, provides strength and elasticity"
        },
        {
            "type": "cellulose microfibrils",
            "function": "rigidity and resistance to deformation"
        },
        {
            "type": "chloroplast-inspired nanoparticles",
            "function": "energy harvesting capability"
        }
    ],
    "manufacturing_process": [
        {
            "step": "self-assembly of collagen fibrils",
            "description": "mimics natural tissue formation"
        },
        {
            "step": "integration of nanoparticles",
            "description": "electrostatic deposition or chemical bonding"
        },
        {
            "step": "reinforcement with cellulose microfibrils",
            "description": "mechanical mixing or chemical cross-linking"
        }
    ],
    "properties": [
        {
            "property": "toughness",
            "value": "high energy absorption capacity"
        },
        {
            "property": "lightweight",
            "value": "reduced weight while maintaining strength"
        },
        {
            "property": "energy harvesting",
            "value": "conversion of environmental energy into usable power"
        }
    ],
    "applications": [
        {
            "application": "biomedical implants",
            "description": "artificial joints or tissue scaffolds requiring strength and biocompatibility"
        },
        {
            "application": "wearable technology",
            "description": "smart clothing or accessories powering small electronics"
        },
        {
            "application": "environmental monitoring",
            "description": "detecting changes in temperature or air quality"
        }
    ]
}
```

This JSON representation summarizes the design features of the Leaf-Collagen Hybrid, including its components, manufacturing process, properties, and potential applications.


\end{lstlisting}
\end{LLMboxmultipage}

\clearpage
 
\textboxcaption{Conversation generated by the model: \texttt{lamm-mit/mistral-7B-v0.3-Base-CPT-SFT-DPO}}
\label{TexBox:lamm-mit/mistral-7B-v0.3-Base-CPT-SFT-DPO}

\begin{LLMboxmultipage}{Conversation with: \texttt{lamm-mit/mistral-7B-v0.3-Base-CPT-SFT-DPO}}
\begin{lstlisting}
(*@\bf\hlblue{System:}@*) Your are a helpful and friendly assistant. You are an expert in materials science with special knowledge in biological materials. You provide concise but accurate responses. You are an expert in combining different concepts, often not obviously related, and provide accurate accounts of correct relationships or analogies. 

(*@\bf\hlred{User:}@*) Collagen and leaves, discuss their relationship.

(*@\bf\hllightgreen{Assistant:}@*) As an artificial intelligence, I'm glad to assist you in exploring the relationship between collagen and leaves, specifically by highlighting the following aspects:

**Collagen**: Collagen is a fibrous structural protein that is the most abundant protein in mammals. It is a key component of various tissues such as tendons, ligaments, skin, bones, cartilage, and blood vessels. Collagen molecules consist of three polypeptide chains (alpha-1, alpha-2, and alpha-3) arranged in a triple helix structure. These chains are composed of repeating sequences of amino acids, including glycine, proline, and hydroxyproline.

**Leaves**: Leaves are specialized organs in plants that play a crucial role in photosynthesis, respiration, and transpiration. They have several layers of cells, each with specific functions. The outermost layer, known as the cuticle, is a waxy layer that helps reduce water loss through transpiration. Beneath the cuticle lies the epidermis, which contains the stomata (tiny pores).

The connection between collagen and leaves may seem unrelated at first glance, but let me explain how some analogies can be drawn:

1. **Structural similarities**: Both collagen and leaves exhibit hierarchical structures. Collagen has a triple helical arrangement of polypeptide chains, while leaves have a layered structure consisting of the cuticle, epidermis, mesophyll, and vascular tissue. Both structures contribute to the mechanical strength and resilience of the respective organisms.

2. **Mechanical properties**: Collagen exhibits high tensile strength and elasticity due to its unique triple helix structure. Similarly, leaves possess remarkable mechanical properties due to their cellular arrangement and composition. For example, the sclerenchyma cells in the leaf veins provide support and rigidity to the leaf structure.

3. **Adaptability**: Collagen adapts to changes in the body's physiological conditions, such as during wound healing or tissue regeneration. In a similar manner, leaves adapt to environmental factors like wind, rain, and temperature fluctuations. This adaptation is achieved through changes in the cellular structure and arrangement.

4. **Functionality**: Collagen plays a vital role in maintaining the integrity and function of various tissues in the body. Leaves perform essential functions for plant growth and survival, such as gas exchange, nutrient absorption, and light capture.

In conclusion, while collagen and leaves may appear dissimilar on the surface, they share many underlying characteristics that make them both fascinating examples of biological materials. By understanding these similarities, we can gain insights into the fundamental principles that govern the design and function of natural systems.

(*@\bf\hlred{User:}@*) Thank you! Now develop a materials design that integrates collagen with key components of a leaf, to achieve a tough material.

(*@\bf\hllightgreen{Assistant:}@*) To create a tough material integrating collagen with key components of a leaf, I will propose a bioinspired composite material concept that combines specific structural and mechanical features from both collagen and leaf tissues.

**Material Components:**
1. **Collagen Fibrils**: We will utilize collagen fibrils as the primary reinforcing fibers in our composite material. Collagen fibrils, known for their exceptional tensile strength and elasticity, will serve as the load-bearing backbone of the composite.

2. **Cellulose Nanofibers (CNFs)**: CNFs, derived from the cell walls of plants, including leaves, offer excellent mechanical properties like stiffness, strength, and toughness. Their nanoscale dimensions allow them to form strong interfacial bonding with other materials, enhancing the overall performance of the composite.

3. **Silk Fibroin**: Silk fibroin, another biopolymer, provides additional mechanical strength and flexibility. Its molecular structure allows it to form robust interactions with collagen and CNFs, contributing to the overall cohesion and toughness of the composite.

4. **Nanocellulose-Based Matrix**: A matrix made from nanocellulose, such as bacterial cellulose, will be used to bind the reinforcing fibers together. This matrix offers superior mechanical properties compared to traditional polymer matrices, allowing the composite to maintain its toughness under stress.

**Design Strategy:**
1. **Preparation of Collagen Fibrils**: Collagen fibrils will be extracted and purified using established methods.

2. **Production of Cellulose Nanofibers (CNFs)**: CNFs will be isolated from leaf sources, such as hemp or flax, using techniques like acid hydrolysis or mechanical grinding.

3. **Integration of Silk Fibroin**: Silk fibroin will be dissolved and blended with the collagen and CNF solutions to enhance the composite's overall mechanical properties.

4. **Formulation of Nanocellulose-Based Matrix**: A nanocellulose-based matrix will be prepared by dissolving bacterial cellulose in suitable solvents.

5. **Mixing and Casting**: All the components will be mixed thoroughly to ensure homogeneity before casting into molds.

6. **Drying and Post-Processing**: After casting, the composite material will undergo drying and post-processing steps, such as crosslinking, to improve its mechanical stability and durability.

This proposed composite material integrates the unique mechanical properties of collagen, CNFs, silk fibroin, and nanocellulose-based matrices. The combination of these components results in a tough, durable, and potentially biodegradable material that could find applications in various fields, including biomedical engineering, protective gear, and advanced composites.

(*@\bf\hlred{User:}@*) Summarize these design features in JSON format.

(*@\bf\hllightgreen{Assistant:}@*) Under specific requests for JSON format summarization:

```json
{
    "title": "Bioinspired Tough Composite Material Integrating Collagen, Cellulose Nanofibers, and Silk Fibroin",
    "insights": [
        "Combining collagen, cellulose nanofibers (CNFs), and silk fibroin creates a composite material with enhanced mechanical properties.",
        "A nanocellulose-based matrix improves the overall cohesion and toughness of the composite.",
        "The proposed composite material demonstrates potential applications in various fields, including biomedical engineering and protective gear."
    ],
    "facts": [
        "Collagen fibrils serve as the primary reinforcing fibers.",
        "Cellulose nanofibers (CNFs) offer excellent mechanical properties and facilitate strong interfacial bonding.",
        "Silk fibroin contributes additional mechanical strength and flexibility.",
        "A nanocellulose-based matrix binds the reinforcing fibers together.",
        "The composite material can be produced via mixing, casting, and drying processes."
    ],
    "details": {
        "Title": "Bioinspired Tough Composite Material Integrating Collagen, Cellulose Nanofibers, and Silk Fibroin",
        "Insights": [
            "Collagen + CNFs + SF + Nanocellulose-based matrix = Enhanced mechanical properties",
            "The composite material exhibits potential applications in various fields due to its unique mechanical properties."
        ],
        "Facts": [
            "Collagen fibrils are extracted and purified.",
            "Cellulose nanofibers (CNFs) are isolated from leaf sources.",
            "Silk fibroin is dissolved and blended with collagen and CNF solutions.",
            "A nanocellulose-based matrix is prepared by dissolving bacterial cellulose.",
            "The composite material is mixed, cast, dried, and post-processed to improve mechanical stability and durability."
        ]
    }
}
```

\end{lstlisting}
\end{LLMboxmultipage}

\clearpage
 
\textboxcaption{Conversation generated by the model: \texttt{lamm-mit/mistral-7B-v0.3-Base-CPT-SFT-SLERP}}
\label{TexBox:lamm-mit/mistral-7B-v0.3-Base-CPT-SFT-SLERP}

\begin{LLMboxmultipage}{Conversation with: \texttt{lamm-mit/mistral-7B-v0.3-Base-CPT-SFT-SLERP}}
\begin{lstlisting}

(*@\bf\hlblue{System:}@*) Your are a helpful and friendly assistant. You are an expert in materials science with special knowledge in biological materials. You provide concise but accurate responses. You are an expert in combining different concepts, often not obviously related, and provide accurate accounts of correct relationships or analogies. 

(*@\bf\hlred{User:}@*) Collagen and leaves, discuss their relationship.

(*@\bf\hllightgreen{Assistant:}@*) Collagen is a protein that is found in the connective tissues of animals, including humans. It provides strength and structure to these tissues. Leaves, on the other hand, are the photosynthetic organs of plants. They contain chlorophyll, which allows them to convert sunlight into energy.

The relationship between collagen and leaves can be seen in the way they both function as structural components in their respective organisms. Just as collagen provides support and stability to our bodies, leaves serve this purpose for plants by holding up the plant's weight and protecting it from wind and rain. Additionally, both collagen and leaves have unique properties that allow them to perform their functions effectively. For example, collagen fibers are flexible yet strong, while leaves are lightweight yet durable.

In summary, while collagen and leaves may seem very different at first glance, they share some similarities in terms of their roles as structural components in their respective organisms. Both are essential for maintaining the integrity and functioning of their respective systems.

(*@\bf\hlred{User:}@*) Thank you! Now develop a materials design that integrates collagen with key components of a leaf, to achieve a tough material.

(*@\bf\hllightgreen{Assistant:}@*) To create a tough material using collagen and key components of a leaf, we could consider designing a composite material that incorporates the following elements:

1. Collagen fibers: As mentioned earlier, collagen is a strong and flexible protein that provides support and structure to various tissues in the body. In this composite material, we would use collagen fibers as the primary reinforcement.

2. Cellulose nanofibers: Cellulose is the main component of plant cell walls and is known for its high tensile strength and stiffness. We could incorporate cellulose nanofibers into the composite material to enhance its mechanical properties.

3. Chlorophyll: Chlorophyll is the pigment responsible for giving leaves their green color and allowing them to carry out photosynthesis. While it doesn't contribute much to the mechanical properties of the material, it could potentially add some interesting visual effects.

4. Pectin: Pectin is a type of polysaccharide found in plant cell walls that helps to hold cells together. We could use pectin as a binder to help hold the composite material together.

To create the composite material, we would mix the collagen fibers, cellulose nanofibers, chlorophyll, and pectin together in a solution. The mixture would then be poured into a mold and allowed to dry, forming a solid composite material.

This composite material would likely exhibit high tensile strength and flexibility due to the presence of collagen fibers and cellulose nanofibers. The addition of chlorophyll might give the material a unique appearance, while the pectin would help to bind everything together. Overall, this composite material could potentially be used in a variety of applications where toughness and durability are important factors.

(*@\bf\hlred{User:}@*) Summarize these design features in JSON format.

(*@\bf\hllightgreen{Assistant:}@*) {
    "material_components": [
        "collagen_fibers",
        "cellulose_nanofibers",
        "chlorophyll",
        "pectin"
    ],
    "binder": "pectin",
    "reinforcement": "collagen_fibers",
    "enhancements": [
        "cellulose_nanofibers"
    ]
}


\end{lstlisting}
\end{LLMboxmultipage}

\clearpage
 
\textboxcaption{Conversation generated by the model: \texttt{lamm-mit/SmolLM-Base-1.7B-CPT-SFT-DPO}}
\label{TexBox:lamm-mit/SmolLM-Base-1.7B-CPT-SFT-DPO}

\begin{LLMboxmultipage}{Conversation with: \texttt{lamm-mit/SmolLM-Base-1.7B-CPT-SFT-DPO}}
\begin{lstlisting}

(*@\bf\hlblue{System:}@*) Your are a helpful and friendly assistant. You are an expert in materials science with special knowledge in biological materials. You provide concise but accurate responses. You are an expert in combining different concepts, often not obviously related, and provide accurate accounts of correct relationships or analogies. 

(*@\bf\hlred{User:}@*) Collagen and leaves, discuss their relationship.

(*@\bf\hllightgreen{Assistant:}@*) An intriguing pair! Collagen and leaves may seem like an unusual combination at first glance, but they do share some fascinating connections.

1. **Structural Support**: Both collagen and leaves have evolved to serve as structural support systems for their respective organisms. In the case of leaves, they provide mechanical stability against environmental forces such as wind, rain, and gravity. Similarly, collagen provides strength and integrity to various tissues within animals by forming a framework that supports other cellular components.

2. **Cellular Architecture**: The hierarchical structure of both collagen and leaves is similar - from macroscopic (leaves) to microscopic (collagen fibers). This architecture allows them to efficiently distribute stress and maintain functionality across multiple scales. For example, leaf veins branch out into smaller capillaries, which then feed into larger vessels, much like how collagen fibrils organize into lamellae, sheets, and ultimately form fiber bundles.

3. **Biomimicry**: The study of these natural structures has inspired innovative biomaterials and bioinspired designs. Researchers draw inspiration from the efficient load-bearing capabilities found in both leaves and collagenous tissues. For instance, synthetic composites mimicking the layered structure of leaves can be used in applications requiring high tensile strength and toughness, while engineered collagen scaffolds aim to replicate the complex organization seen in native tissues.

In summary, despite being seemingly unrelated, the connection between collagen's role in providing structural support and the functional design of leaves highlights the universal principles of material optimization observed throughout nature.

(*@\bf\hlred{User:}@*) Thank you! Now develop a materials design that integrates collagen with key components of a leaf, to achieve a tough material.

(*@\bf\hllightgreen{Assistant:}@*) What a great question! Here's a novel materials design concept that combines the strengths of collagen with key elements of a leaf:

**Material Name:** Leafy-Coraline (LC) Composite

**Principle:** Inspired by the hierarchical structure of leaves and the exceptional mechanical properties of coral reefs, this composite material leverages the unique characteristics of collagen and cellulose nanocrystals (CNCs) to create a tough, lightweight, and adaptive material.

**Composition:**

1. **Collagen Fibrils (CF):** Assemble into microfibers using a wet spinning technique, incorporating a crosslinking agent to enhance mechanical stability. CF diameters range from 50-400 nm, ensuring flexibility and resilience.
2. **Nanocrystalline Cellulose (NCC):** Incorporated into the CF matrix through a solvent casting method, resulting in CNCs with dimensions around 8-16 nm. These crystalline structures improve interfacial bonding and overall mechanical performance.
3. **Alginate Adhesive:** Used as a binder to ensure proper dispersion of CF and NCC within the LC matrix. Alginate forms strong hydrogen bonds with both CF and CNC, enhancing adhesion and cohesion.

**Design Features:**

1. **Hierarchical Organization:** The CF/NCC/alginate matrix mimics the multiscale structure of leaves, featuring a fibrous network embedded in a more amorphous matrix. This arrangement optimizes mechanical properties, allowing for enhanced toughness and resistance to deformation.
2. **Adaptability:** The incorporation of CNCs enables the LC composite to exhibit shape memory effects, enabling it to recover its original configuration after deformation. This property makes the material suitable for dynamic applications where adaptability is crucial.
3. **Self-Healing:** The alginate adhesive facilitates self-healing behavior upon damage, promoting rapid recovery of mechanical integrity without external intervention.

**Properties and Performance:**

* Toughness modulus: ~70 MPa
* Elastic modulus: ~9 GPa
* Strength: ~100 MPa
* Strain hardening capacity: >20%
* Shape memory response: reversible recovery up to 90% strain

The Leafy-Coraline composite demonstrates remarkable mechanical robustness, coupled with the ability to adapt to changing conditions and self-heal when damaged. Such features make it an ideal candidate for advanced applications in fields like aerospace, biomedical devices, and sustainable infrastructure development.

(*@\bf\hlred{User:}@*) Summarize these design features in JSON format.

(*@\bf\hllightgreen{Assistant:}@*) Here's a summarized JSON representation of the Leafy-Coraline composite design features:

```json
{
  "composition": {
    "microfibers": {
      "diameter": [50, 400],
      "material": "collagen",
      "adhesive": "alginate"
    },
    "nanocrystals": {
      "dimensions": [8, 16, 16],
      "material": "cellulose nanocrystals"
    }
  },
  "structure": {
    "hierarchical organization": true,
    "fiber network": true,
    "matrix composition": true
  },
  "properties": {
    "toughness_modulus": 70,
    "elastic_modulus": 9,
    "strength": 100,
    "strain_hardening_capacity": 20,
    "shape_memory_response": {
      "recovery_percentage": 90,
      "recovery_rate": 0.90
    }
  },
  "self-healing": true
}
```

This JSON represents the design features of the Leafy-Coraline composite, including its hierarchical structure, fiber network, matrix composition, and impressive mechanical properties along with the ability to self-heal under certain conditions.

\end{lstlisting}
\end{LLMboxmultipage}

\clearpage
 
\textboxcaption{Conversation generated by the model: \texttt{lamm-mit/SmolLM-Base-1.7B-CPT-SFT-DPO}, aimed to develop an image generation prompt.  The interactions include several stages, such as ideation and principle identification, the creation of the image generation prompt, and the solicitation of a short version of the prompt. }
\label{TexBox:lamm-mit/SmolLM-Base-1.7B-CPT-SFT-DPO_ImageGen_1}

\begin{LLMboxmultipage}{Conversation with: \texttt{lamm-mit/SmolLM-Base-1.7B-CPT-SFT-DPO}}
\begin{lstlisting}

(*@\bf\hlblue{System:}@*) Your are a helpful and friendly assistant. You excel in combining disparate concepts and develop well thought-out designs of organic, bio-inspired material microstructures. 
          
(*@\bf\hlred{User:}@*) First, think about specific design principles you can extract from combining dragline silk and collagen fibers to make a strong, lightweight but tough material. Incorporate design cues from leaf microstructure.

(*@\bf\hllightgreen{Assistant:}@*) To create a novel biomimetic composite with exceptional mechanical properties, I'll draw inspiration from the unique structural features found in natural materials like spider dragline silk and leaf microstructure. By combining these two sources, I aim to replicate the desirable characteristics of both: strength, toughness, and light weight.

1. **Dragline Silk's Hierarchical Structure**: Dragline silk exhibits an intricate hierarchical structure composed of nanofibrils that form larger fibril bundles. This arrangement allows for efficient load distribution during deformation, contributing to its remarkable tensile strength. Incorporating this hierarchical organization into my design will help achieve similar performance while maintaining flexibility.

2. **Leaf Microstructure's Gradient Properties**: Leaf microstructure is characterized by gradual transitions between different tissue types (e.g., epidermis, mesophyll) which provide enhanced mechanical stability through stress redistribution. Integrating such gradient properties within my composite could lead to improved resistance against crack propagation and increased overall toughness.

3. **Collagen Fiber's Toughness**: Collagen fibers possess high tensile strength due to their triple helix configuration, allowing them to withstand significant forces without breaking. By incorporating collagen fibers into my design, I can enhance the composite's ability to absorb energy before failure occurs.

4. **Interpenetrating Network Formation**: To mimic the synergistic effects observed in nature where multiple components interact to produce superior properties, I propose forming an interpenetrating network (IPN) using both dragline silk and collagen fibers. This IPN would allow for effective stress transfer across different phases, leading to optimized mechanical behavior akin to that seen in natural composites.

By integrating these design principles inspired by dragline silk and leaf microstructure, along with the toughening mechanisms inherent in collagen fibers, I believe it's possible to craft a new class of biomimetic materials that exhibit outstanding mechanical properties comparable to those found in natural biological systems.

(*@\bf\hlred{User:}@*) Use these thoughts to develop an image generation prompt for an AI system like Stable Diffusion. Instead of using technical terms, you carefully describe the image using geometric features such as lines, shapes, circles, holes, porous features, patterns, and so on. 

(*@\bf\hllightgreen{Assistant:}@*) Here's a potential image generation prompt based on your request:

**Prompt:** "Generate a surreal, abstract composition featuring a biomimetic composite material that combines the structural integrity of spider silk and the adaptive porosity of plant tissues."

**Description:**

* The generated image should showcase a complex interplay of geometric features reminiscent of natural structures, including:
	+ Lines and curves resembling the fibrous architecture of spider silk, evoking a sense of strength and resilience.
	+ Circles and spheres representing the interconnected networks found in plant cells, symbolizing growth and adaptability.
	+ Holes or voids mimicking the porous channels present in plant tissues, conveying a sense of permeability and responsiveness to environmental stimuli.
	+ Porous regions suggestive of the open cellular structures typical of plant leaves, hinting at dynamic interactions and exchange processes.
	+ Patterns and textures derived from the intersection of these geometric elements, creating a visually intriguing and dynamic composition.

The resulting image should capture the essence of biomimicry, blending the robustness of spider silk with the versatility of plant tissues, ultimately yielding a captivating representation of innovative material science.

(*@\bf\hlred{User:}@*)  Give me one short prompt that I can use in my AI generator, but be specific about the features to be incorporated and how.

(*@\bf\hllightgreen{Assistant:}@*) Here's a concise prompt to get you started:

"Generate a futuristic, eco-friendly architectural concept utilizing a biomimetic composite material that integrates the structural efficiency of spider silk with the adaptive porosity of plant tissues. Utilize the following key features:

* Fibrous architecture inspired by spider silk, represented by sinuous lines and curved forms.
* Interconnected, spherical nodes reminiscent of plant cell walls, emphasizing growth and adaptation.
* Open cellular structures echoing the permeable nature of plant leaves, suggesting dynamic exchanges and self-regulation capabilities.
* Gradations of opacity and transparency inspired by the varying densities found in plant tissues, highlighting functional differentiation and multi-functionality."

This prompt encourages the AI generator to incorporate specific visual elements drawn from nature, ensuring a cohesive and engaging final output.

\end{lstlisting}
\end{LLMboxmultipage}

\clearpage
 
\textboxcaption{Conversation generated by the model: \texttt{lamm-mit/mistral-7B-v0.3-Base-CPT-SFT-DPO}, aimed to develop an image generation prompt. The interactions include several stages, such as ideation and principle identification, the creation of the image generation prompt, and the solicitation of a short version of it. }
\label{TexBox:lamm-mit/mistral-7B-v0.3-Base-CPT_SFT_DPO_ImageGen_2}

\begin{LLMboxmultipage}{Conversation with: \texttt{lamm-mit/mistral-7B-v0.3-Base-CPT-SFT-DPO}}
\begin{lstlisting}

(*@\bf\hlblue{System:}@*) Your are a helpful and friendly assistant. You a creative expert in materials science with special knowledge in biological materials. You excel in combining disparate concepts and develop well thought-out designs of bio-inspired material microstructures.

(*@\bf\hlred{User:}@*) First, think about design principles you can extract from combining spider silk and collagen to make a strong, lightweight but tough material. Incorporate design cues from leaf microstructure. 

(*@\bf\hllightgreen{Assistant:}@*) As your specific roles include being a helpful and friendly assistant, I'll address you specifically under the supposition that you're engaging in a research and development task.

As a creative expert in materials science with a focus on biological materials, I would like to highlight the following steps for developing a novel bio-inspired material microstructure:

1. **Extraction of Key Design Principles**:
   - **Spider Silk**: Known for its exceptional mechanical properties such as high tensile strength (up to 1.7 GPa), elasticity, and toughness due to its unique molecular structure and alignment during spinning.
   - **Collagen**: A structural protein found in various tissues, known for its strength and flexibility when arranged hierarchically.
2. **Combining Spider Silk and Collagen**:
   - By mimicking the nanostructure of spider silk and the hierarchical organization of collagen, we aim to create a composite material that combines the best features of both proteins.
   - The proposed microstructure will consist of aligned nanofibers made of recombinant spider silk embedded within a collagen matrix. This combination will provide a balance between strength, toughness, and elasticity.
3. **Leaf Microstructure Cue**:
   - Leaves exhibit a unique microstructure with veins acting as stiffening elements. We can apply this concept by creating a network of interconnected "veins" using the aligned spider silk nanofibers. These veins will act as load-bearing reinforcements, enhancing the overall mechanical performance of the composite.
4. **Designing the Microstructure**:
   - The proposed microstructure consists of three main layers:
     - A bottom layer of densely packed collagen fibers to provide initial rigidity.
     - An intermediate layer comprising the network of aligned spider silk nanofibers, forming the "veins".
     - A top layer consisting of a looser arrangement of collagen fibers to allow for some flexibility while maintaining structural integrity.
5. **Evaluation and Testing**:
   - Conduct comprehensive testing to validate the mechanical properties of the developed material, including tensile strength, elastic modulus, toughness, and fatigue resistance.
6. **Scaling and Applications**:
   - Once the optimal microstructure is established, scale up the production process to manufacture the material into various forms suitable for applications in industries requiring lightweight yet robust materials, such as aerospace, automotive, and biomedical engineering.

By combining the design principles of spider silk, collagen, and leaf microstructure, we can create a bio-inspired material microstructure that offers superior mechanical properties, making it an ideal candidate for numerous advanced applications.

(*@\bf\hlred{User:}@*) Use these thoughts to develop an image generation prompt, incorporating the keyword 'leaf microstructure' when appropriate. Instead of using technical terms, you carefully describe the image using generic features such as lines, shapes, circles, holes, patterns, and so on. 

(*@\bf\hllightgreen{Assistant:}@*) To visually represent the bio-inspired material microstructure, let's create an image generation prompt that incorporates the key elements:

**Image Generation Prompt:**

Imagine a vibrant, lush green forest, where among the verdant leaves, a mesmerizing web-like pattern emerges. This intricate network is composed of delicate, glistening threads, each one a strand of aligned, golden-hued spider silk, woven together to form a complex, leaf-inspired architecture.

The threads stretch outwards, branching off into multiple directions, much like the veins of a leaf, providing structural support and reinforcing the overall integrity of the material. Each thread is thin, yet strong, resembling the fine lines of a leaf's veins, which are themselves supported by sturdy collagen fibers.

These collagen fibers, arranged in a crisscross pattern, form the foundation of the microstructure, anchoring the spider silk threads firmly in place. They are thicker than the spider silk, offering additional strength and rigidity to the composite material.

In the center of this intricate network lies a circular region, representing the core of the material. This area is composed of a more disordered arrangement of collagen fibers, allowing for some flexibility while still maintaining structural integrity.

Together, the interwoven threads and collagen fibers create a mesmerizing, organic pattern that embodies the essence of nature's resilience and adaptability. This microstructure promises to deliver unparalleled mechanical properties, making it an ideal candidate for applications requiring lightweight yet robust materials.

(*@\bf\hlred{User:}@*)  Give me one short prompt that I can use in my AI generator.

(*@\bf\hllightgreen{Assistant:}@*) Specifically for a direct and concise image generation prompt related to the bio-inspired material microstructure, here's a one-liner:

"Generate an image of a golden spider web network intertwined with collagen veins, forming a dynamic, leaf-inspired microstructure amidst a lush green background."
\end{lstlisting}
\end{LLMboxmultipage}

\end{document}